\documentclass[journal]{IEEEtran}  

\IEEEoverridecommandlockouts                              

\usepackage[utf8]{inputenc}
\usepackage[T1]{fontenc}

\usepackage{graphics} 
\usepackage{epsfig} 
\usepackage{amsmath} 
\usepackage{amssymb}  
\usepackage{optidef}
\usepackage{multirow}
\usepackage{comment}
\usepackage{subcaption}
\usepackage[linesnumbered,ruled,vlined,noend]{algorithm2e}

\usepackage[shortlabels]{enumitem}
\usepackage{booktabs}
\usepackage{adjustbox}
\usepackage{xspace}
\usepackage{balance}
\usepackage{tabularx}
\usepackage{longtable}
\usepackage[bookmarks=false,pagebackref=true,breaklinks=true,colorlinks]{hyperref}
\usepackage[dvipsnames]{xcolor}
\usepackage{dblfloatfix}

\newcommand{\FK}{\mathrm{FK}}
\newcommand{\cfg}{\mathbf{x}}
\newcommand{\Cfg}{\mathbf{X}}
\newcommand{\shortname}{DiffCo\xspace}
\newcommand{\rulesep}{\unskip\ \vrule\ }
\newcommand{\FKPH}{{\mathrm{FKPH}}}
\newcommand{\cscore}{\psi}
\newcommand{\supportlabel}{\mathbf{Y_S}}
\newcommand{\updated}[1]{{#1}} 

\newcommand{\second}[1]{{#1}} 
\SetKw{Break}{break}

\title{\LARGE \bf
\shortname: Auto-\underline{Diff}erentiable Proxy \underline{Co}llision Detection with Multi-class Labels for Safety-Aware Trajectory Optimization
}

\author{Yuheng Zhi$^{1}$, Nikhil Das$^{1}$ and Michael Yip$^{1}$
\thanks{$^{1}$Yuheng Zhi, Nikhil Das and Michael Yip are with the Department of Electrical and Computer Engineering, University of California San Diego, La Jolla, CA 92093 USA
        {\tt\small \{yzhi, nrdas, yip\}@ucsd.edu}}%
}

\begin{document}

\maketitle
\thispagestyle{plain}
\pagestyle{plain}

\begin{abstract}
The objective of trajectory optimization algorithms is to achieve an optimal collision-free path between a start and goal state. In real-world scenarios where environments can be complex and non-homogeneous, a robot needs to be able to gauge whether a state will be in collision with various objects in order to meet some safety metrics. The collision detector should be computationally efficient and, ideally, analytically differentiable to facilitate stable and rapid gradient descent during optimization. However, methods today lack an elegant approach to detect collision differentiably, relying rather on numerical gradients that can be unstable. We present \shortname, the first, fully auto-differentiable, non-parametric model for collision detection. Its non-parametric behavior allows one to compute collision boundaries on-the-fly and update them, requiring no pre-training and allowing it to update continuously in dynamic environments. It provides robust gradients for trajectory optimization via backpropagation and is often 10-100x faster to compute than its geometric counterparts. \shortname also extends trivially to modeling different object collision classes for semantically informed trajectory optimization. 
\end{abstract}


\section{Introduction}
\subsection{Collision Detection in Robotics}
Motion planning algorithms and trajectory optimization problems often require solving for trajectories that steer clear of obstacles. 
Simply evaluating whether a robot is going to collide with any obstacle, for any step or line segment in the trajectory, requires 
a geometric collision checker. These collision checkers are algorithms that input robot and environment scene information and compare mesh elements (e.g., triangles) of the robot with meshes of the surrounding objects and obstacles. 
\second{Flexible Collision Library (FCL), LIBCCD \cite{pan2012fcl, libccd} and their base algorithms, Gilbert-Johnson-Keerthi (GJK) and Expanding Polytope Algorithm (EPA) \cite{gjk, epa} are the gold standards today that follow this polygonal collision checking approach. }
With these approaches, often hundreds to tens of thousands of possible colliding elements pairs need to be tested in order to verify whether a waypoint along a trajectory is in collision or not. This computational requirement is then repeated not only on every waypoint, but also on every edge between waypoints using a fine discretization of the path to ensure all intermediate points along the trajectory are also collision-free. 
Due to the high frequency of queries and the complexity of each query, collision detection is the most computationally expensive operation for all motion planning problems \cite{kingston2018sampling, elbanhawi2014sampling}. 

\begin{figure}[t]
    \centering
    \includegraphics[width=.95\linewidth, trim=1.15cm 0.4cm .4cm .5cm, clip]{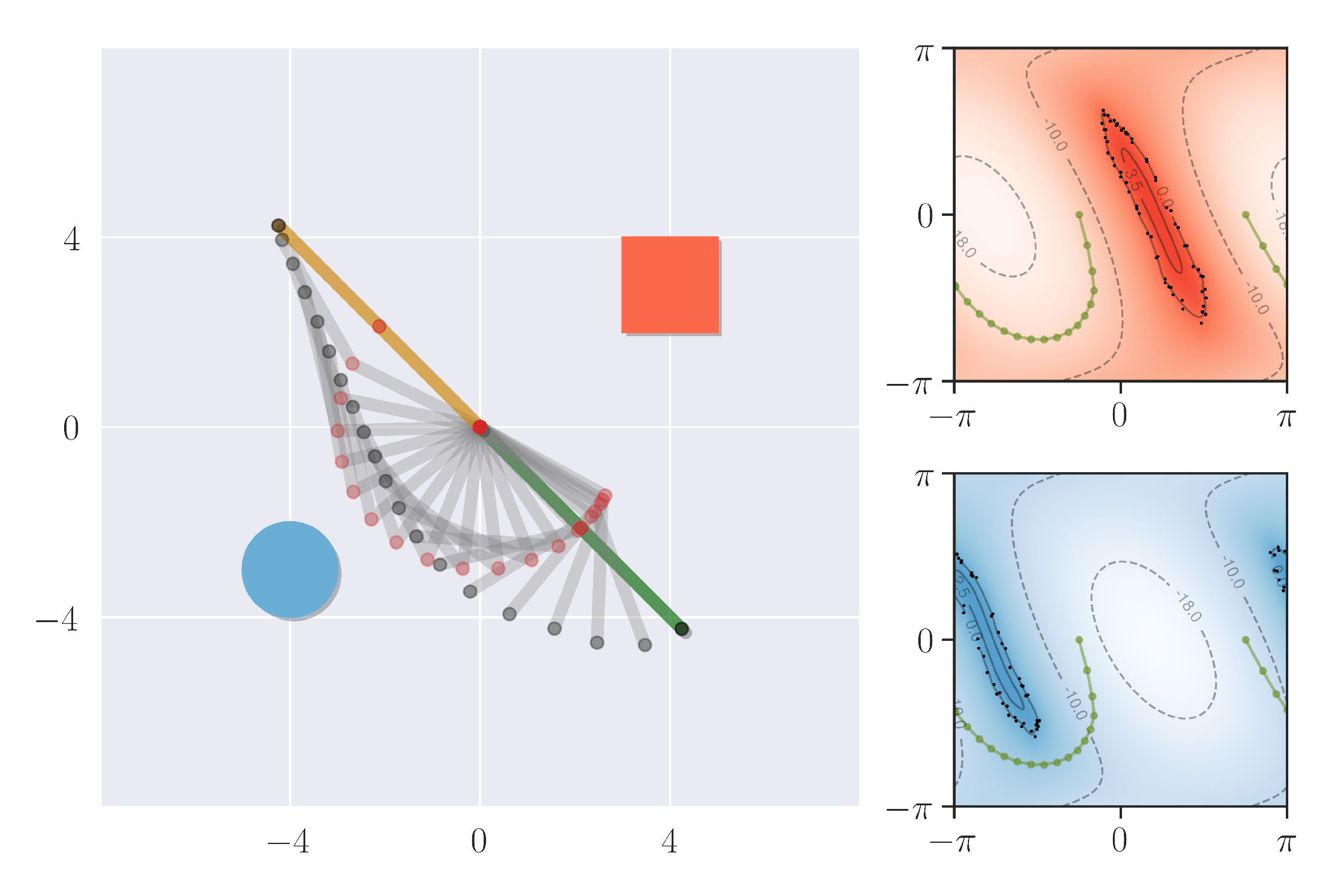} 
    \caption{An example of an optimized robot trajectory with the proposed proxy differentiable collision detector \shortname. The solution is shown in both the workspace and the configuration space. 
    The position of the first joint of the 2-DOF planar robot is fixed and both joints have no angular limits. In the left figure, the start state is green, the goal is brown, and the waypoints are gray.
    The clearance constraint of the orange object is set to be 10 times larger than the blue one, thus resulting in a trajectory that stays further to the orange object than to the blue one.
    The value of a point on one of the right figures is the \shortname collision score between the robot and the obstacle of the same color as the figure.
    The green lines show the trajectory in $\mathcal{C}$-space. Note the wrapping-around property of the angular space.}
    \label{fig:opening}
\end{figure}

Machine learning has recently been introduced to learn binary classifiers that are used in place of geometric collision checks during motion planning and  trajectory optimization. These binary classifiers output labels for $\mathcal{C}_\mathrm{free}$ or $\mathcal{C}_\mathrm{obs}$ for any input configuration. They are functions that can approximate the mapping from a robot configuration to a label over the entire configuration space and thus are often referred to as \textit{proxy} collision detectors \cite{das2020learning, das2020forward, pan2015efficient, huh2016learning}. 
The purpose of having a proxy model to evaluate for collision is primarily towards addressing the computational burden of geometric checking, as these proxy models can be much more computationally efficient.
A common pipeline of these proxy collision detectors is to first get the initial ground-truth collision labels from geometric collision detectors for a set of randomly sampled configurations at time $t=0$, and then learn the initial model of the configuration space using these samples. Some proxy collision detectors are designed for online (non-stationary) distributions, and so incorporate an active learning stage which usually involves efficient sampling strategies to refine the model and account for movements of obstacles at every timestep \cite{das2020learning}. 

Compared to traditional geometric collision detectors, proxy collision detectors involve additional steps of initial model training and online model updating; however, they still result in computational gains even with every additional cost taken into account \cite{das2020learning, pan2015efficient, huh2016learning}. 
This is attributed to their sparse, computationally lightweight approximations. The training algorithms are also specially designed to achieve fast convergence so that they may run in real-time. 
Finally, the updating step is usually much more data efficient than the training step \cite{das2020learning,das2020forward}; the sampling strategies used to update the models minimize the number of new samples by utilizing the existing label boundaries to directly inform sampling, allowing them to track changing environments or to refine their models in a sample-efficient way.

\subsection{The Need for Differentiable Collision Detection in Trajectory Optimization}
Main-stream trajectory optimization methods rely on gradient-based, constrained, non-linear optimizers to get a solution \cite{trajopt, chomp, kalakrishnan2011stomp}, so they can only work with differentiable objectives and constraints, which brings the need for differentiable collision detectors. Numerical differentiation is acceptable for some algorithms, e.g. CHOMP relies on a pre-computed distance-to-collision map for collision distance estimation and computing configuration gradients \cite{chomp}. GRIPS provides a pre-computed distance measure for wheeled mobile robots \cite{heiden2018grips}. However, the number of states needed to be pre-computed grows exponentially as the degree of freedom increases, so analytically differentiable collision estimations are preferred for computational efficiency. Furthermore, with the rise of graphical processing units (GPU) and the significant developments and ease of auto-differentiation made available through deep learning packages, as well as the increasing popularity of end-to-end neural network-based motion planning approaches \cite{ qureshi2019motion, ichter2018learning, oraclenet, qureshi2020constrained, qureshi2018deeply}, a differentiable model for collision detection is especially desirable and would be easy to integrate. 

Finally, an interesting consideration for planning a feasible path for a robot in the real world is that most obstacles and objects are not alike and avoidance among objects should ideally not be treated equally. Knowledge about the object labels and their \textit{semantics} can inform the safety or severity of moving close to them when computing motion plans. These labels can be used to define metrics that emphasize avoiding accidental collisions with sensitive objects, such as humans in the workspace, or risky objects (glass, liquid-filled containers, stacks of objects, electronics, etc.); in robotic surgery, instruments, or organs have different sensitivities that must be taken into account. Thus, as an additional consideration when detecting collisions, an associated label is also advantageous so that \textit{semantically informed planning} can be conducted. 
This idea has never been integrated at the level of the collision detector itself, which if possible, could allow planners and trajectory optimizers to, at the lowest level, pick and choose priorities in avoiding collision between object classes with different risk profiles.




\begin{figure*}[htbp!]
    \centering
    \includegraphics[width=.9\linewidth]{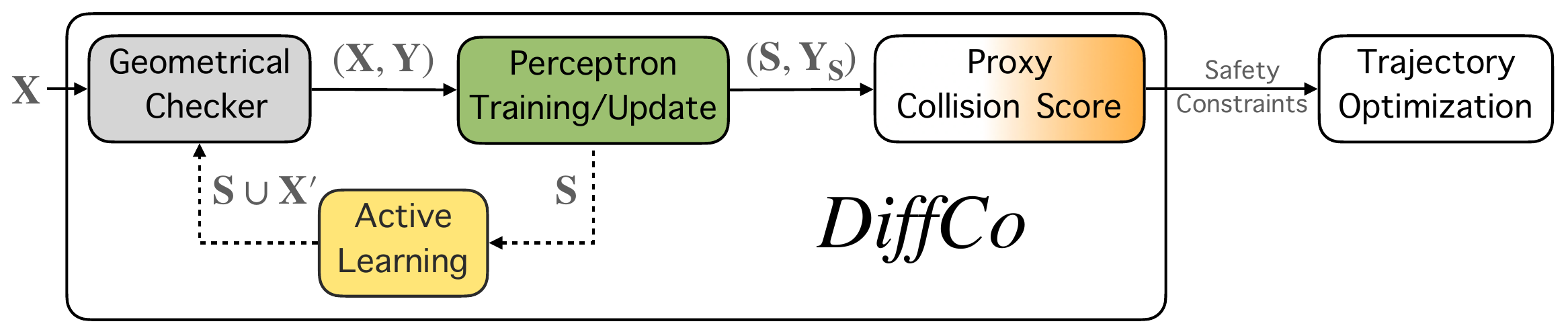}
    \caption{
    Overall algorithm pipeline of building a differentiable proxy collision detector with active learning updates. The dashed lines describe active learning data generation procedure based on \textit{exploration} and \textit{exploitation}. $\Cfg$ is an initial set of robot configurations, $\mathbf{Y}$ is the set of corresponding collision labels generated by the geometrical collision checker. $\mathbf{S}$ is a sparse set of support configurations extracted from $\Cfg$ and $\mathbf{Y_S}$ is the corresponding labels. $\Cfg'$ are configurations proposed by the active learning approach to keep track of changes in the environment. Instance-level auto-differentiable safety constraints such as collision avoidance can be constructed from the proxy collision detector.}
    \label{fig:main_flow}
\end{figure*}

\subsection{Contributions in This Work} 

In this paper, we introduce a \textit{differentiable proxy collision detector} called \shortname that checks collision with objects using only differentiable operations. The \textit{proxy} collision detector generates a score of collision using a non-parametric, kernel-based function approximator. 
This provides a substantial advantage over parametric models such as neural networks that are neither easily updated nor able to adapt to more complex environments. Furthermore, we explore the use of class-specific \shortname models to provide independent proxy collision detection per object class in the environment (i.e., cat versus table, all inanimate objects versus all living creatures, etc.). Most importantly, we describe the entire model in a completely differentiable way, enabling it to be easily integrated into auto-differentiation packages available in deep learning libraries such as PyTorch. We show how the method, being non-parametric, retains the ability to adapt its model online to changing environments using an active learning approach. We finally show how the model can be used effectively in trajectory optimization, where its gradient may be used for finding feasible paths that are safety-optimized and semantically informed based on different object classes.


To evaluate the capabilities of this model over a range of configuration spaces, we demonstrate results across toy problems to full 7-DOF robot manipulators. 
The proposed approach is shown to be computationally efficient for trajectory optimization, to the point that it is faster than a geometric collision checker would be solving the same problem. Fig.~\ref{fig:opening} shows the conceptual figure of a 2-DOF planar robot trajectory optimized using the collision gradients given by \shortname, both in the workspace and configuration space. 


\subsection{More Related Work}

\paragraph{Distance to Collision}
Not many previous works have explicitly studied the problem of differentiable collision checking, but many have worked on modeling the distance to collision, which is a continuous indicator of collision. For collision-free configurations, the distance to collision is the minimum distance from points on the robot to obstacles; for in-collision configurations, the distance to collision is the negative value of the maximal penetration depth. 
Distance to collision is sometimes included as a byproduct of the geometrical collision check, e.g., 
Gilbert–Johnson–Keerthi distance algorithm (GJK) and the Expanding Polytope Algorithm (EPA) \cite{gjk, epa}. Although accurate, these methods are computationally expensive in complicated scenarios. \second{A common practice is to enclose the robot links by a series of convex shape primitives like spheres and capsules; a pre-computed signed distance field (SDF) of the environment is queried by the primitives to approximate distance to collision \cite{chomp, kalakrishnan2011stomp, gpmp2}. A more fine-grained approach also exists to strictly convexify geometrical shapes up to arbitrary precision to achieve smooth distance to collision \cite{escande2014strictly}.} On the other hand, most proxy collision detectors have yet to explore this notion of distance to collision. 

A recent work using neural networks \cite{kew2019neural} and another work using Gaussian Processes \cite{das2020stochastic} have shown initial success in estimating distance to collision using proxy collision detectors. The first work takes a robot configuration and a workspace configuration as input and uses neural networks for estimating distance to collision.  Unfortunately, its model is constrained to only those environments covered by the pre-defined \textit{workspace configuration space} in the training set and has yet to be shown to be scalable. The second work by Das and Yip \cite{das2020stochastic} uses Gaussian Process to overcome the rigidity and training data cost of the neural network definition through its use of the non-parametric Gaussian Process model, and as a bonus, provides a probabilistic estimation of distance to collision which is very useful in real life due to noisy sensors; however, 
the GP model itself is not very computationally efficient without significant modifications for online usage \cite{wilcox2020solar}. 
In the same paper, a deterministic, kernel regressor is also proposed that helps to reduce computational burden but at the cost of providing false labels if sensor noise is considered. 

Other related works have succeeded in their particular applications and are noted here. An approach for estimating penetration depths (rather than distance to collision) between two objects trains an SVM to model the in-collision configuration space \cite{pan2013efficient}. This method could theoretically be extended to capture distance to collision, but was not done in the paper. Furthermore, the SVM model would be too computationally expensive to update for online robot control in dynamic environments. 
Flacco et al. \cite{flacco2012depth} uses the concept \textit{depth space} to estimate distance to obstacle points using sensor information for online collision avoidance, but using the approach to conduct global motion planning may bring unnecessary perception computation as the environmental information is usually assumed to be known by the planning algorithms. It also cannot be used to calculate distance to collision for hypothetical configuration queries sent out by planning algorithms. RelaxedIK \cite{rakita2018relaxedik} employs neural networks to approximate self-collision distances between links simplified into line segments, which can successfully avoid self-collision of humanoid robots with high DOF. However, extending the same method to general object collision checking can be challenging due to the amount of data required to train the neural network.


\paragraph{Planning with Semantics}
Besides collision detection, some motion planning algorithms uses knowledge about the content of objects in the environment, because some classes of objects should be treated differently 
due to their unique behavioral patterns or different levels of importance \cite{rajaram2016refinenet, objcontent2, objcontent3}. An algorithm designed for motion planning in autonomous driving uses a semantic hierarchy to forecast the movement of pedestrians, bicycles, and motor vehicles \cite{uber2020perceive}. The \shortname approach that will be introduced in this paper is able to provide the object category information needed in these algorithms.

\paragraph{Trajectory Optimization}
CHOMP \cite{chomp} and TrajOpt \cite{trajopt} are trajectory optimization algorithms that use gradient information. 
They rely on either geometrical collision detection algorithm or pre-computed maps to get gradients. On simple tasks, even when initialized by an inadmissible trajectory, they can usually yield more optimal, feasible trajectories using less time than sampling-based motion planning algorithms. In this paper, however, more general and accessible non-linear optimization algorithms, Adam \cite{adam} and SLSQP \cite{slsqpKraft1994algorithm, slsqpKraft1988software}, are used to perform trajectory optimization under constraints.

\section{Method}
In the following section, the components comprising the auto-differentiable proxy collision detector will be described: 
(A) will describe modeling the configuration space using a multi-label kernel perceptron preserving knowledge of object categories; 
(B) will describe the approach in achieving a differentiable model using polyharmonic splines;
\updated{(C)} will extend the method to modeling dynamical environments using an active learning strategy to update the learned model upon obstacle changes, such as movement, rotation, appearance, or disappearance;
Finally, \updated{(D)} will describe the constrained trajectory optimization problem that employs the proxy collision estimator to compute constraint values in a way that can be easily integrated with deep learning packages such as PyTorch.  
Fig.~\ref{fig:main_flow} gives an overview of the pipeline. 
See Table~\ref{tab:symbols} for a full list of mathematical symbols used.

\subsection{Multi-label Kernel Perceptron Model and Training}\label{sec:perceptron}

The first task towards learning a differentiable proxy collision detector is to model the configuration space. 
A configuration of a robot with $D$ degrees of freedom is a $D$-dimensional vector $\cfg \in \mathbb{R}^D$, where the $i$-th entry is the value of the $i$-th joint angle position. We consider a cluttered workspace with obstacles across $C$ categories. Configurations in collision with obstacles of category $c\in C$ forms a subspace of collisions $\mathcal C_\mathrm{obs}^c$, and its complement is $\mathcal C_\mathrm{free}^c$.

We propose to model the configuration space by finding a sparse set of configurations
near the boundary between each pair of $\mathcal{C}_\mathrm{obs}^c$ and $\mathcal{C}_\mathrm{free}^c$. 
These configurations contain important geometrical information in the workspace. We name these representative configurations as \textit{support configurations}. Since a similar concept exists in the formulation of Support Vector Machines \cite{svm}, we also use the term \textit{support points}, \textit{support set} interchangeably. 



To find a support set of configurations, the first step is to collect a dataset of $N$ \second{uniformly sampled} robot configurations, $\mathbf{X}\in \mathbb{R}^{N\times D}$ labeled by the categories of obstacles the robot will collide with, $\mathbf{Y} \in \{+1, -1\}^{N\times C}$. 
Configuration $\mathbf{x}_i \in \mathbb{R}^D$ is the $i$-th row of $\mathbf{X}$. Its corresponding label $\mathbf{y}_i \in \{+1, -1\}^C$ is the $i$-th row of $\mathbf{Y}$. $\mathbf{y}_i$ has $C$ entries, and the $c$-th entry $\mathbf{y}_{ic}$ labels whether configuration $\mathbf{x}_i$ is in collision with objects of category $c$ ($+1$ for \textit{in-collision} and $-1$ for \textit{collision-free}).
This $(\Cfg, \mathbf{Y})$ dataset needs to be multi-label to help capture the category information of objects as it is possible to be in collision with several obstacles at once. One can also define each instance of objects to be a unique category to achieve instance-level modeling, a strategy that will be discussed later. 
The configurations for the initial dataset, at time = 0, are collected by uniformly sampling the configuration space with a geometrical collision checking algorithm such as FCL \cite{pan2012fcl} to generate labels. This most computationally expensive step of the method (typically still only a few seconds) is all done when a robot is turned on, and yet only needs to run once. Changes to the environment will be updated to the model during active learning at a much faster rate.  

Given the dataset ($\mathbf{X},\mathbf{Y}$), the goal of the multi-label perceptron is to find a sparse set of support configurations $\mathbf{S}\in \mathbb{R}^{M\times D}~ (M \leq N)$ \cite{kernelperceptron}. This is done via  a sparse update rule. 
The first key component of kernel perceptron is the employed kernel function $k(\cdot, \cdot)$ and the corresponding kernel matrix $\mathbf{K} \in \mathbb{R}^{N\times N}$. The $(i, j)$-th element of $\mathbf{K}$ is calculated by
\begin{equation}
    \mathbf{K}_{ij} = k(\mathbf{x}_i, \mathbf{x}_j), \quad \forall \mathbf{x}_i, \mathbf{x}_j \in \mathbf{X}.
\end{equation}
The kernel function $k$ depends on the system and the task, though the radial basis function (RBF) is popular due to its generality \cite{das2020learning}. In this work, we will use the Forward Kinematics (FK) kernel \cite{das2020forward} as it tends to build more accurate and more sparse models for robot manipulators. The FK kernel chooses several points along the robot body that define its gross shape, and performs kernel distance evaluation between those points. Fundamentally, it applies forward kinematics transform prior to evaluating an RBF distance (see Appendix for FK kernel equations). Note that other kernel functions may have been used instead if one is not using a robot manipulator, without loss of generality (as demonstrated in \cite{das2020learning} and \cite{das2020forward}). 

The trainable parameters of a \textit{vanilla} kernel perceptron are an $N$-dimensional vector of weights, each associated with a sampled configuration. However, if we are performing multi-label classification, we need to define the parameters as a matrix of weights $\mathbf{W} \in \mathbb{R}^{N\times C}$, each row of which is associated with a configuration in the dataset, initialized all as 0. The hypothesis matrix $\mathbf{H} \in \mathbb{R}^{N\times C}$ is computed by
\begin{equation}
    \mathbf{H} = \mathbf{K}\mathbf{W},
    \label{eq:hypothesis}
\end{equation}
which are the predicted scores of each configuration being in collision with obstacles of each category. The matrix of margins $\mathbf{M} \in \mathbb{R}^{N\times C}$ is defined to be
\begin{equation}
    \mathbf{M} = \mathbf{Y}\odot \mathbf{H} = \mathbf{Y}\odot \mathbf{(KW)},
\end{equation}
where $\odot$ stands for element-wise product. A correctly classified configuration should have only positive values in its corresponding row of $\mathbf{M}$. Thus the training objective of this kernel perceptron is to choose a $\mathbf{W}$ so that all elements of $\mathbf{M}$ are positive,
\begin{equation}
    \mathbf{W} \in \left\{\mathbf{W}~|~\mathbf{M}_{ij} > 0,\ \forall i,j \in \{1,2,...,N\}\right\}.
\end{equation}

To allow the fast nature of this pipeline, the kernel perceptron is trained with a sparse update rule described in the loop in Algorithm~\ref{algo:sparse}.
\begin{algorithm}[tb]
\DontPrintSemicolon 
\KwIn{Configurations $\mathbf{X}$, corresponding collision labels $\mathbf{Y}$, initial weights $\mathbf{W}$, initial hypothesis $\mathbf{H}$.}
\KwOut{Support configurations $\mathbf{S}$ and corresponding labels $\supportlabel$}
$\text{Completed} \leftarrow \{\text{false}\} \times C$\;
\For{it $\leftarrow 1$ to maxIteration}{
    $\mathbf{M} \leftarrow \mathbf{Y}\odot \mathbf{H}$\;
    \For{$c\leftarrow 1$ to $C$}{
        \tcp{Update misclassification}
        $i\leftarrow \arg\min_i \mathbf{M}_{ic}$\;
        \uIf{kernelNotComputed$(\mathbf{x}_i)$}{ \label{algline:checkkernel}
            $\mathbf{K}_{(\cdot)i} \leftarrow k(\mathbf{x}_i, \mathbf{X})$, $\mathbf{K}_{i(\cdot)} \leftarrow \mathbf{K}_{(\cdot)i}$\;
        }
        \uIf{$\mathbf{M}_{ic} \leq 0$}{
            $ \delta \leftarrow (\mathbf{y}_{ic}-\mathbf{H}_{ic})/\mathbf{K}_{ii}$\;
            $\mathbf{W}_{ic} \leftarrow \mathbf{W}_{ic} + \delta$\;
            $\mathbf{H}_{(\cdot)c} \leftarrow \mathbf{H}_{(\cdot)c } + \delta \mathbf{K}_{(\cdot)i}$\;
            \textbf{continue}\;
        }
        \tcp{Zero redundant weights}
        $\mathbf{M}'_{(\cdot)c}\leftarrow \mathbf{Y}_{(\cdot)c}\odot(\mathbf{H}_{(\cdot)c}-\mathbf{W}_{(\cdot)c}\odot\mathrm{diag}~\mathbf{K})$\;
        $i\leftarrow \arg\max_i \mathbf{M}'_{ic}$ subj. to $\mathbf{W}_{ic} \neq 0$\; 
        \uIf{$\mathbf{M}'_{ic} > 0$}{
            $\mathbf{H}_{(\cdot)c} \leftarrow \mathbf{H}_{(\cdot)c} - \mathbf{W}_{ic} * \mathbf{K}_{(\cdot)i}$\;
            $\mathbf{W}_{ic} \leftarrow 0$\;
            \textbf{continue}\;
        }
        \tcp{No more updates in $c$}
        
        $\text{Completed}[c] \leftarrow true$\;
    }
    \uIf{all(Completed)}{
        \textbf{break}\;
    }
}
$\mathbf{S} \leftarrow$ \textit{removeConfig}$(\mathbf{X}, \cfg_i)$ st. $\mathbf{W}_{i(\cdot)} = \mathbf{0}$\;
$\supportlabel \leftarrow$ \textit{removeLabel}$(\mathbf{Y}, \mathbf{Y}_i)$ st. $\mathbf{W}_{i(\cdot)} = \mathbf{0}$\;

\KwRet{$\mathbf{S}, \supportlabel$}
\caption{\sc SparseMultilabelPerceptron}
\label{algo:sparse}
\end{algorithm}
Das and Yip \cite{das2020learning} described a single-label version of this algorithm and provided theoretical convergence guarantees. Here we extend the method to a multi-label update rule.
\begin{figure*}
    \centering
    \begin{subfigure}{0.39\linewidth}
    \includegraphics[width=\linewidth, trim={0, -0.35\linewidth, 0, -0.35\linewidth}, clip=True]{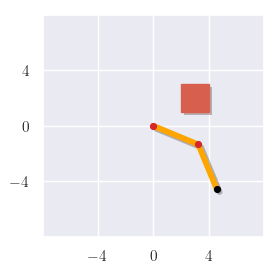}
    \caption{Workspace}
    \end{subfigure}
    \rulesep
    \begin{subfigure}{0.59\linewidth}
    \includegraphics[width=\linewidth]{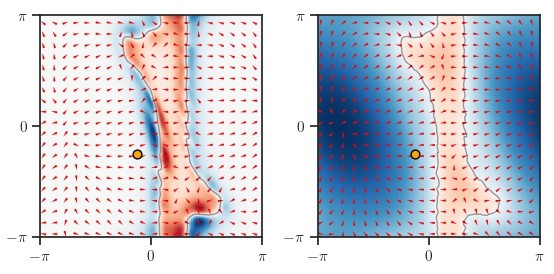}
    \includegraphics[width=\linewidth]{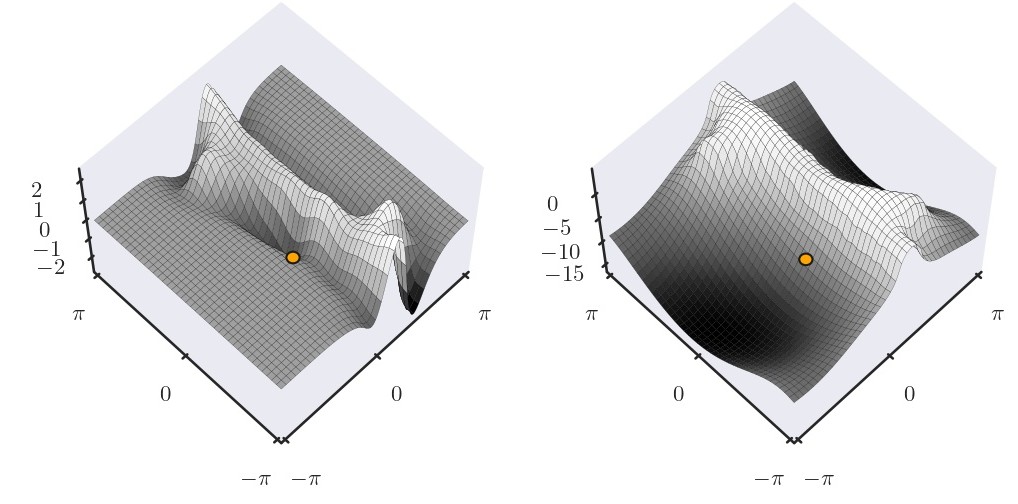}
    \caption{$\mathcal{C}$-space using rational quadratic function (left) and polyharmonic function (right)}
    \end{subfigure}
    
    \caption{A demonstration of the differences with kernel function choice. Given the two-link robot on the left, the middle figures visualize raw hypothesis score from a kernel perceptron trained using a rational quadratic function. The right figures are the proposed polyharmonic function. Red regions mark higher score, blue regions mark lower score, white regions are close to 0. The orange dots are the current configuration of the robot. Red arrows point to the direction of the \textbf{negative} gradients, which \textit{should} indicate an action that guides the robot away from collision. One can see that, though the rational quadratic (and also, by extension, the Gaussian) kernel is more popular in general for kernel methods, the gradient directions sometimes do not lead the robot in a direction away from collision, whereas the polyharmonic function does. Also, there are many local minimums in both $\mathcal{C}_\mathrm{free}$ and $\mathcal{C}_\mathrm{obs}$, which means the hypothesis score given by rational quadratic kernel does not make a consistent differentiable collision detector. (The 3D score maps are Gaussian-filtered to increase readibility.)}
    \label{fig:originalscore}
\end{figure*}
At a high level, for every iteration, the method finds the configuration with the lowest margin and increases its weight by $ \delta = (\mathbf{y}_{ic}-\mathbf{H}_{ic})/\mathbf{K}_{ii} $, which consequently forces its margin to be positive one (line 5-12). Also, it tries to set the weight of the configuration with the largest positive margin to 0 if its margin remains positive after this operation (line 13-18). This update rule helps keeping the number of support points with non-zero weights minimum. The algorithm terminates when all the margins become positive and the weight of the largest-margin point can not be set to zero, or when it reaches the maximum number of iterations. 

After training, all redundant samples, i.e., samples with all-zero weights $\mathbf{W}_i = \mathbf{0} \in \mathbb{R}^{C}$, will be removed from the dataset to achieve sparsity. The remaining set of configurations are the \textit{support configurations}. In practice, the size of the support set $|\mathbf{S}|$ is usually smaller than $|\mathbf{X}|$ by 1-2 orders of magnitude \cite{das2020learning}, which is how the proxy detectors can maintain low computational complexity.


Algorithm \ref{algo:sparse} can be used for both initial training and, notably, for active learning updates to account for changing environments, which will be described in Sec.~\ref{sec:active}. When it is used for initial training, the weights should be initialized all as $\mathbf{0}$, and the hypothesis $\mathbf{H}$ should also be $\mathbf{0}$. The entries of $\mathbf{K}$ are not calculated until necessary. The function $\textit{kernelNotComputed}(\mathbf{x}_i)$ checks if the kernel values of $\mathbf{x}_i$ have been calculated. The entries of $\mathbf{x}_i$ are only calculated when it reaches the line \ref{algline:checkkernel} and $\textit{kernelNotComputed}(\mathbf{x}_i)$ returns true.


In summary, the above multi-label kernel perceptron approach is able to find a set of support configurations for each category of obstacles. It can be viewed as a more general version of the Fastron algorithm described in \cite{das2020learning}. While the model built by \cite{das2020learning} can only output binary predictions of \textit{in-collision} or \textit{collision-free}, our method is able to predict a composite collision status that identifies which category of obstacles the robot collides with.

\subsection{A Differentiable Model for Collision Detection}\label{sec:diff}
While proxy collision detectors typically use  Gaussian or Rational Quadratic kernels to map the hypothesis function in Eq.~\eqref{eq:hypothesis}, when visualized across the configuration space, the hypothesis function has non-intuitive peaks and valleys. This means that, if taking the gradient of the hypothesis score map, the direction of motion may not always move the configurations away from collision but may instead move them into collision.
See Fig.~\ref{fig:originalscore} for a visual illustration. This is because these popular kernels have the nature of outputting near-zero values for configurations that are far from the support configurations. This causes the kernel perceptron to value the regions in the centers of free and obstacle regions to be close to zero, which intuitively should be larger values than their surroundings. 
Thus, we propose the use of an \textit{unbounded} kernel function when computing the collision score, the polyharmonic radial basis function $k_{\mathrm{PH}}$. The polyharmonic function is defined as
\begin{align}
    k_{\mathrm{PH}}(\cfg, \cfg') &=
    \begin{cases}
        r^k &\text{when }k=1,3,5,...,\\
        r^k\ln(r) &\text{when }k=2,4,6,...,
    \end{cases}\\
    \text{where }r &= \|\cfg-\cfg'\|_2,
\end{align}
and is used as an alternative to the Gaussian or Rational Quadratic Kernel used in proxy collision detection \cite{das2020learning,das2020forward}. The output, shown in Fig. \ref{fig:originalscore}, produces gradients that allow sampled configurations to follow an efficient path away from collision.

The resulting kernel matrix on the support set $\mathbf{S}$ is denoted as $\mathbf{K}_\mathrm{PH} \in \mathbb{R}^{M\times M}$, 
\begin{equation}
    \begin{aligned}
        \mathbf{K}_{\mathrm{PH,}ij} = \begin{cases}
            k_\mathrm{PH}(\mathbf{x}_i, \mathbf{x}_j) & \text{if } \exists c \in C, \mathbf{W}_{ic}\mathbf{W}_{jc}\neq 0, \\
            0 & \text{otherwise}.
        \end{cases}
    \end{aligned}
\end{equation}
Unless otherwise stated, we use $k=1$ for experiments in this paper; adjusting values of $k$ changes the drop-off rate of the kernel function as the test point moves further away from the support set. The interpolation objective is
to find a set of interpolation weights $\mathbf{A}\in \mathbb{R}^{M\times C}$ so that
\begin{align}
    \mathbf{K}_\mathrm{PH}\mathbf{A} = \supportlabel.
    \label{eq:interpolation}
\end{align}
This linear equation can be solved by off-the-shelf software packages. Even though the original formulation of polyharmonic spline involves an additional set of weights for linear terms \cite{hangelbroek2013density},
no practical improvements were observed when using it, so we eliminated it to keep the proposed algorithm in its simplest form.

With the usage of forward kinematics transform, $\FK(\cfg)$ in proxy collision checking \cite{das2020forward}, we find it beneficial to input the control point positions instead of raw configurations to $k_\mathrm{PH}$, leading to a new interpolation kernel function
\begin{align}
    k_\FKPH(\cfg, \cfg') = k_{\mathrm{PH}}(\FK(\cfg), \FK(\cfg')).
\end{align}
The new interpolation kernel function $k_\FKPH$ differs from the kernel function $k_{\FK}$ used in kernel perceptron. 
Naturally, $\mathbf{K}_\FKPH \in \mathbb{R}^{M\times M}$ refers to the corresponding kernel matrix generated by $k_\FKPH(\cdot, \cdot)$ on the support set. Now, given a query configuration $\cfg_q$, we can calculate its collision score by a function $\cscore: \mathbb{R}^{D} \rightarrow \mathbb{R}^{C}$
\begin{align}
    \cscore(\cfg_q) = k_\FKPH(\cfg_q, \mathbf{S})\mathbf{A}, \label{eq:score}
\end{align}
where $k_\FKPH(\cdot, \mathbf{S}): \mathbb{R}^{D} \rightarrow \mathbb{R}^{1\times M}$ is a function that calculates kernel values of a query configuration $\cfg_q$ against all support points using the kernel function $k_\FKPH$. $\cscore(\cfg_q)$ produces a $C$-dimensional vector. Its $c$-th entry indicates the collision score between the robot and obstacles in category $c$. 
In-collision configurations receive \textbf{positive} scores for the categories they collide with, while collision-free configurations get \textbf{negative} values.

It's important to note the actual numerical values of the collision scores do not approximate the geometrical distances because the interpolation target is just $\supportlabel$, the binary collision checking labels of $+1$ and $-1$. This model output does not represent any physical world measurements, but rather describes the distance to margin in the kernel space.
Its gradients are still useful in informing an efficient path to move out of and away from obstacles.
Using binary geometrical collision checking labels to set up interpolation targets for the differentiable model has reduced overhead compared to computing exact distances. 
We will later show by experiments that this collision score reasonably correlates the exact collision distance, especially for configurations at the boundary. 

Finally, of note, the collision score function $\cscore$ is analytically differentiable, whose gradients w.r.t. the query configuration
$\nabla_{\cfg_q}\cscore_c,\ \forall c \in \{1,2,...,C\}$
can be conveniently calculated by any automatic differentiation library via backpropagation. We use PyTorch \cite{paszke2019pytorch} to implement the proposed algorithm for easy access to the queried configuration's gradients.

\begin{algorithm}[tbp]
\DontPrintSemicolon 
\KwIn{For every loop: previous support configurations $\mathbf{S}$, perceptron weights $\mathbf{W}$ and hypothesis $\mathbf{H}$.}
\KwOut{For every loop: updated support configurations $\mathbf{S}$, perceptron weights $\mathbf{W}$ and hypothesis $\mathbf{H}$, and interpolation weights $\mathbf{A}$.}
\While{true}{
    $\Cfg'\leftarrow \emptyset$\;
    \tcp{exploitation}
    \For{$i\leftarrow 1\ to\ \nu$}{
        \ForEach{$\cfg \in \mathbf{S}$}{
            $\Cfg' \leftarrow \Cfg'\cup\{\cfg'\sim \mathcal{N}(\cfg, \sigma^2I)\}$
        }
    }
    \tcp{exploration}
    \For{$i \leftarrow\ 1\ to\ \zeta$}{
        $\Cfg' \leftarrow \Cfg'\cup\{\cfg'\sim U(\theta_{\mathrm{min}}, \theta_{\mathrm{max}})\}$
    }
    \tcp{ground truth query}
    $\mathbf{Y} \leftarrow \text{\sc trueChecker}(\mathbf{S}\cup \Cfg')$\;
    $\mathbf{S}, \supportlabel \leftarrow \text{\sc SparseMultilabelPerceptron}( \mathbf{S}\cup \Cfg',~\mathbf{Y},~\mathbf{W}\cup\{0\}, \mathbf{H}\cup\{0\})$\;
    $\mathbf{K}_{\FKPH}\leftarrow k_{\FKPH}(\mathbf{S}, \mathbf{S})$\;
    $\mathbf{A} \leftarrow \text{\sc Solve}(\mathbf{K}_{\FKPH}\mathbf{A}=\supportlabel)$\;
}
\caption{{\sc Active Learning Pipeline}}
\label{algo:active}
\end{algorithm}
\subsection{Adaptation to Dynamical Environment via Active Learning} \label{sec:active} 
A motion planning or trajectory optimization algorithm aiming at real-world applications should employ a collision detector that adapts to dynamical environments in which obstacles are subject to move. This especially raises a challenge to any collision detection methods that require pre-computation and cannot update their modeling of the environment efficiently. For example, pre-computed cost maps usually need to be fully re-computed when changes happen in the workspace \cite{pan2015efficient, chomp}. An efficiently updatable (adaptive), differentiable, low-complexity collision detection model is thus highly desirable in these situations.

We employ an efficient active learning strategy as described in Algorithm~\ref{algo:active} that discusses an update law to be executed at each new timestep. \second{This algorithm can be considered as a generalized version of the update law described in \cite{das2020learning} for a multi-class differentiable model.}

The idea is to efficiently sample random points in the configuration space so that these samples will reveal the movements of obstacles, find new obstacles in the scene, or remove disappeared obstacles. 
The sampling procedure is composed of two parts. 
The first is \textit{exploitation}. 
This is done by sampling $\nu$ times from a gaussian distribution around each of the most recent support points
to account for minor boundary changes, 
\begin{align}
\cfg' \sim \mathcal{N}(\cfg,\sigma^2I),\quad \forall \cfg\in \mathbf{S}.
\end{align}
\updated{This stage is under the assumption, and an empirically true result, that almost all support configurations in $\mathbf{S}$ are very close to the boundary of $\mathcal{C}_\mathrm{free}$ and $\mathcal{C}_\mathrm{obs}$. Since this boundary changes gradually when the objects do not move too fast, just by sampling around the previous support points, many samples will be located near the boundary of the current timestep. In the model update step, adjusting the model to fit these critical samples will largely maintain its accuracy. The values of $\nu$ and $\sigma$ should be selected to accommodate the speed of the objects: if the objects are moving fast, a large $\sigma$ should be selected, otherwise the model may fail to track the C-space boundary, and vice versa. It never hurts the model accuracy to use a large $\nu$, but it increases the time consumed on getting ground-truth labels and updating the model. }

The second stage is \textit{exploration}, which tries to discover new obstacles that previously do not exist in the environment, given a certain time or sample limit. This is done using a random sampler in the configuration space, 
\begin{equation}
\cfg'\sim U(\theta_{\mathrm{min}},\theta_{\mathrm{max}})
\end{equation}
for $\zeta$ samples\second{, where $\theta_{\mathrm{min}}$ and $\theta_{\mathrm{min}}$ are joint limits.}. \updated{Similar to $\nu$, it is always good for model accuracy to use a large $\zeta$, but it increases the computational overhead of model updating. A good trade-off between accuracy and performance can be achieved by adjusting $\nu, \zeta$.}



\subsection{Trajectory Optimization with Variable Safety Bias}\label{sec:traj} 
The proposed multi-label differentiable collision detector can be applied to many scenarios and algorithmic pipelines, especially those where the robot must be aware of different classes of obstacles and appropriately optimize their plans to safely avoid them in a risk-aware manner. This involves iteratively solving for the trajectory at every control step, as common practice for model-predictive control and rapid replanning. 

To formulate the trajectory optimization problem that minimizes the cost of end effector movement while keeping collision-free with different categories of obstacles, we have:
\begin{mini}|s|
    {\cfg_{1:T}}{f(\cfg)}{}{}
    \addConstraint{\cscore_c(\cfg_t)+\epsilon_c}{\leq 0 \ \ \forall c\in\{1,...,C\}, 
    \forall t} 
    \addConstraint{\|\cfg_{t+1}-\cfg_t\|}{<\omega_{\mathrm{max}} \quad t=1...T-1}
    \addConstraint{\|\mathrm{ee}(\cfg_{t+1})-}{\mathrm{ee}(\cfg_t)\|<v_{\mathrm{max}} \quad t=1...T-1}
    \addConstraint{\theta_{\mathrm{min}} \leq \cfg_t}{\leq \theta_\mathrm{max} \quad t=1...T}
    \addConstraint{\cfg_1}{=\cfg_{\mathrm{start}}}
    \addConstraint{\cfg_T}{=\cfg_{\mathrm{goal}}}.
\end{mini}
$T$ is the total number of waypoints (or timesteps). 
$\cscore_c(\cdot)$ again is the proxy collision score defined in kernel space (Eq.~\eqref{eq:score}); 
$\epsilon_{c}$'s are categorical safety biases that try to reduce false-negatives given by \shortname. $\epsilon_{c}$'s should be positive $\forall c\in \{1,...,C\}$, and $\epsilon_{c_1}$ should be larger than $\epsilon_{c_2}$ if objects of category $c_1$ are more important than objects of $c_2$. 
$\omega_{\mathrm{max}}$ limits the total change of joint angles between adjacent waypoints; $\theta_{\mathrm{min}}$ and $\theta_{\mathrm{max}}$ are limits of the joint angles.
$\mathrm{ee}(\cdot)$ represents the position of the end-effector, and $v_{\mathrm{max}}$ sets the maximum velocity for the end effector. 
The objective $f(\cfg)$ can be arbitrarily defined by any differentiable function. We define it as
\begin{equation}
    \begin{aligned}
    f(\cfg) = \sum_{t=1}^{T-1} & w_{p}\|\FK(\cfg_{t+1})-\FK(\cfg_{t})\|^2 \\
    + &w_\theta\|\cfg_{t+1}-\cfg_{t}\|^2
    \end{aligned}
\end{equation}
where $\FK(\cdot)$ produces the Cartesian coordinates of the robot control points on each link, allowing us to have a weighted objective to minimize a combination of Cartesian movement and joint movement. 

Problems in this formulation have been well studied and can be solved by off-the-shelf optimization software \cite{adam, paszke2019pytorch, chomp, trajopt}. Optimization algorithms have their distinct ways of handling constraints. As an example, when we use the Adam algorithm implemented in PyTorch library, 
the constraints are converted into weighted terms to add to the original objective $f(\cfg)$, 
\begin{align}
    h(\cfg) = f(\cfg) + \sum_i \mu_i|g_{\leq}^i(\cfg)|^+ + \sum_j \beta_j |g_{=}^j(\cfg)|,
\end{align}
where $|\cdot|^+ = \max(\cdot, 0)$ and $|\cdot|$ takes the absolute value. $g_{\leq}^i$'s are for inequality constraints and $g_{=}^j$'s are for equality ones. The constraints are converted to $g(\cfg)$'s simply by moving all non-zero terms of the equality/inequality to the left-hand side. The new optimization problem $\min_\cfg h(\cfg)$ is unconstrained.

This algorithm works for either admissible or inadmissible initializations, i.e., it will try to find a locally optimal admissible trajectory even starting from an inadmissible initial guess. In practice, we use both an initial seed of a straight line in the configuration space between the start and the end configurations, as well as random seeds. Verification of the admissibility of the generated output is performed for each instance.

\begin{figure}[tbp]
    \centering
    \includegraphics[width=0.3\linewidth]{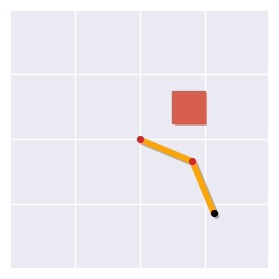}
    \includegraphics[width=0.3\linewidth]{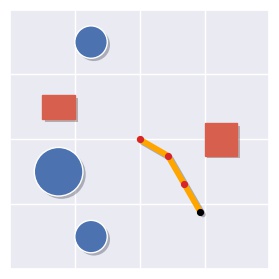}
    \includegraphics[width=0.3\linewidth]{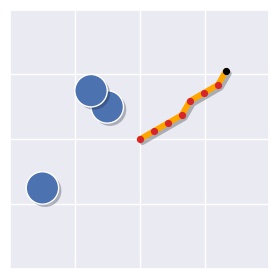}
    \includegraphics[width=0.35\linewidth]{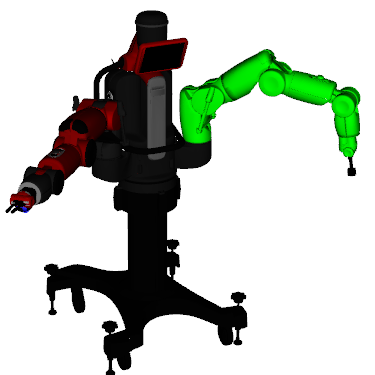}
    \includegraphics[width=0.35\linewidth]{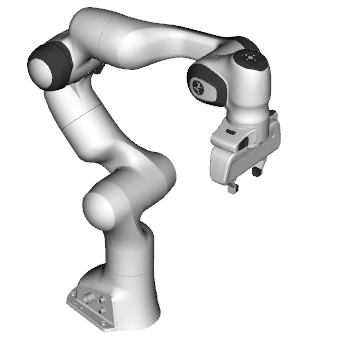}
    \caption{2, 3, 7 DOF planar robots, Baxter robot and Panda robotic arm involved in experiments. While the planar robots may be simple in Cartesian space, their configuration spaces can be of the same complexities as the 3D robots of the same DOF. For the Baxter robot, only its 7-DOF left arm is considered by all experiments unless otherwise specified. For the Panda arm, only its 7-DOF arm is considered.} 
    \label{fig:robot_gallery}
\end{figure}

\section{Experiments and Results}
In this section, we first evaluate the score given by 
\shortname both quantitatively and qualitatively. Then we show how to do trajectory optimization with safety constraints efficiently using \shortname. Next, we show the effectiveness of the active learning strategy in a dynamical environment. Finally, we provide a series of comprehensive quantitative comparisons between \shortname and the standard geometrical collision checking algorithm FCL \cite{pan2012fcl} in trajectory optimization tasks. The experimented robots are in Fig.~\ref{fig:robot_gallery}. \second{For each robot, we select the control points to be unique joint positions, including the tip of the end effector. This resulted in 2, 3, 7 control points for the planar robots, respectively, and 4 for the Baxter left arm and the Panda arm.}

\updated{For all experiments, \shortname is implemented using PyTorch \cite{paszke2019pytorch}, which is a set of Python bindings of the underlying C++ source code. FCL is also deployed using Python bindings of the C++ library \second{(stable version v0.6.1) \cite{pan2012fcl, pythonfcl}}. 
\second{For training \shortname, the binary checking functionality of FCL is used, while continuous-valued distance to collision is queried when FCL is used as a baseline collision detector in evaluation.}
The optimization algorithms are from off-the-shelf Python packages. Namely, Adam is from PyTorch; SLSQP and Trusted-Region Constrained are from SciPy \cite{trustregion, 2020SciPy-NMeth}. All experiments are done on a workstation with 32GB memory and an Intel Xeon Silver 4214 CPU, which has 24 cores@2.20GHz.} Our code is available at \href{https://github.com/ucsdarclab/diffco}{https://github.com/ucsdarclab/diffco}.

\begin{figure}[tbp!]
    \centering
    \includegraphics[width=\linewidth,  trim=62 10 49 20, clip]{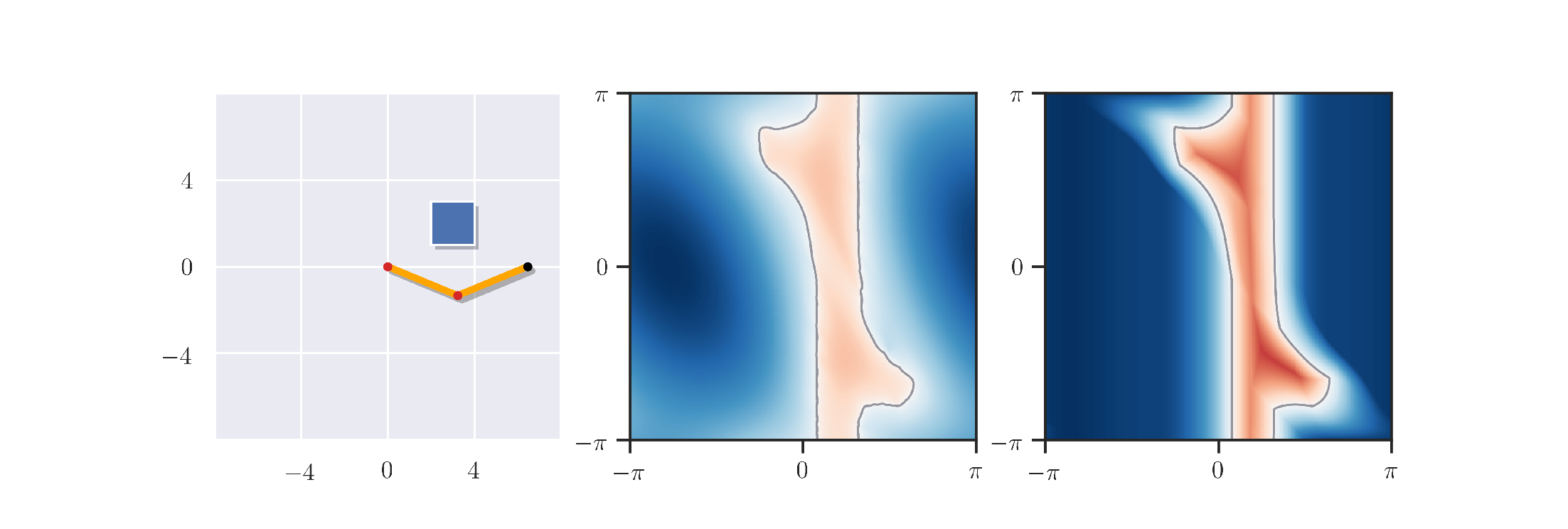}
    \centering
    \includegraphics[width=\linewidth,  trim=62 10 49 20, clip]{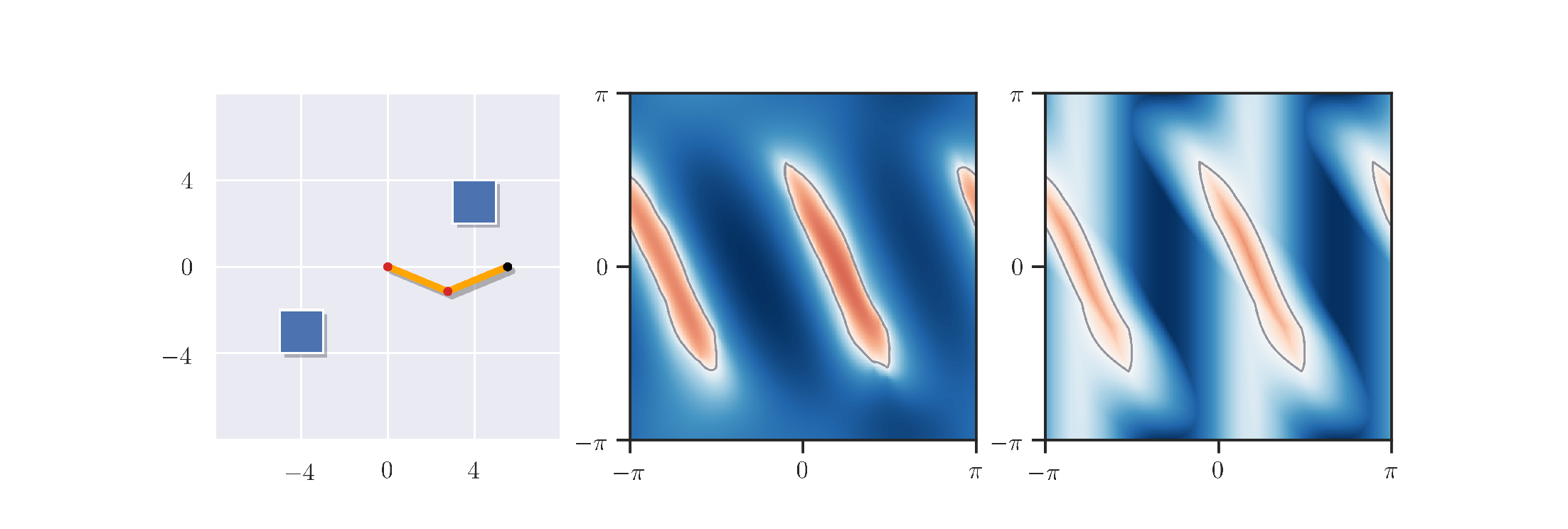}
    \centering
    \includegraphics[width=\linewidth,  trim=62 10 49 20, clip]{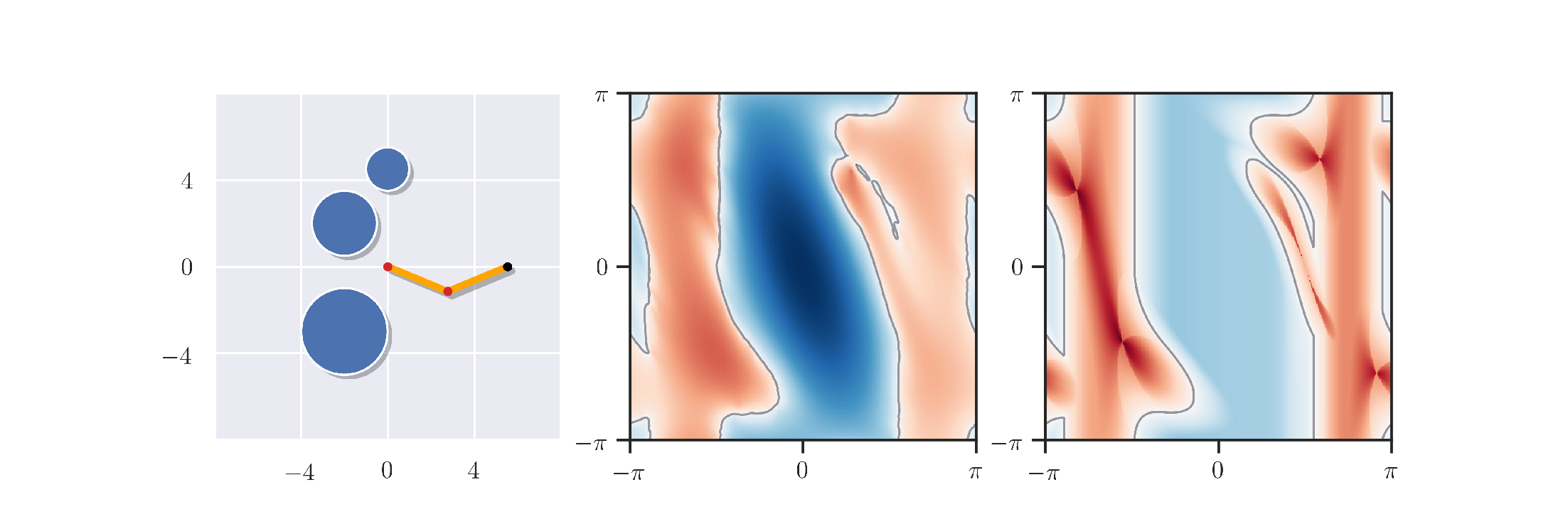}
    \caption{\shortname score (middle) for a 2-DOF robot (left) in different settings, compared to the result given by FCL (right). They are visually similar in terms of trends and shapes of the value surface. 
    In-collision regions are colored orange while the remaining areas are collision-free.}
    \label{fig:vis}
\end{figure}

\begin{figure}[tbp]
    \centering
    \begin{subfigure}{0.48\linewidth}
    \includegraphics[width=\linewidth, trim={7 15 26 34},clip=true]{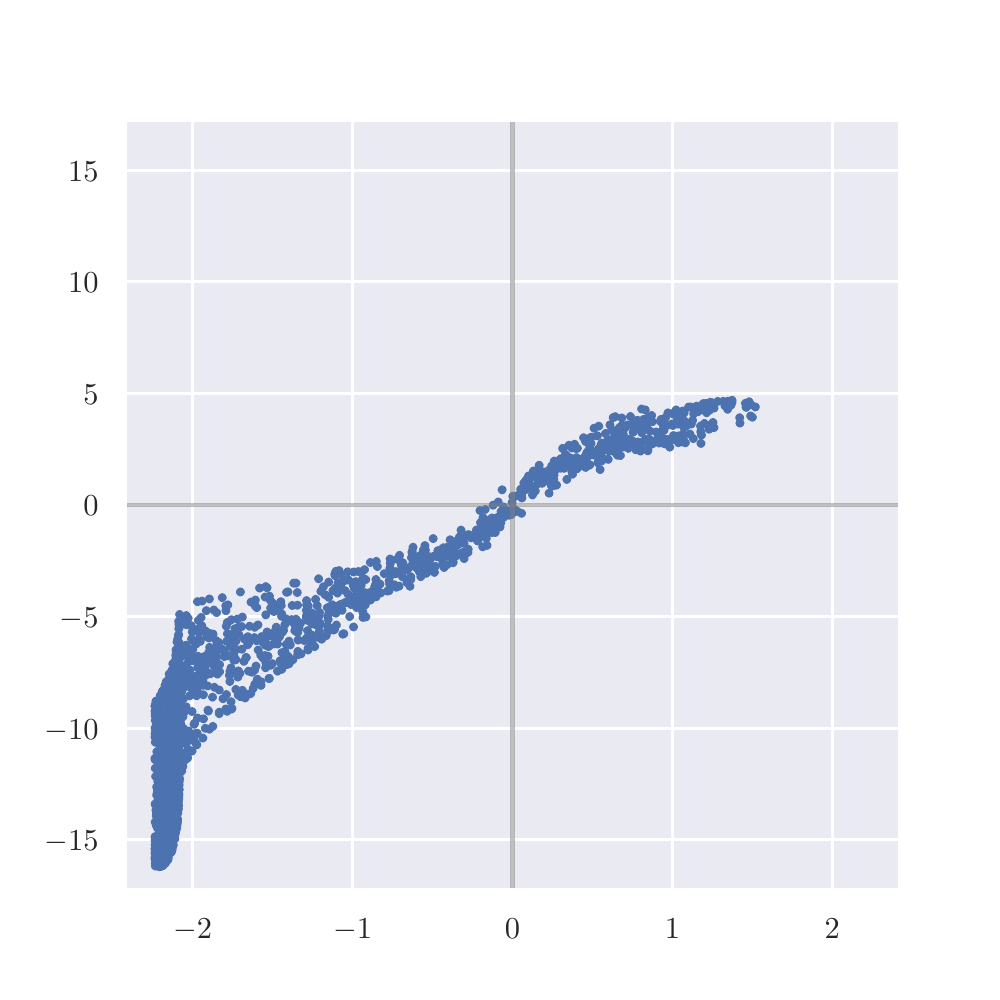}
    \caption{2 DOF, $1^\text{st}$ scene}
    \end{subfigure}
    \begin{subfigure}{0.48\linewidth}
    \includegraphics[width=\linewidth, trim={7 15 26 34},clip=true]{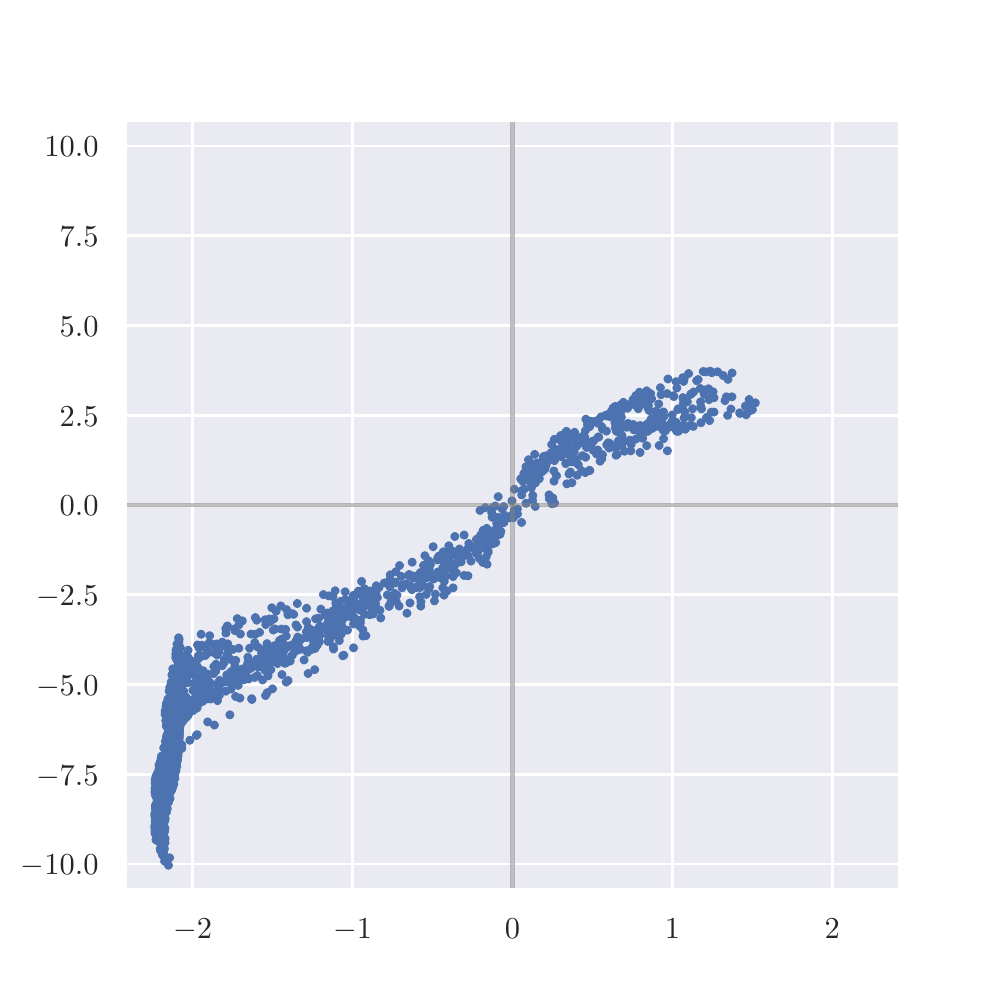}
    \caption{2 DOF, $1^\text{st}$ scene, w/o FK}
    \end{subfigure}
    
    \begin{subfigure}{0.48\linewidth}
    \includegraphics[width=\linewidth, trim={7 15 26 34},clip=true]{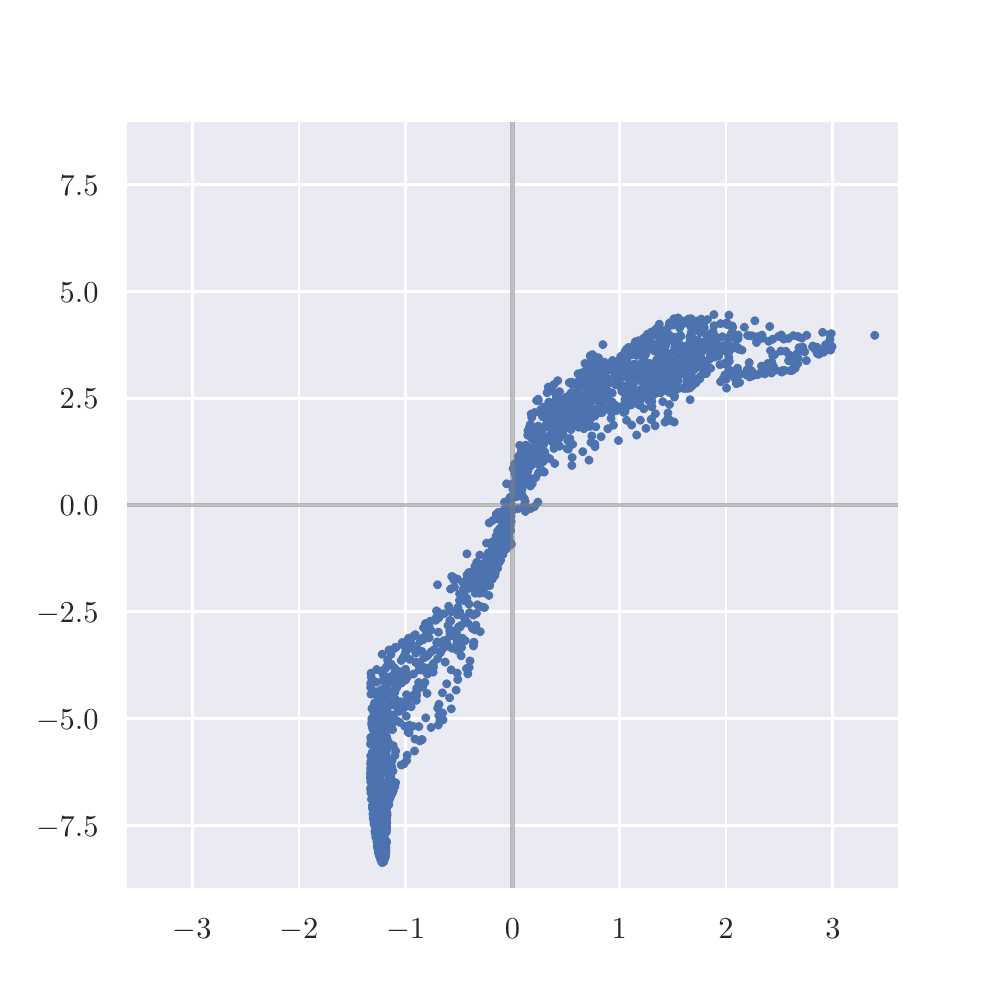}
    \caption{2 DOF, $2^\text{nd}$ scene}
    \end{subfigure}
    \begin{subfigure}{0.48\linewidth}
    \includegraphics[width=\linewidth, trim={7 15 26 34},clip=true]{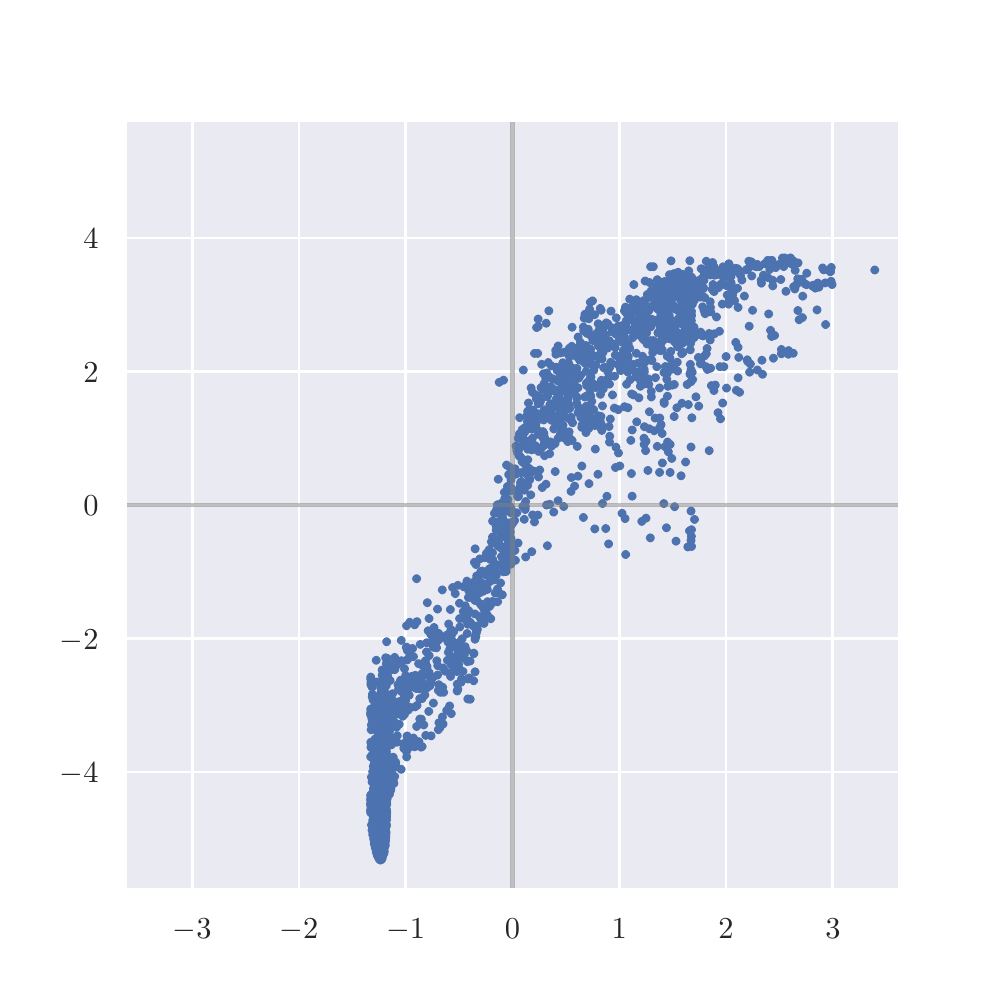}
    \caption{2 DOF, $2^\text{nd}$ scene, w/o FK}
    \end{subfigure}
    
    \begin{subfigure}{0.48\linewidth}
    \includegraphics[width=\linewidth, trim={7 15 26 34},clip=true]{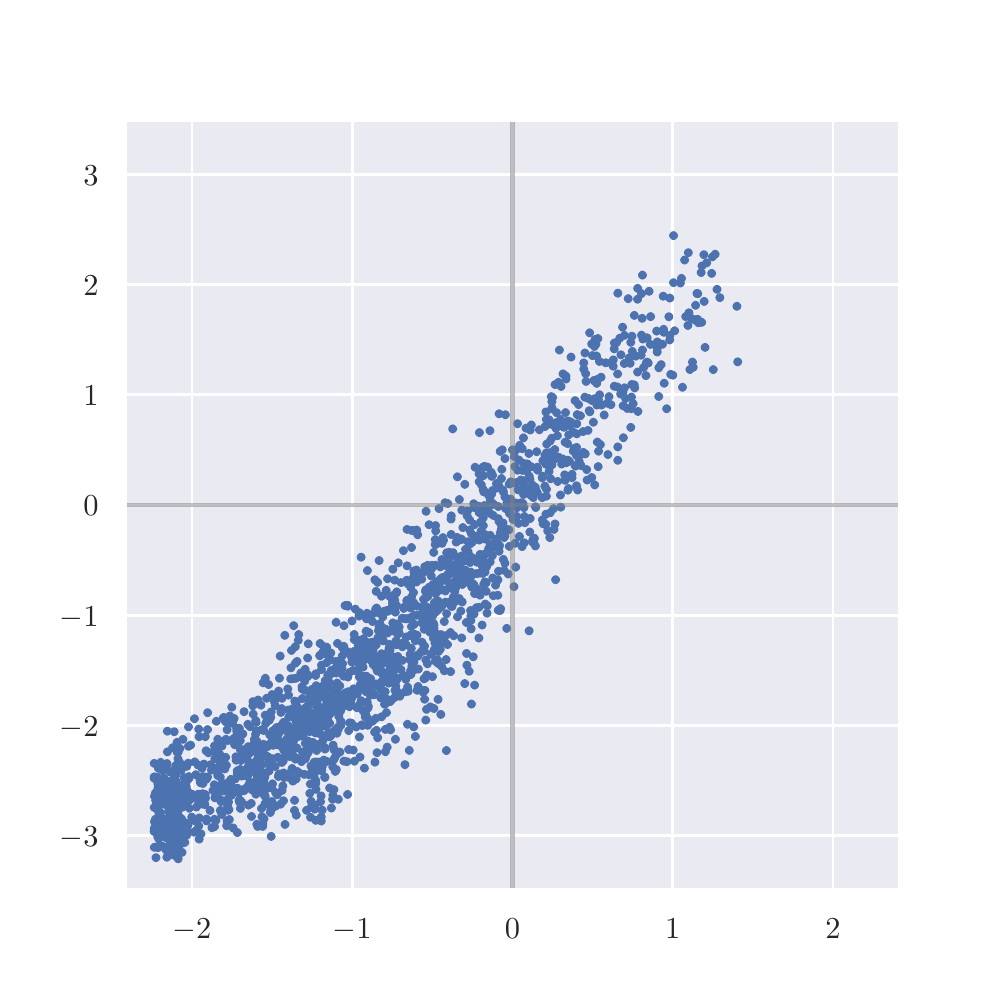} 
    \caption{7 DOF, $1^\text{st}$ scene}
    \end{subfigure}
    \begin{subfigure}{0.48\linewidth}
    \includegraphics[width=\linewidth, trim={7 15 26 34},clip=true]{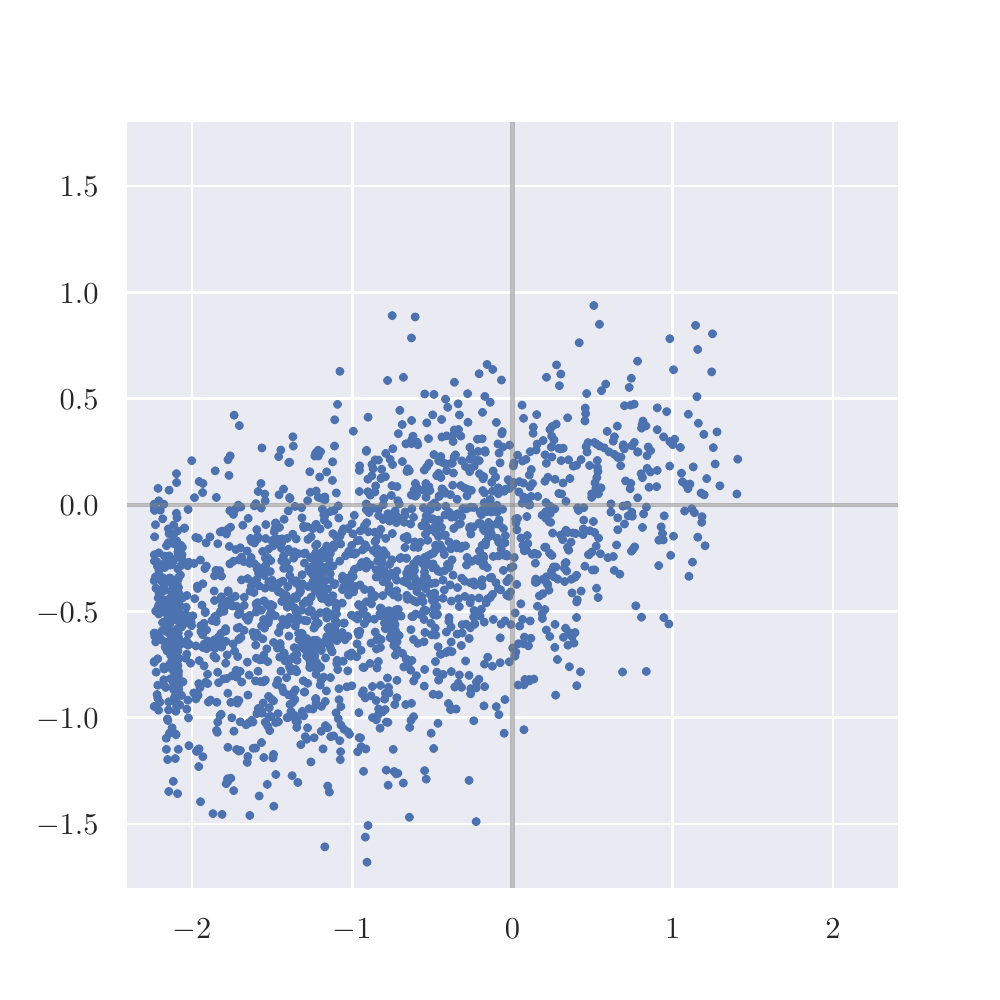}
    \caption{7 DOF, $1^\text{st}$ scene, w/o FK}
    \end{subfigure}
    
    \begin{subfigure}{0.48\linewidth}
    \includegraphics[width=\linewidth, trim={7 15 26 34},clip=true]{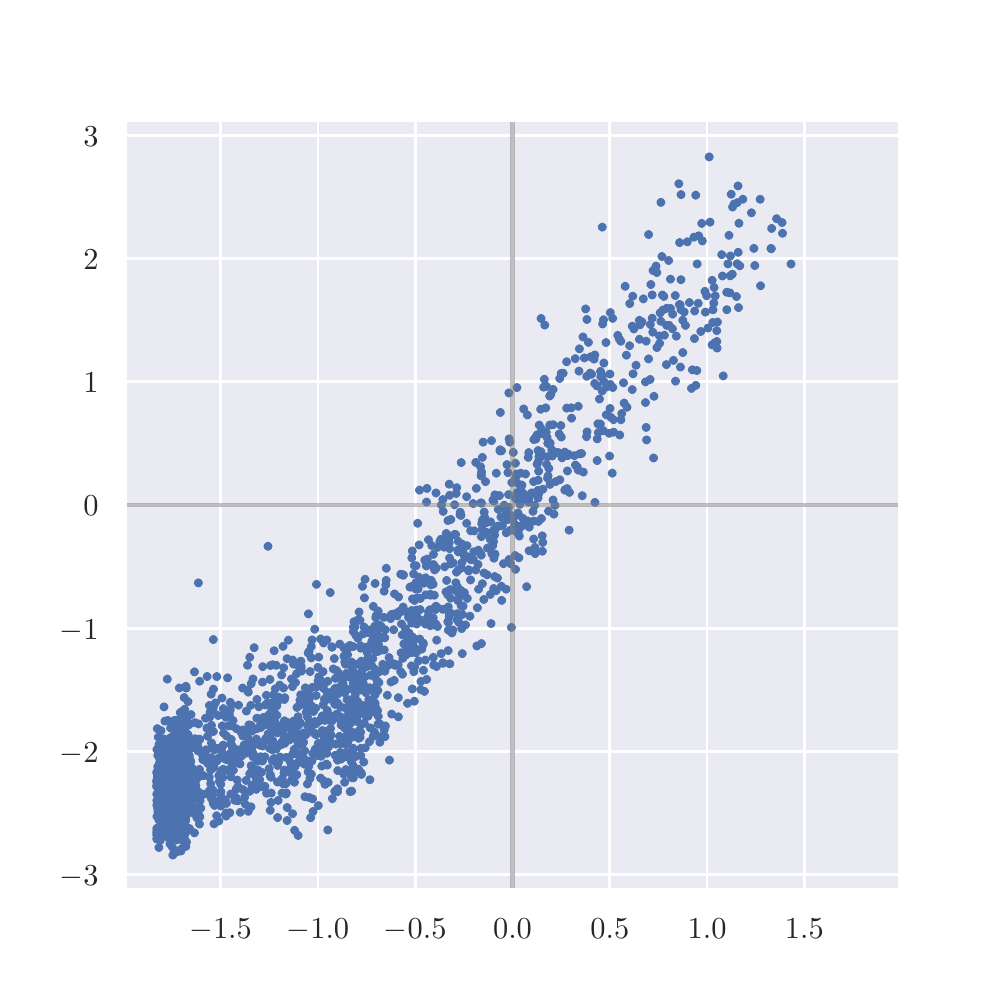}
    \caption{7 DOF, $2^\text{nd}$ scene}
    \end{subfigure}
    \begin{subfigure}{0.48\linewidth}
    \includegraphics[width=\linewidth, trim={7 15 26 34},clip=true]{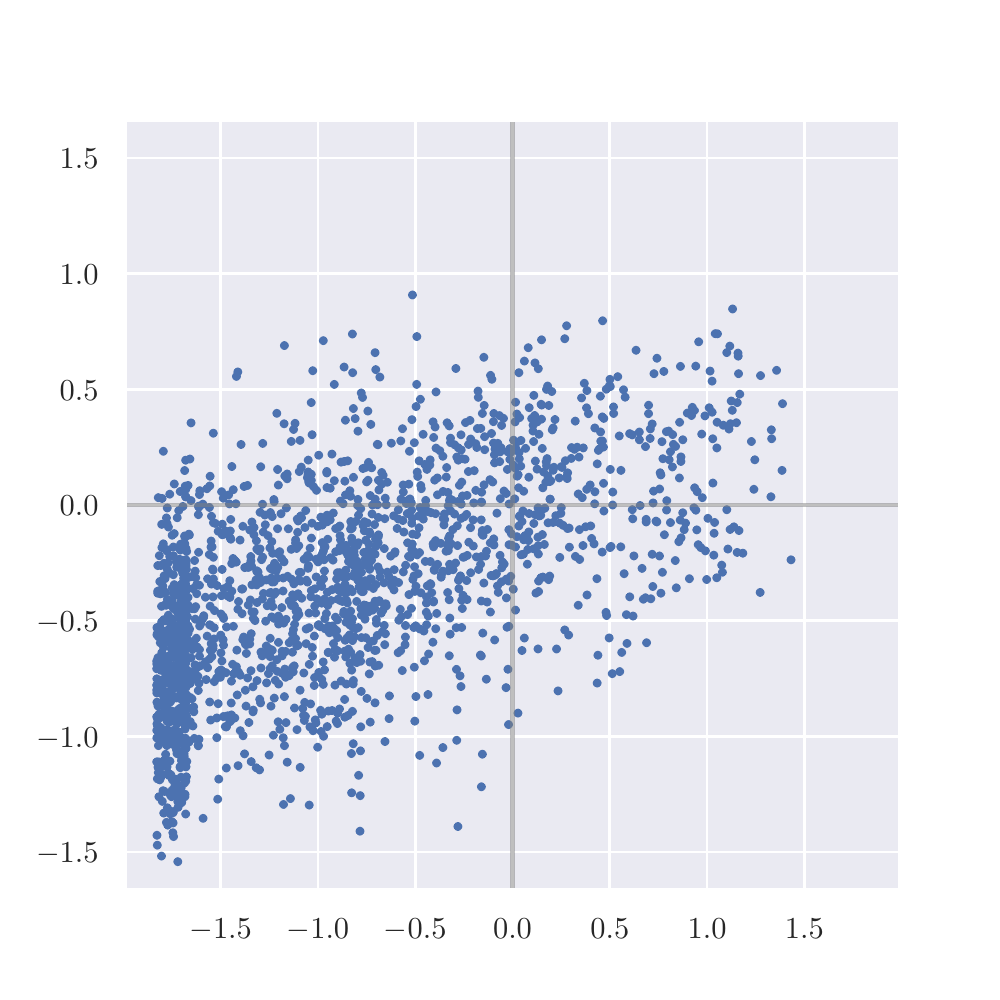}
    \caption{7 DOF, $2^\text{nd}$ scene, w/o FK}
    \end{subfigure}
    
    \caption{Correlation between \shortname collision score ($\mathbf{y}$-axis) and results given by FCL ($\mathbf{x}$-axis) for a 2DOF and 7DOF arm. (a-d) shows the 2DOF results, (e-h) shows the 7DOF results, and (b,d,f,h) shows the results when not integrating the FK kernel into the proxy collision detector. We show that good correlation can be seen between the score and the geometrical distance to collision, and (b,d,f,h) shows how incorporating the robot kinematics when doing a proxy collision detection is important to getting good correlation.
    } 
    \label{fig:correlation}
\end{figure}

\begin{figure}[tbp]
    \centering
    \begin{subfigure}{0.48\linewidth}
    \includegraphics[width=\linewidth, trim={7 15 26 34},clip=true]{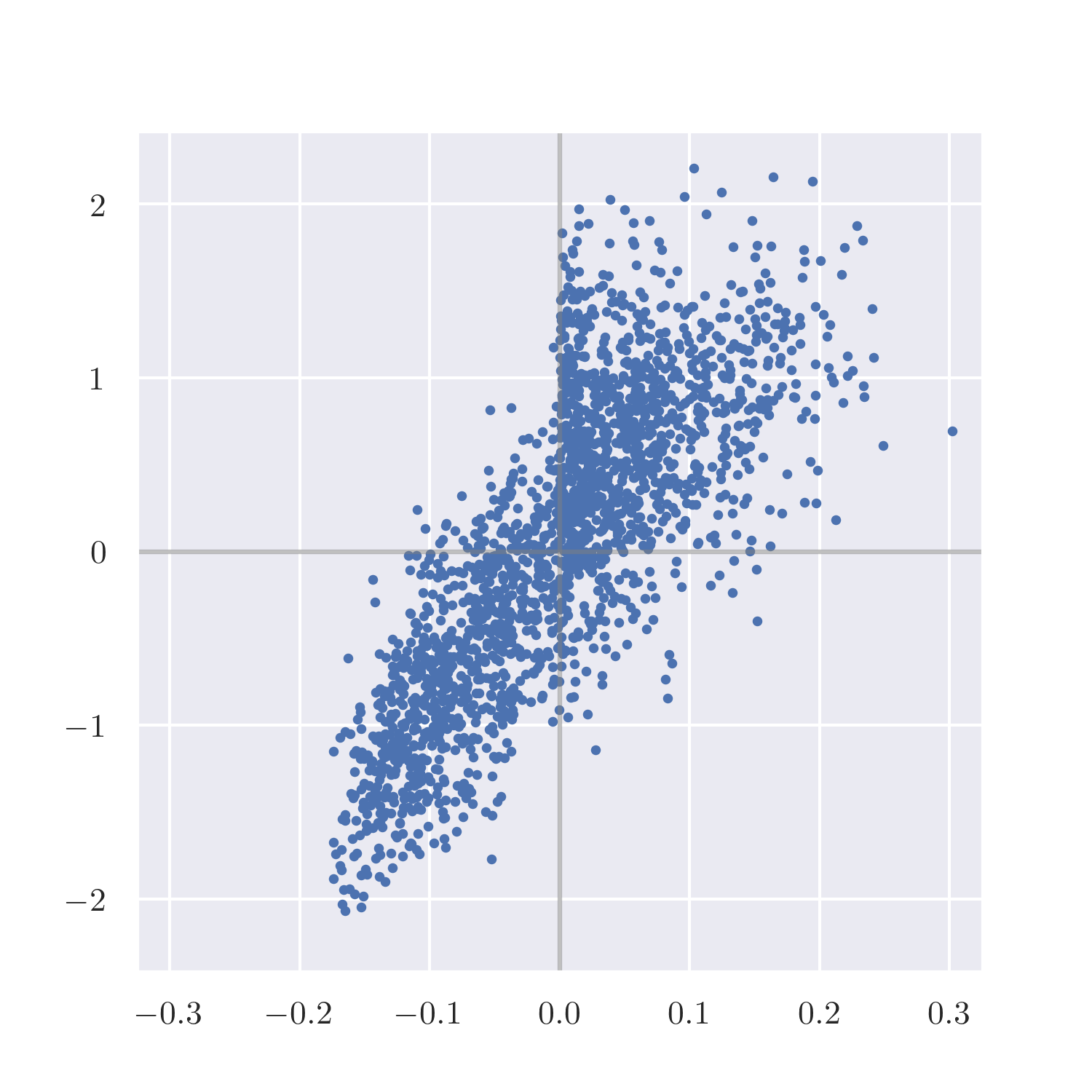}
    \caption{with FK transform}
    \end{subfigure}
    \begin{subfigure}{0.48\linewidth}
    \includegraphics[width=\linewidth, trim={7 15 26 34},clip=true]{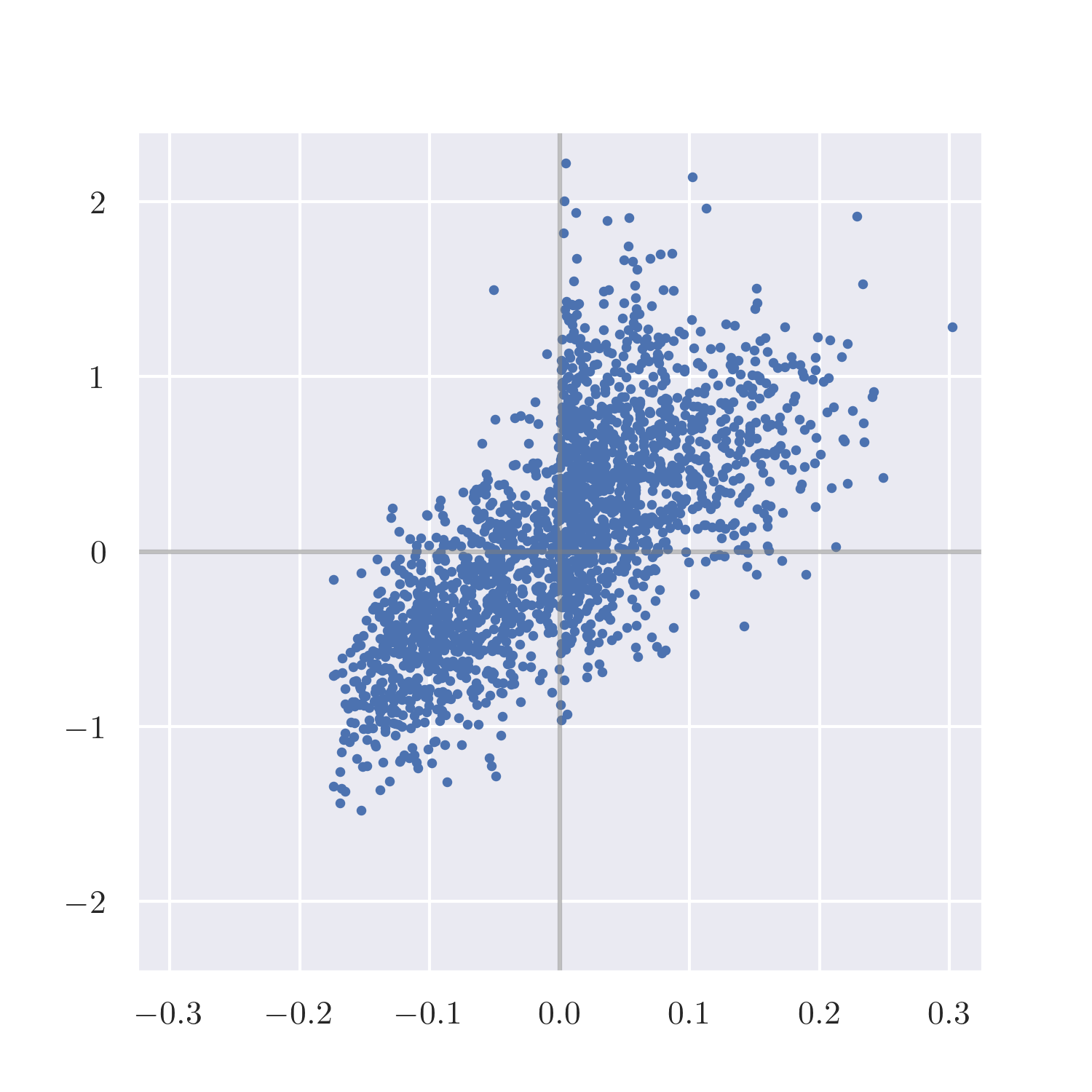}
    \caption{w/o FK transform}
    \end{subfigure}
    
    \caption{Correlation between \shortname collision score ($\mathbf{y}$-axis) and results given by FCL ($\mathbf{x}$-axis) for self-collision of a 14DOF dual-arm Baxter robot. 
    } 
    \label{fig:selfcollisioncorrelation}
\end{figure}

\subsection{Evaluation of \shortname Score}

Fig.~\ref{fig:vis} shows the similarity between \shortname collision score and the geometrical signed distance to collision given by FCL \cite{pan2012fcl}. \updated{We formulate the distance to collision similar to the way shared by many motion planning algorithms \cite{chomp, Merkt2018, das2020stochastic, trajopt, uber2020perceive}:
\begin{align}
    d_{\mathcal{R}}(\cfg) =
    \begin{cases}
        -\min\limits_{i, j} {distance}(\mathcal{R}_i(\cfg), o_j) & \cfg\in \mathcal{C}_\mathrm{free},\\
        \max\limits_{i, j} {penetration}(\mathcal{R}_i(\cfg), o_j) & \cfg\in \mathcal{C}_\mathrm{obs},
    \end{cases}
    \label{eq:collisiondistance}
\end{align}
\noindent where $\cfg$ is the robot configuration being tested. 
For collision-free configurations, the signed distance to collision is the negative value of the shortest distance between points of all robot links $\mathcal{R}_i$'s and points of all obstacles $o_j$'s. For in-collision configurations, it's defined to be the greatest penetration depth (the minimum amount intersecting robot links must move to be out of collision \cite{collisionbook}).} 
So, the \textbf{in-collision} configurations have \textbf{positive} distances and \textbf{collision-free} configurations have \textbf{negative} ones. In Fig.~\ref{fig:vis}, the graphs generated using the polyharmonic kernel with $\FK$ transforms ($\FKPH$ kernel) shows good correlation, especially when it gets close to the boundaries of C-space obstacles, which means the proposed differentiable collision detector can be used to compute a direction towards the outside of a C-space obstacle when the robot configuration is in collision.


\begin{figure}[tb]
    \centering
    \includegraphics[width=0.49\linewidth]{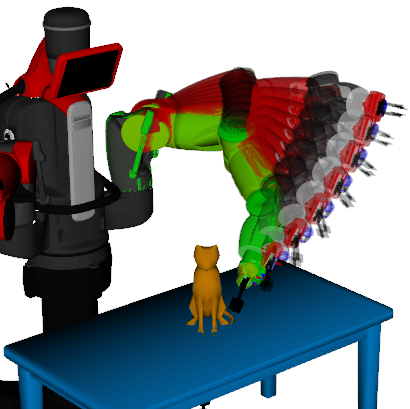} 
    \includegraphics[width=0.49\linewidth]{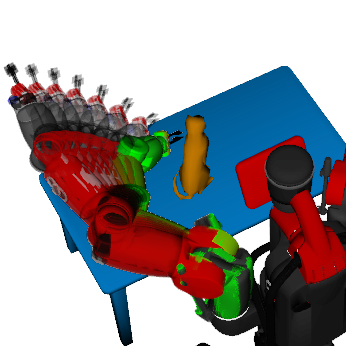}
    \caption{An example of what happens when the cat on the table is given a high weight of collision cost. The gradients of \shortname collision score are able to produce a trajectory that guides the Baxter robot to escape from high-risk configurations. The starting configuration is colored green, which is close to the cat. 
    }
    \label{fig:high-risk}
\end{figure}

In Fig.~\ref{fig:correlation}, the correlation between the collision score given by the proposed method and the geometrical signed distance given FCL \cite{pan2012fcl} is visualized. \second{A good correlation means that the learned collision score is a faithful indicator of the closeness to collision.} Here we show that using the $\mathrm{FK}$ transform provides better correlation to the FCL values than without. Most of the time, this is the case because the FK maps the joint-space configurations to a Cartesian space prior to the kernel distance evaluation, and thus is much more closely correlated to the geometrical distance-to-collision value that is usually natively measured in Cartesian space (by GJK, FCL, or other) as well. 

\updated{This formulation of distance to collision can also be generalized to self-collision by replacing $o_j$ with $\mathcal{R}_j(\cfg)$ in Eq.~\eqref{eq:collisiondistance}, and \shortname can be used to model self-collision without any technical modification. The correlation between the \shortname collision score and the true signed distance to self-collision of a 14-DOF dual-arm Baxter robot is visualized in Fig.~\ref{fig:selfcollisioncorrelation}.}

\begin{figure}[tbhp]
    \centering
    \frame{\includegraphics[width=.49\linewidth, height=.49\linewidth, trim=6.2cm 5.8cm 1.9cm 2.2cm, clip]{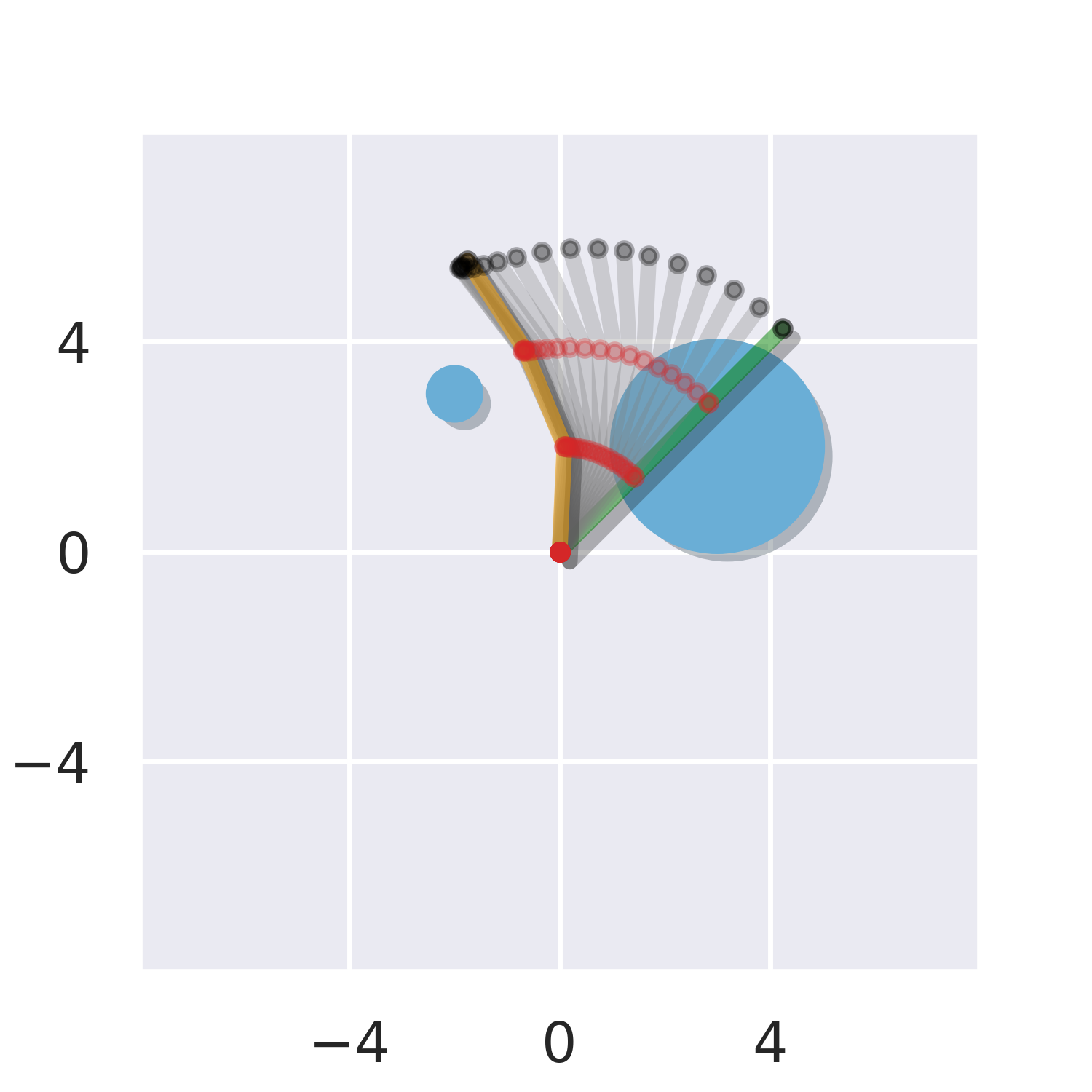}}
    \frame{\includegraphics[width=.49\linewidth, height=.49\linewidth, trim=6.2cm 5.8cm 1.9cm 2.2cm,, clip]{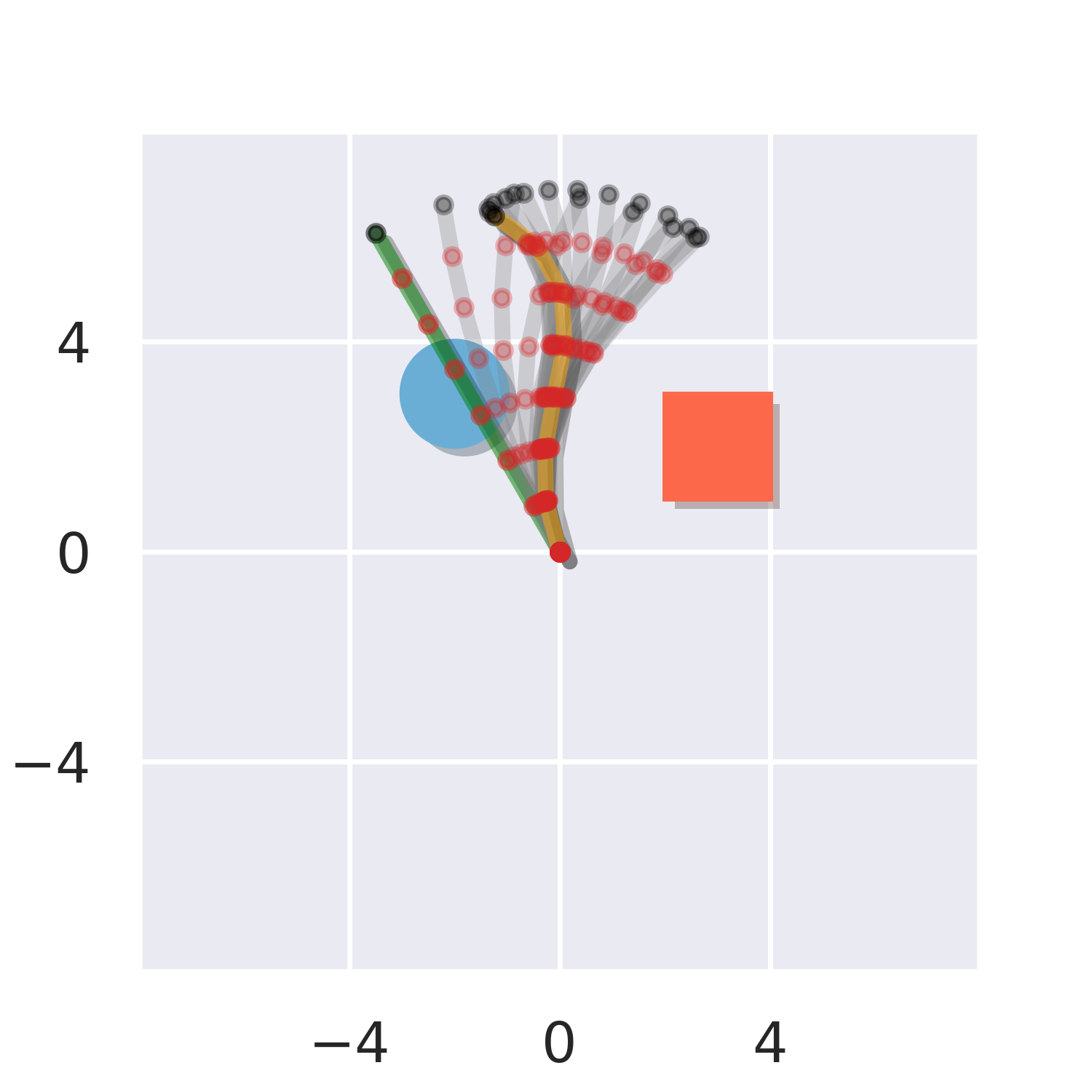}}
    \frame{\includegraphics[width=.49\linewidth, height=.49\linewidth]{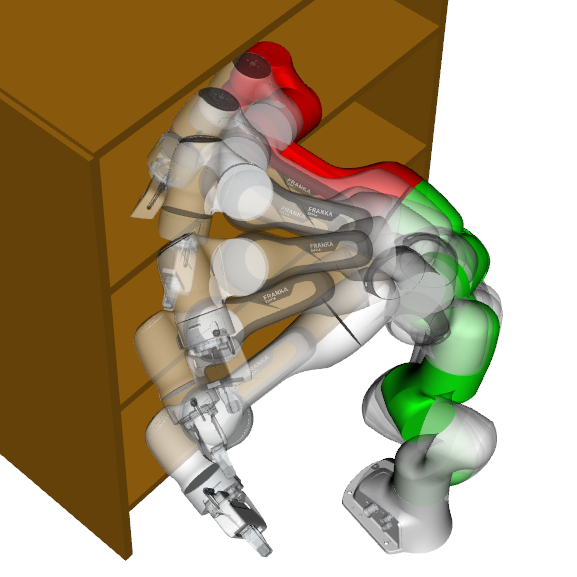}}
    \frame{\includegraphics[width=.49\linewidth, height=.49\linewidth]{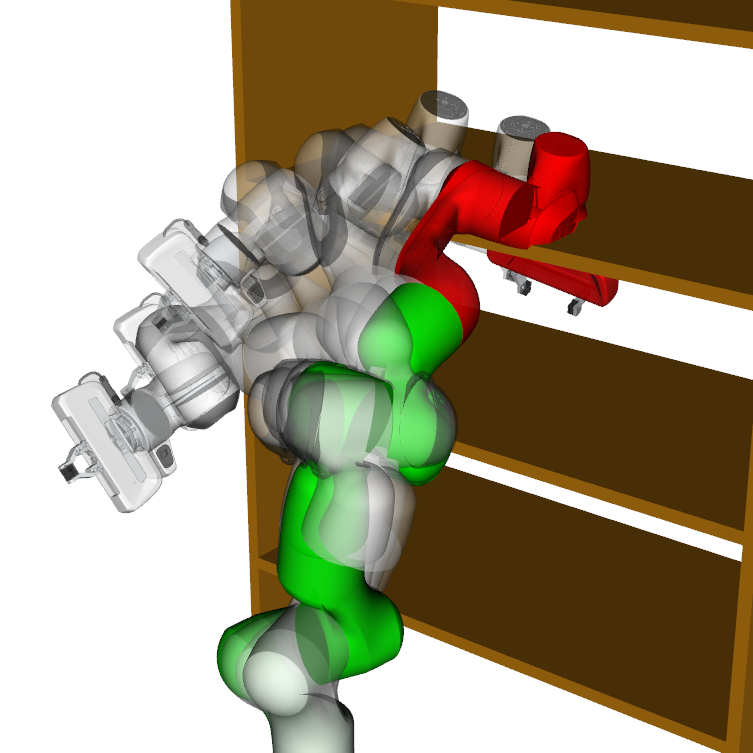}}
    \frame{\includegraphics[width=.49\linewidth, height=.49\linewidth]{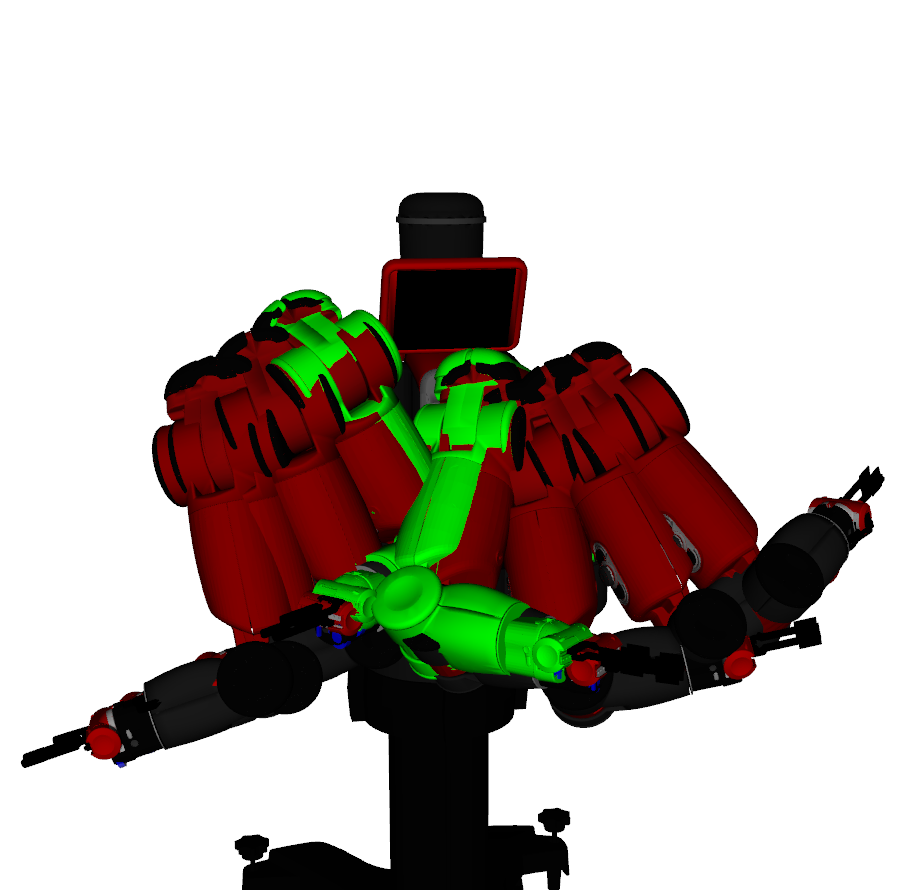}}
    \frame{\includegraphics[width=.49\linewidth, height=.49\linewidth]{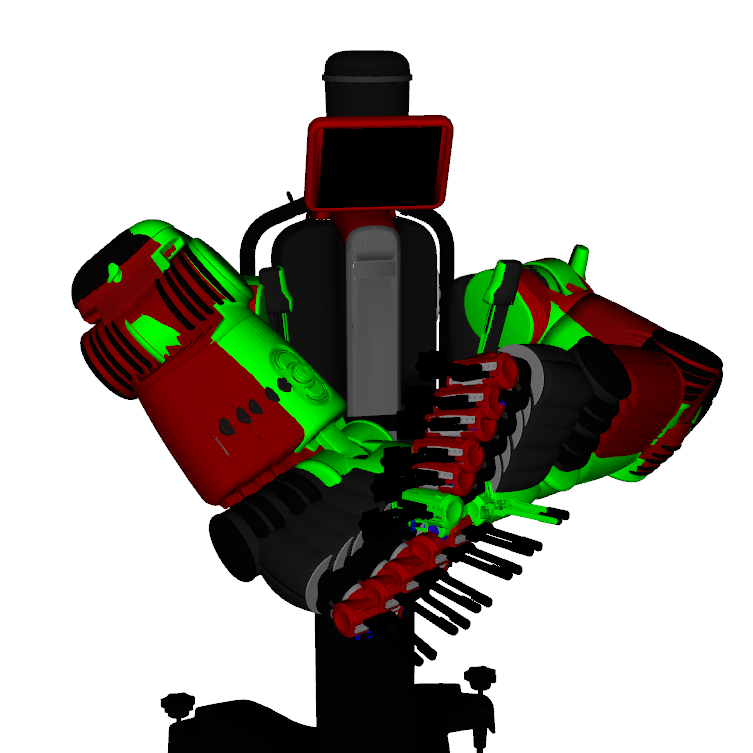}}
    \caption{The proposed differentiable proxy collision detector \shortname can be used to fix in-collision \updated{and self-collision} configurations. Starting configurations are green.} 
    \label{fig:fix}
\end{figure}
\begin{figure}
    \centering
    \includegraphics[width=.6\linewidth, trim=42 38 15 15, clip ]{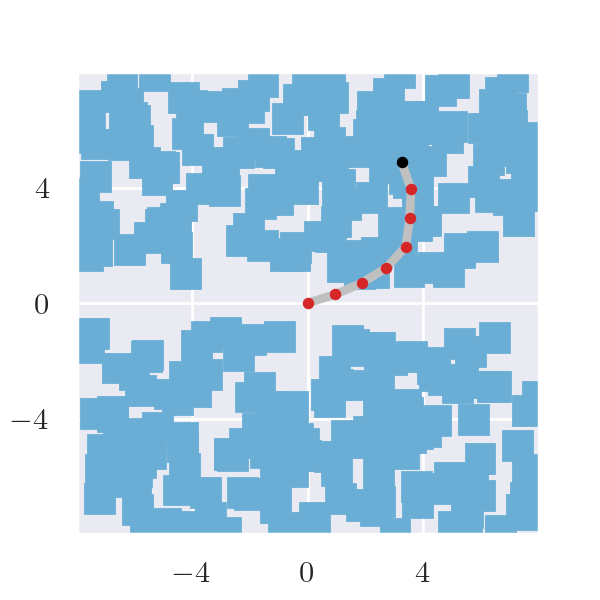}
    \caption{A cluttered environment used to test the efficiency of sampling from $\mathcal{C}_\mathrm{free}$. It contains 300 randomly placed rectangle obstacles with a narrow passage in the middle. Only 1.0\% of the C-space is collision-free.} 
    \label{fig:narrow}
\end{figure}

\subsection{Escaping from High-risk Areas through Gradient Descent}
We now show how the proposed differentiable collision detection method \shortname may use its gradients to produce a configuration path that locally guides the robot to escape from high-risk configurations that are close to obstacles. Fig.~\ref{fig:high-risk} shows an example of a Baxter robot that is told that the cat on the table is a high-risk object.
Given a configuration nearly touching the table obstacle, we let PyTorch's Adam optimizer run for some iterations until the collision score is reduced to be under a threshold. We record all the intermediate waypoints and visualize the trajectory on the robot, showing its path away from the risky object is path-efficient. 


Another set of examples demonstrate what happens when you start with a configuration in collision. Fig.~\ref{fig:fix} shows the proposed differentiable collision detector can be used to fix these in-collision and self-collision configurations. \updated{Note that the dual-arm Baxter robot has 14 DOF, and its self-collision requires checking all pairs of its links, which is computationally expensive. \shortname is able to resolve self-collision in a unified way as it resolves robot-obstacle collision.}

\begin{algorithm}[tbp]
\DontPrintSemicolon 
\newcommand{\opt}{{\sc Opt}}
\newcommand{\usediffco}{\textit{use\shortname}}
\KwIn{\shortname model $\cscore$, safety bias $\epsilon$, Boolean \textit{use\shortname}, number of required states $T$, gradient-based optimizer \opt, number of updating steps $K$.}
\KwOut{$T$ states in $\mathcal{C}_\mathrm{free}$}
$n \leftarrow 0$, $\Cfg\leftarrow \emptyset$\;
\While{n < T}{
    $\cfg \leftarrow$ {\sc randSample()}\;
    \If{\usediffco}{
    \tcp{Projection using \shortname}
    \For{$i\leftarrow 1\ \text{to}\ K$}{
        \lIf{$\cscore(\cfg)+\epsilon < 0$}{
        \Break}
        $\cfg \leftarrow$\opt($\cfg, \cscore, \nabla_{\cfg}\cscore$)
    }}
    \tcp{Verification}
    \If{$\text{\sc trueChecker}(\cfg)=$ \textit{free}}{
        $\Cfg\leftarrow\Cfg\cup\{\cfg\}$\;
        $n\leftarrow n+1$\;
    }
}
\KwRet{$\Cfg$}
\caption{{\sc Sampling C-Space using \shortname}} 
\label{algo:diffcosampling}
\end{algorithm}
\begin{table}[tbp]
\centering
\caption{Comparison between 2 sampling pipelines with and without \shortname projection to obtain 1000 valid configurations in the environment in Fig.~\ref{fig:narrow}.} 
\label{tab:diffcosampling}
\begin{tabular}{@{}lcccc@{}}
\toprule
Method                & Valid States & FCL Check & \shortname Check & Time (s) \\ \midrule
Random    & 1000                & 97817               & -                          & 135.41   \\
\shortname & 1000                & 19275               & 56330                      & 65.22    \\ \bottomrule
\end{tabular}
\end{table}


\updated{This property makes \shortname potentially useful for efficiently sampling from $\mathcal{C}_\mathrm{free}$. 
We design a naive sampling pipeline as described in Algorithm~\ref{algo:diffcosampling}. The value of \textit{use\shortname} controls whether the pipeline uses \shortname to project sample configurations out of collision. Line 3 takes a random sample uniformly from the C-space. Lines 5 computes the \shortname collision score. If it is smaller than zero, it means \shortname believes $\cfg$ is collision-free and the optimizing procedure is stopped; if not, the optimizer will try to decrease the score by updating $\cfg$ using the gradient $\nabla_{\cfg}\cscore$ (Lines 6-7). Line 8 verifies if $\cfg$ is in $\mathcal{C}_\mathrm{free}$. Line 9-10 add $\cfg$ to the set of valid configurations if it passes the verification.}

\begin{figure*}[hbtp]
    \vspace{-0.3cm}
    \centering    
    \begin{subfigure}{0.49\linewidth}
    \includegraphics[width=0.49\linewidth, trim=42 38 15 15, clip] {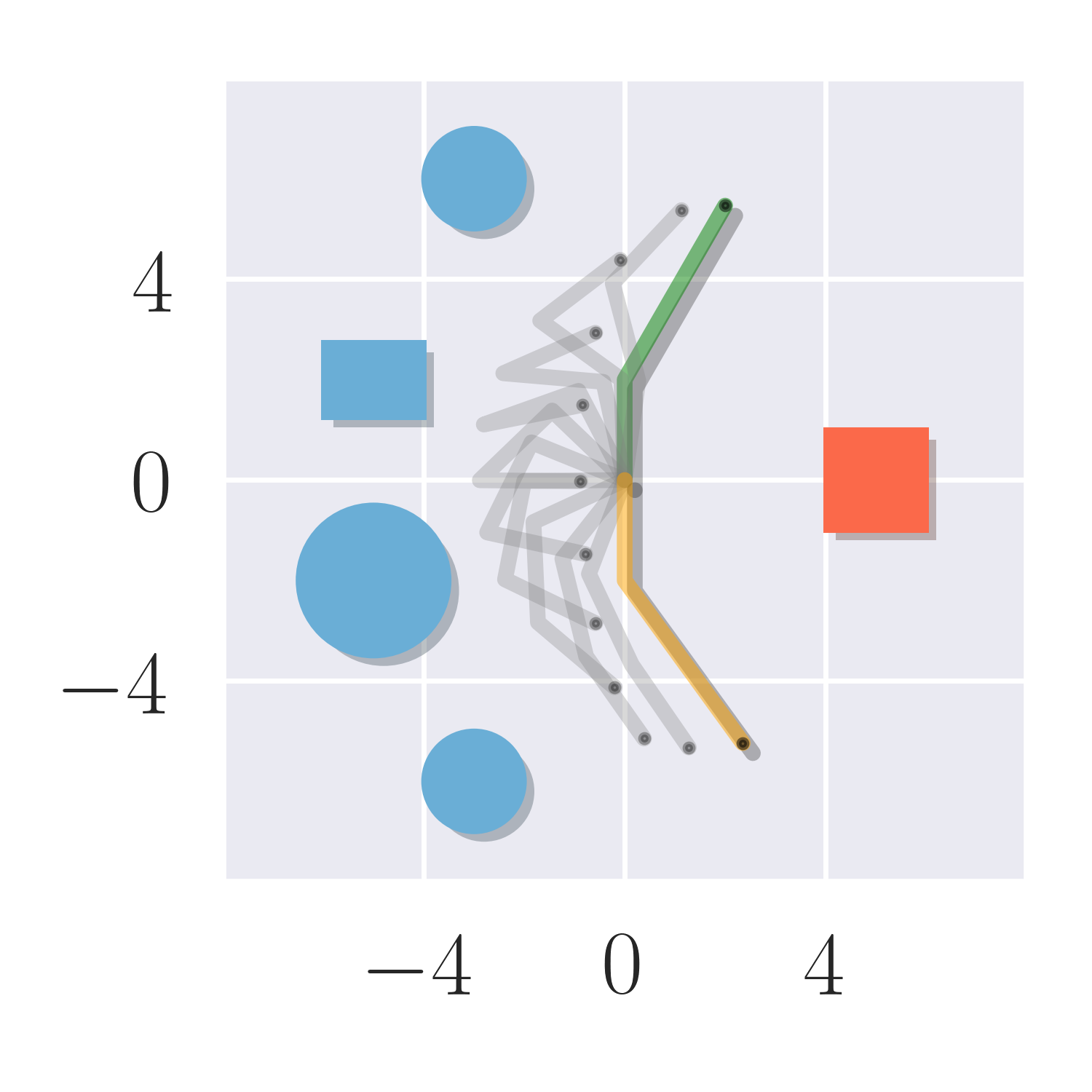}
    \includegraphics[width=0.49\linewidth, trim=42 38 15 15, clip] {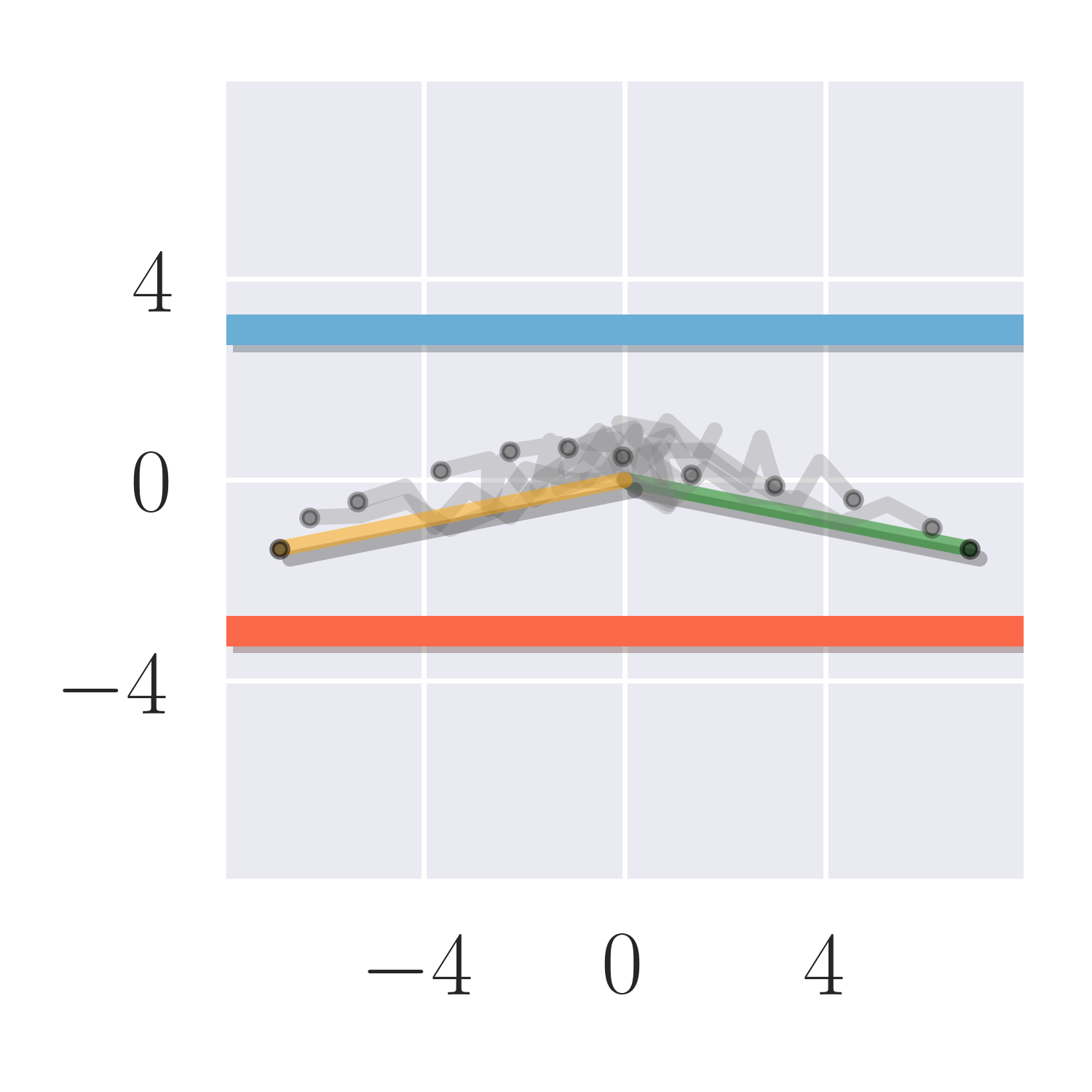}
    \caption{$10\times$ bias}
    \label{fig:avoidimportantobj}
    \end{subfigure}
    \begin{subfigure}{0.49\linewidth}
    \includegraphics[width=0.49\linewidth, trim=42 38 15 15, clip] {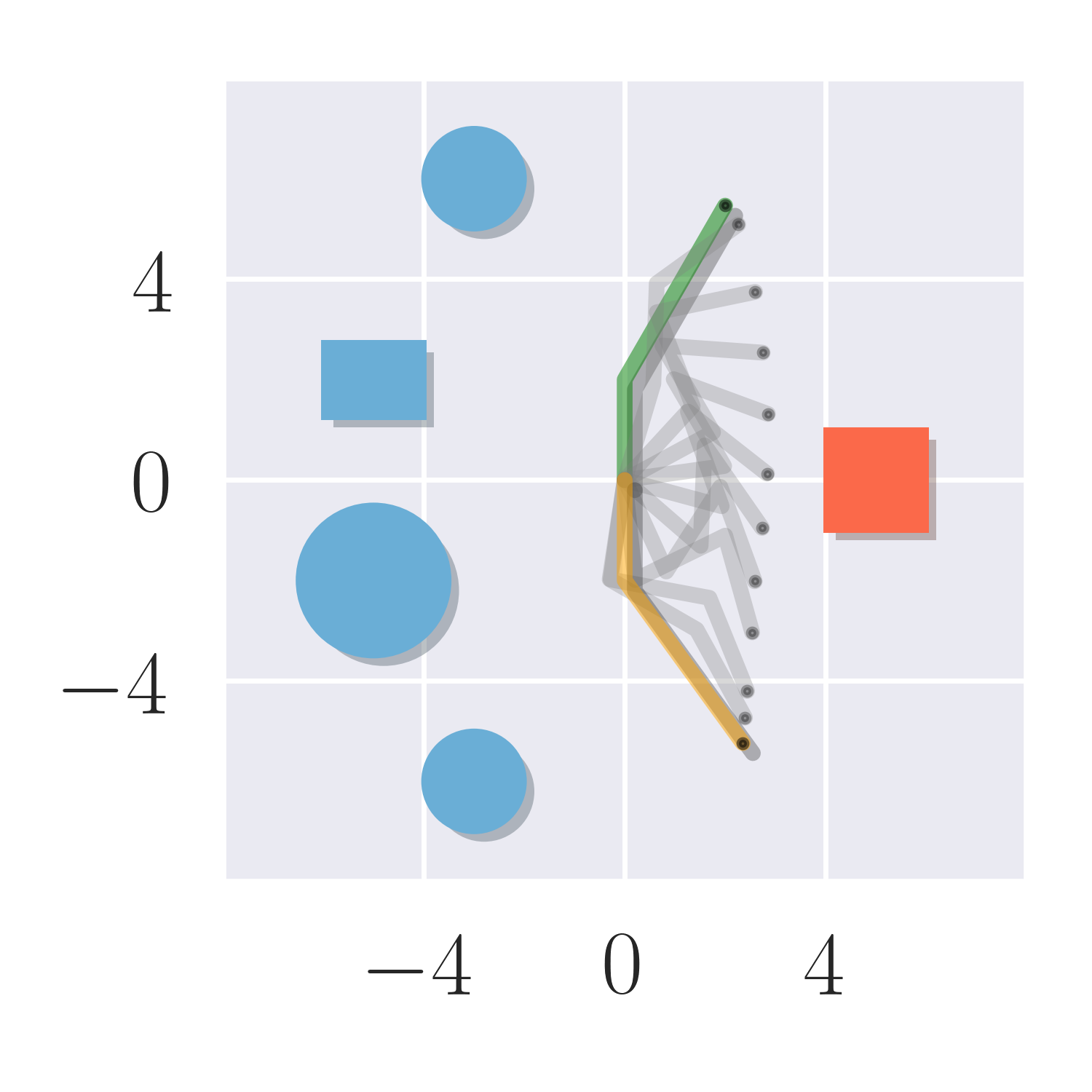}
    \includegraphics[width=0.49\linewidth, trim=42 38 15 15, clip] {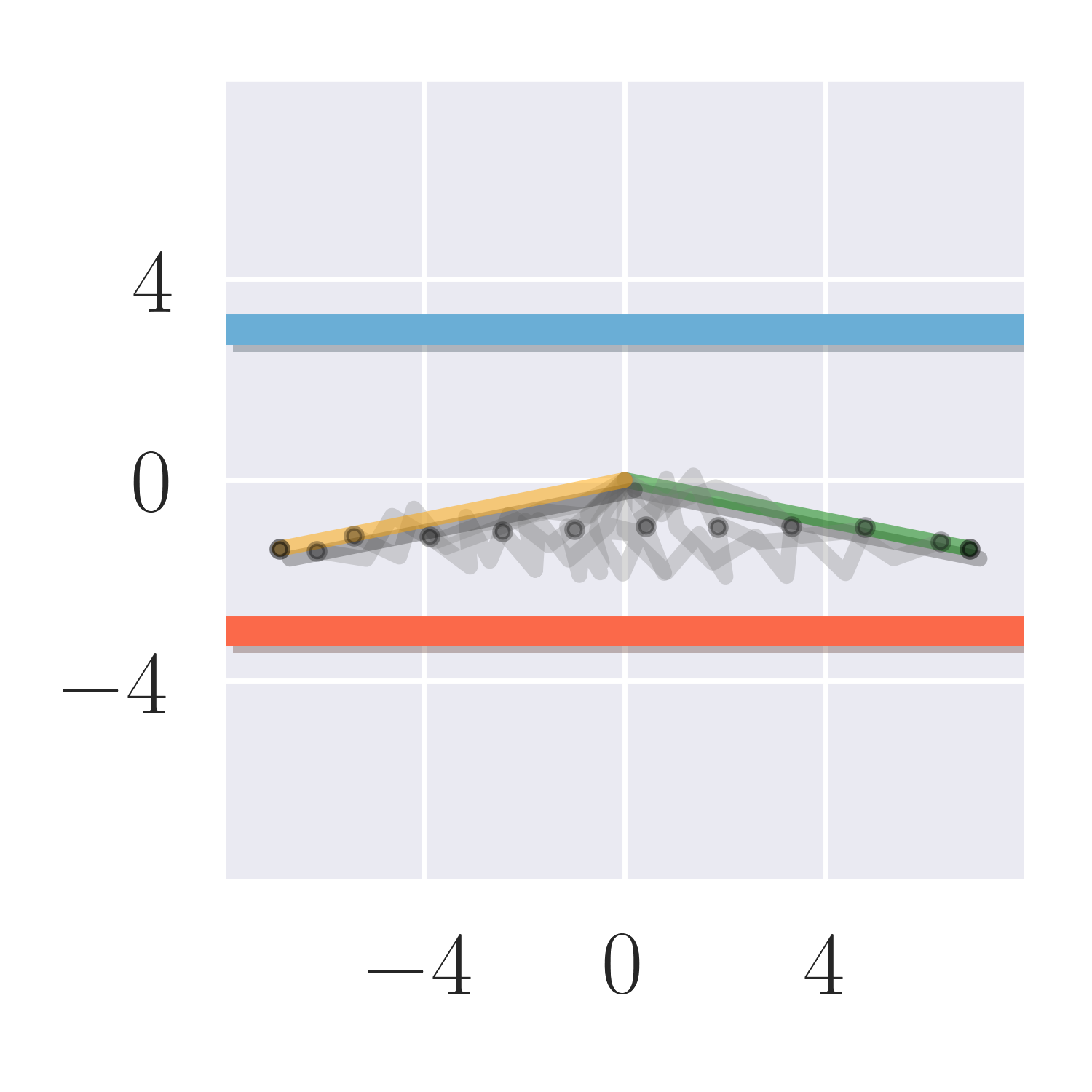}
    \caption{Equal bias}
    \label{fig:equalmargin}
    \end{subfigure}
    \caption{\updated{The two robots are 3-link and 7-link planar robots, respectively.} Our method is able to provide distinguished safety biases for more important objects. (a) exhibits the solutions when a 10$\times$ safety bias is attached to the orange object, which tend to keep larger distance to the important object. In (b), all objects receive equal safety biases, thus the solutions tend to obtain the minimum cost of moving control points.} 
    \label{fig:safetybias}
\end{figure*}

\begin{figure*}[hbtp]
    \centering
    
    \begin{subfigure}{0.24\textwidth}
    \includegraphics[width=\linewidth]{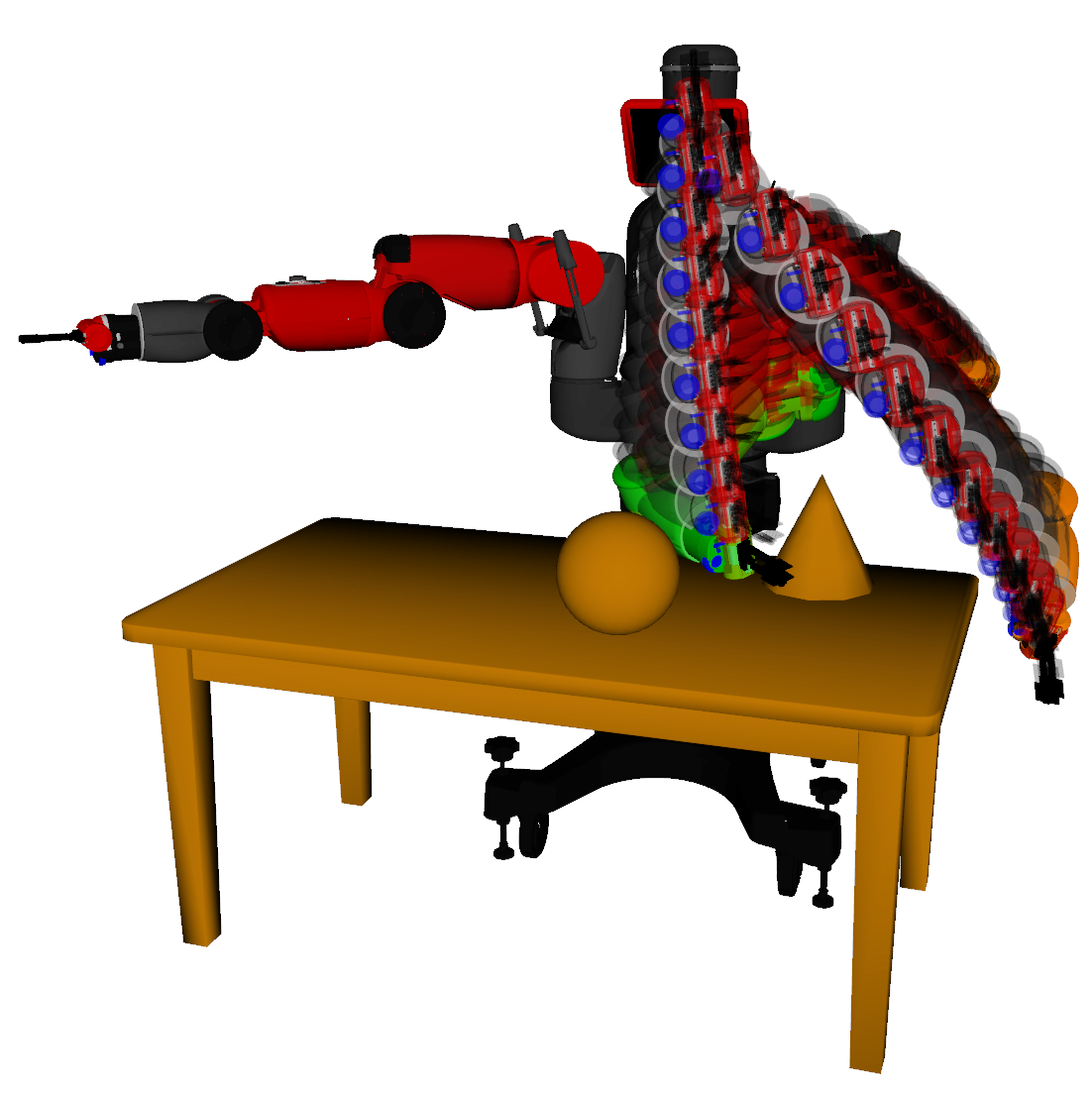}
    \caption{RRT, 1st simple scene.}
    \end{subfigure}
    \begin{subfigure}{0.24\textwidth}
    \includegraphics[width=\linewidth]{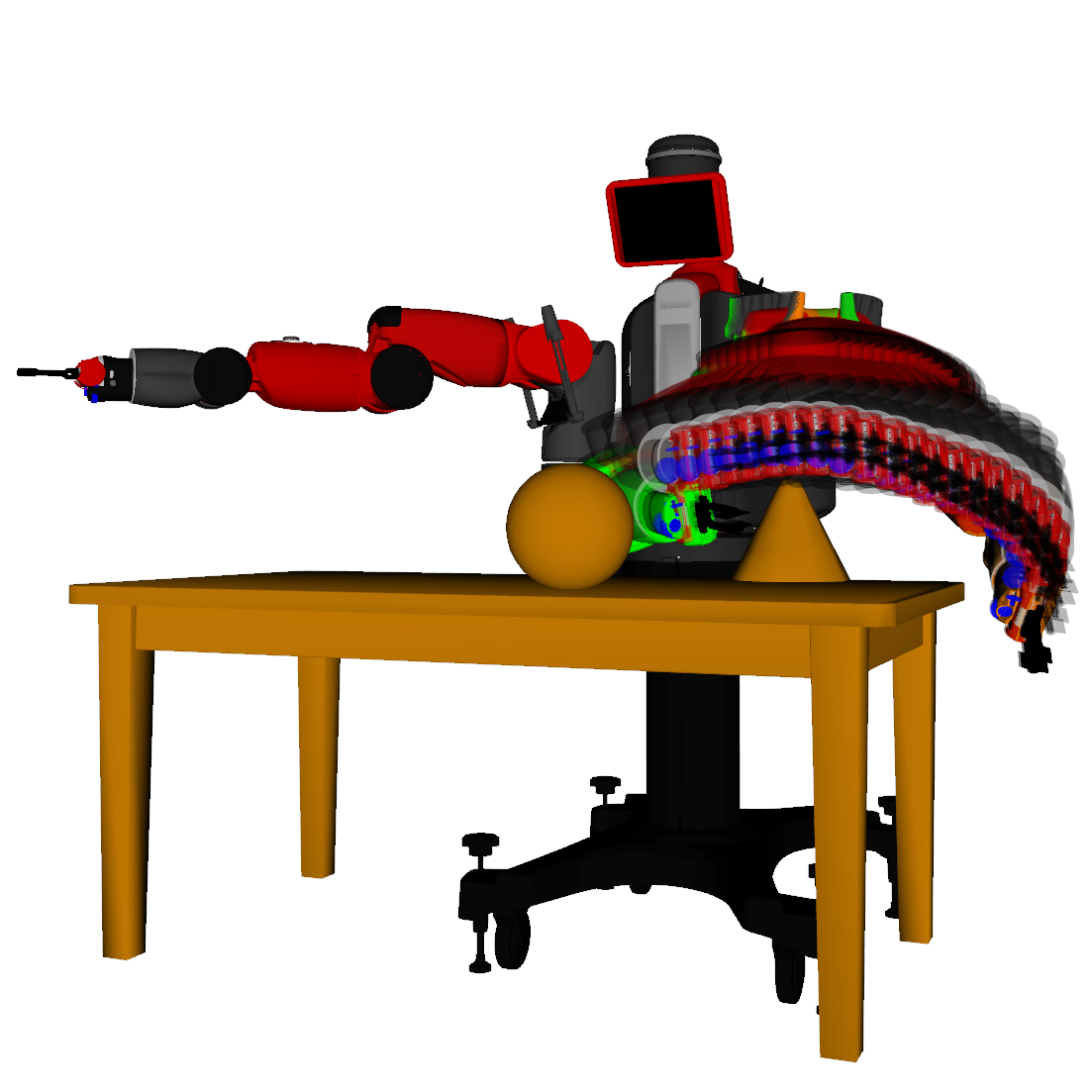} 
    \caption{Optimized, 1st simple scene.}
    \end{subfigure}
    \begin{subfigure}{0.24\textwidth}
    \includegraphics[width=\linewidth]{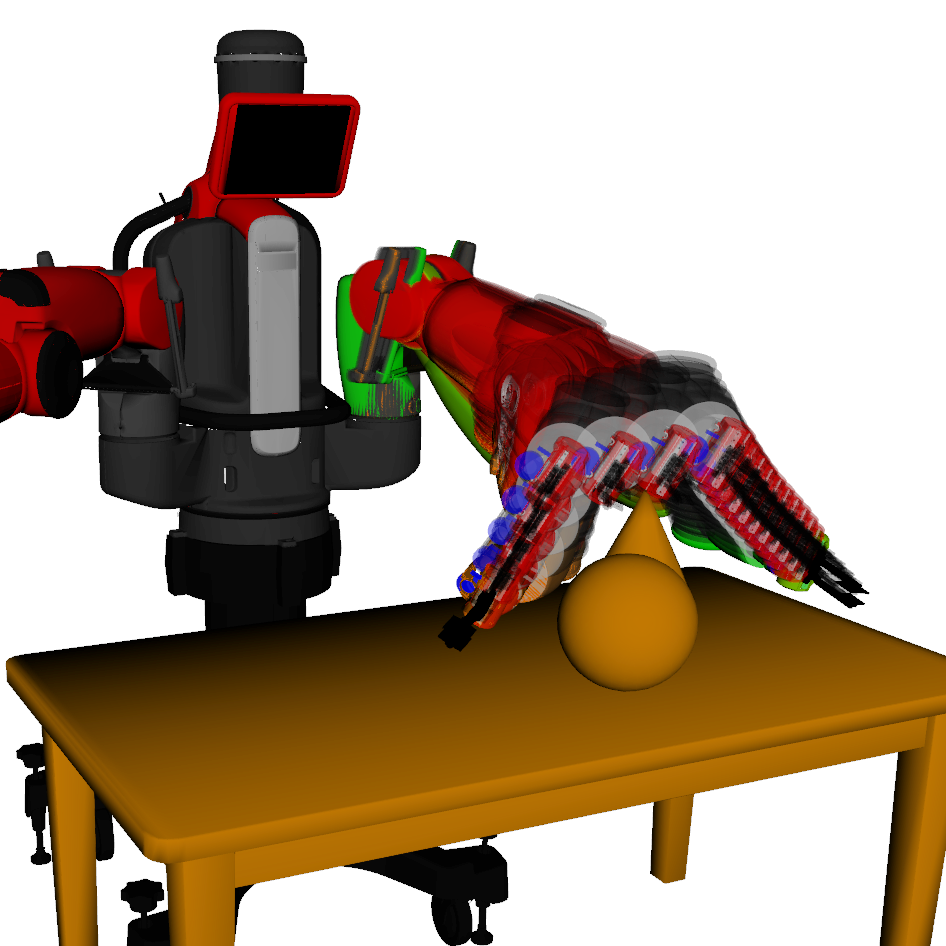}
    \caption{RRT, 2nd simple scene.}
    \end{subfigure}
    \begin{subfigure}{0.24\textwidth}
    \includegraphics[width=\linewidth]{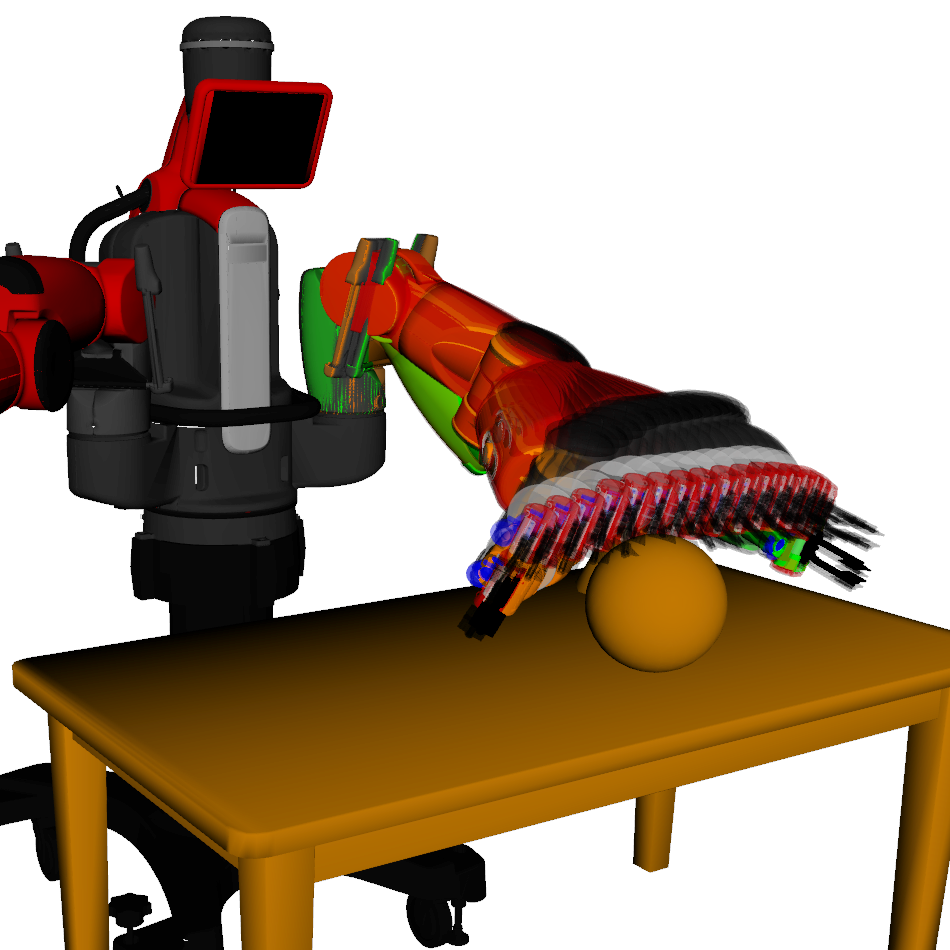}
    \caption{Optimized, 2nd simple scene.}
    \end{subfigure}
    \begin{subfigure}{0.24\textwidth}
    \includegraphics[width=\linewidth]{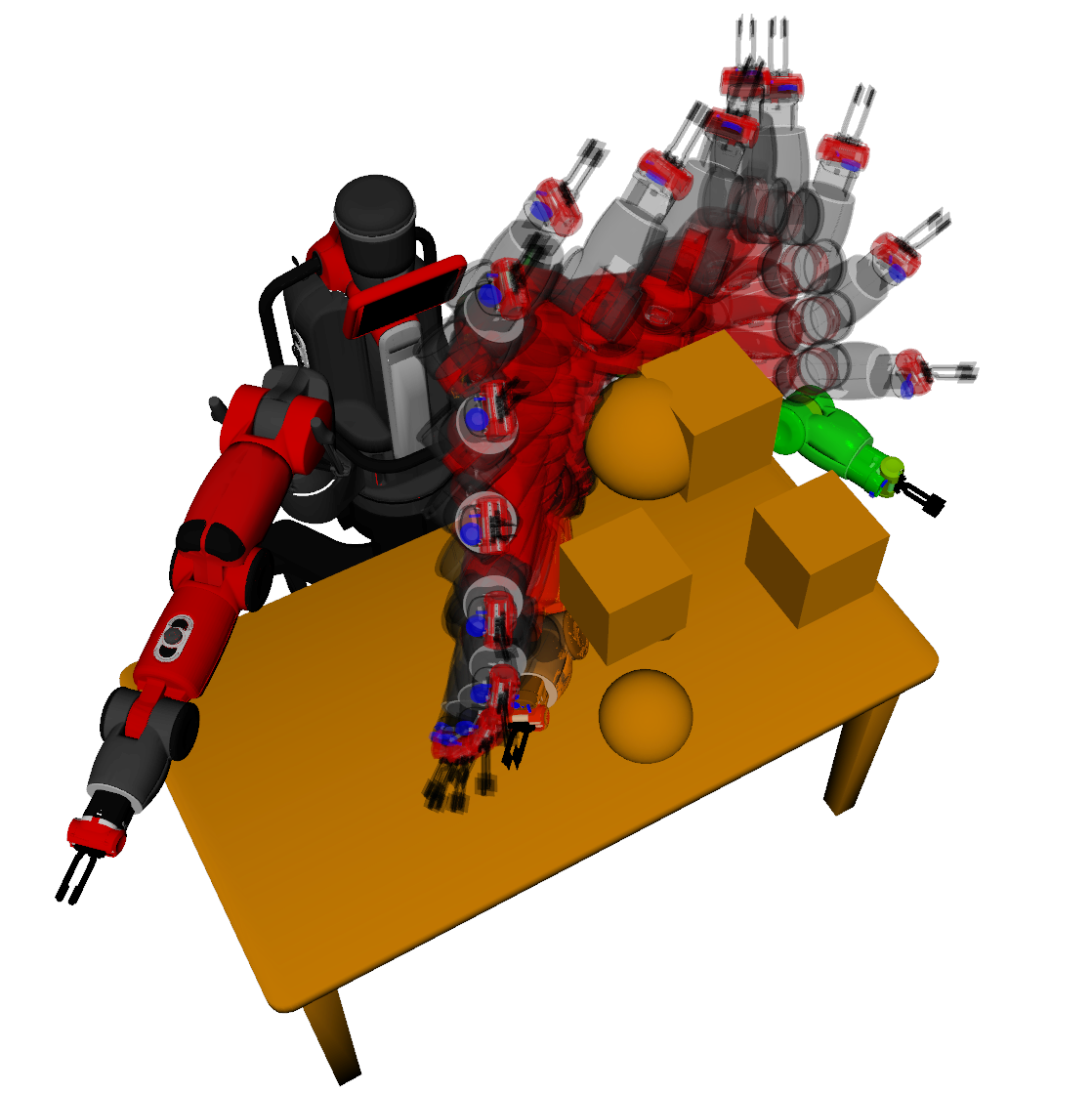}
    \caption{RRT, complex, 1st view.}
    \end{subfigure}
    \begin{subfigure}{0.24\textwidth}
    \includegraphics[width=\linewidth]{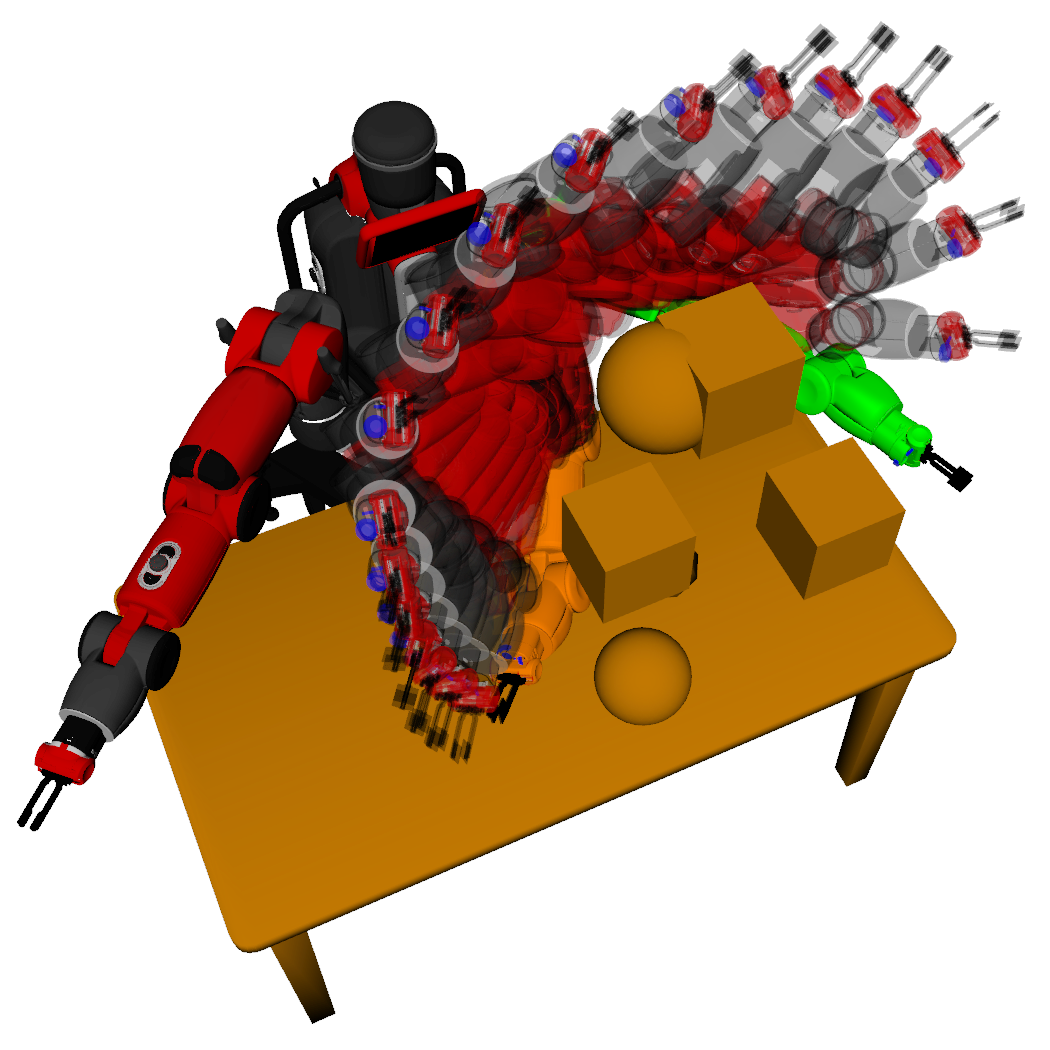} 
    \caption{Optimized, complex, 1st view.}
    \end{subfigure}
    \begin{subfigure}{0.24\textwidth}
    \includegraphics[width=\linewidth]{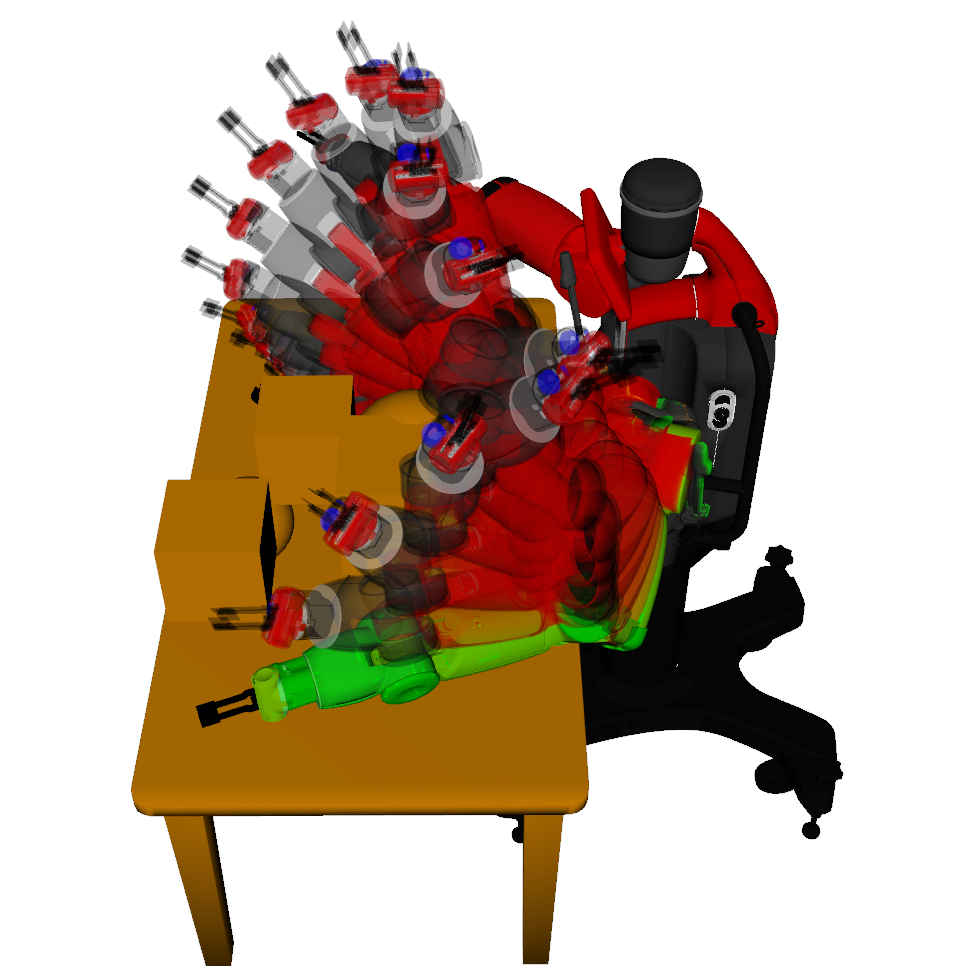}
    \caption{RRT, complex, 2nd view.}
    \end{subfigure}
    \begin{subfigure}{0.24\textwidth}
    \includegraphics[width=\linewidth]{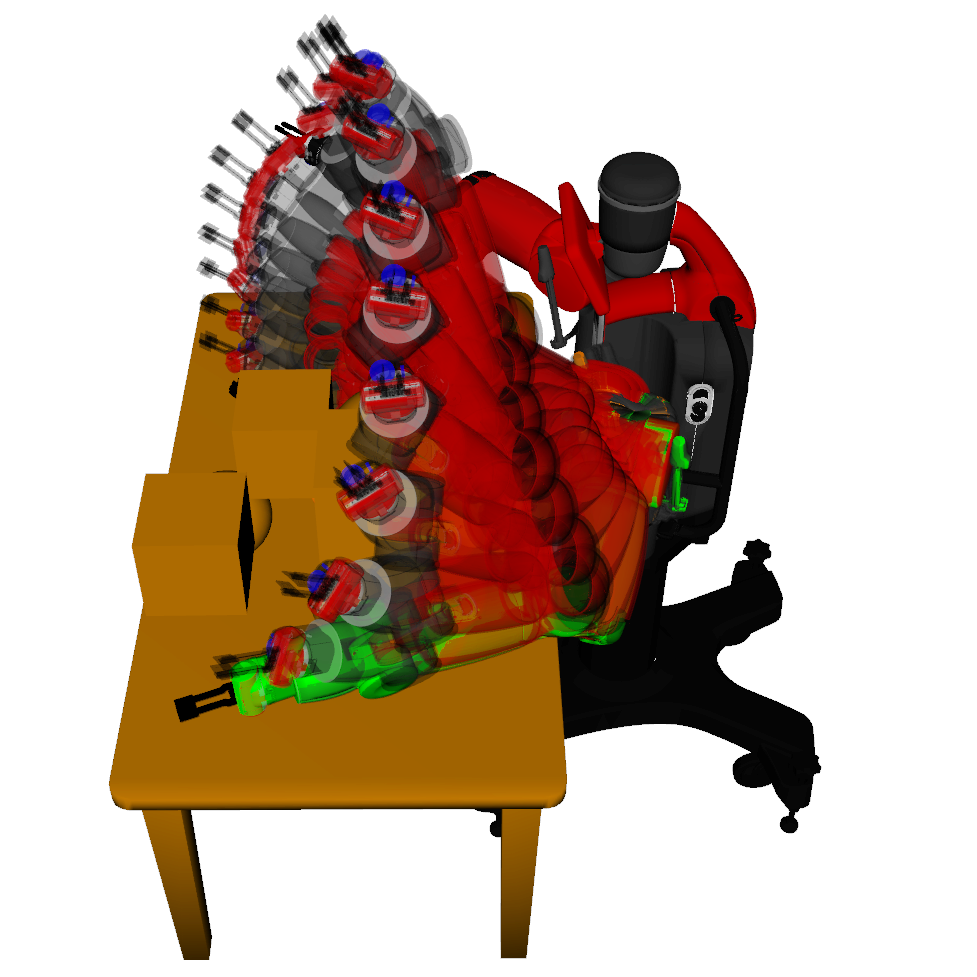}
    \caption{Optimized, 2nd view.}
    \end{subfigure}
    
    \begin{subfigure}{0.24\textwidth}
    \includegraphics[width=\linewidth]{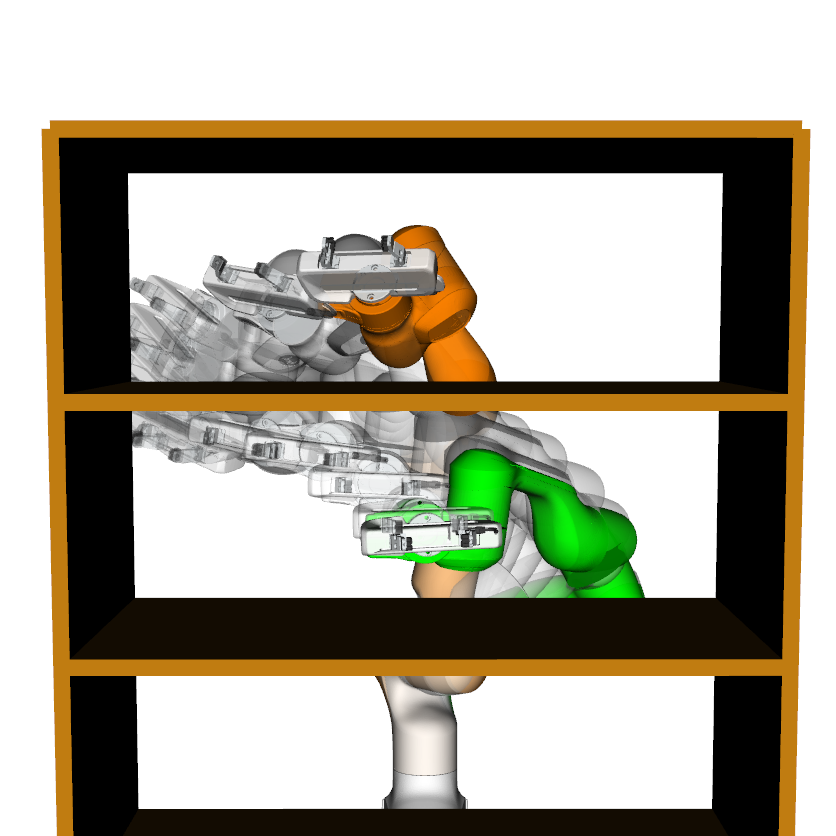}
    \caption{RRT, bookshelves, 1st view.}
    \end{subfigure}
    \begin{subfigure}{0.24\textwidth}
    \includegraphics[width=\linewidth]{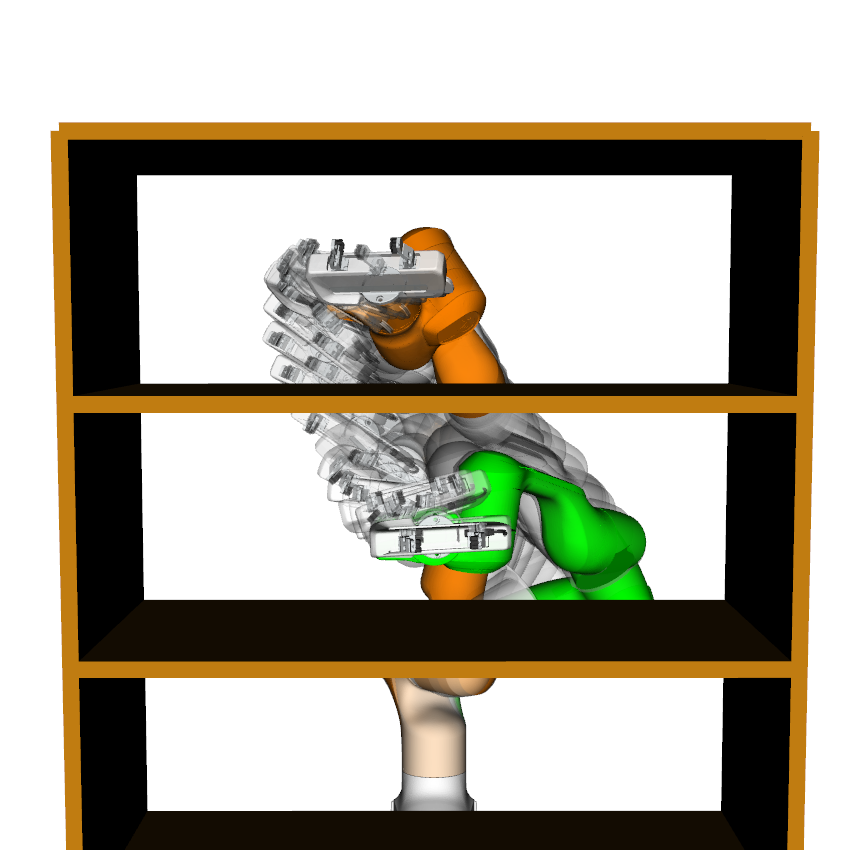} 
    \caption{Optimized, 1st view.}
    \end{subfigure}
    \begin{subfigure}{0.24\textwidth}
    \includegraphics[width=\linewidth]{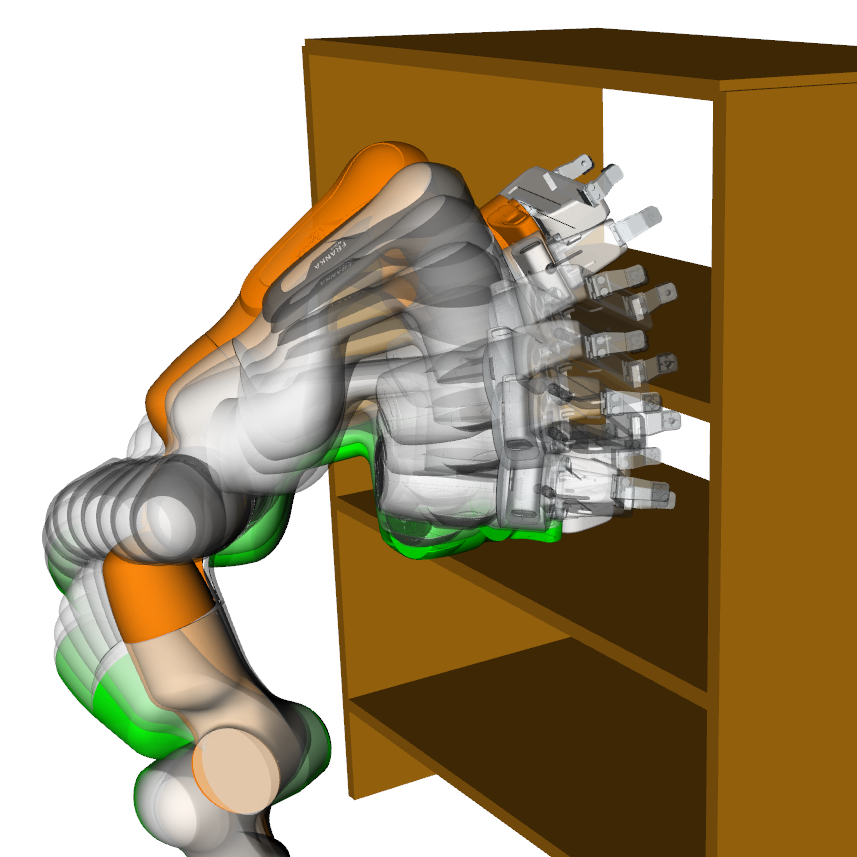}
    \caption{RRT, bookshelves, 2nd view.}
    \end{subfigure}
    \begin{subfigure}{0.24\textwidth}
    \includegraphics[width=\linewidth]{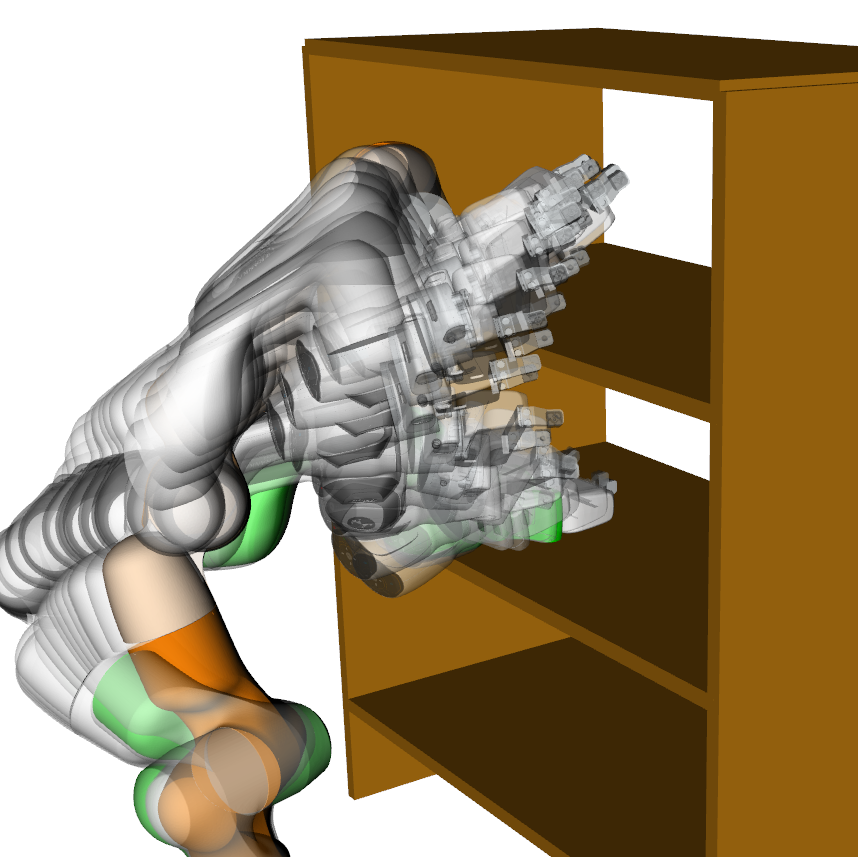}
    \caption{Optimized, 2nd view.}
    \end{subfigure}
    
    \caption{(a, c, e, g, i, k) are trajectories of the Baxter robot (left arm) and Panda arm generated by RRT, the accompanying figures are optimized trajectories produced by our method using the corresponding RRT results as initializations. It shows the proposed differentiable collision detection method can work with trajectory optimization methods to optimize existing results given by completeness-guaranteed motion planning algorithms to produce smoother and safer solutions with equivalent or lower cost. \second{The optimization of the 4 paths took 1.09, 0.94, 1.44, and 1.15 seconds, respectively.}}
    \label{fig:optimizerrt}
\end{figure*}

\begin{figure*}[hb]
    \centering
    \includegraphics[width=.24\linewidth, trim=10 25 26 44, clip]{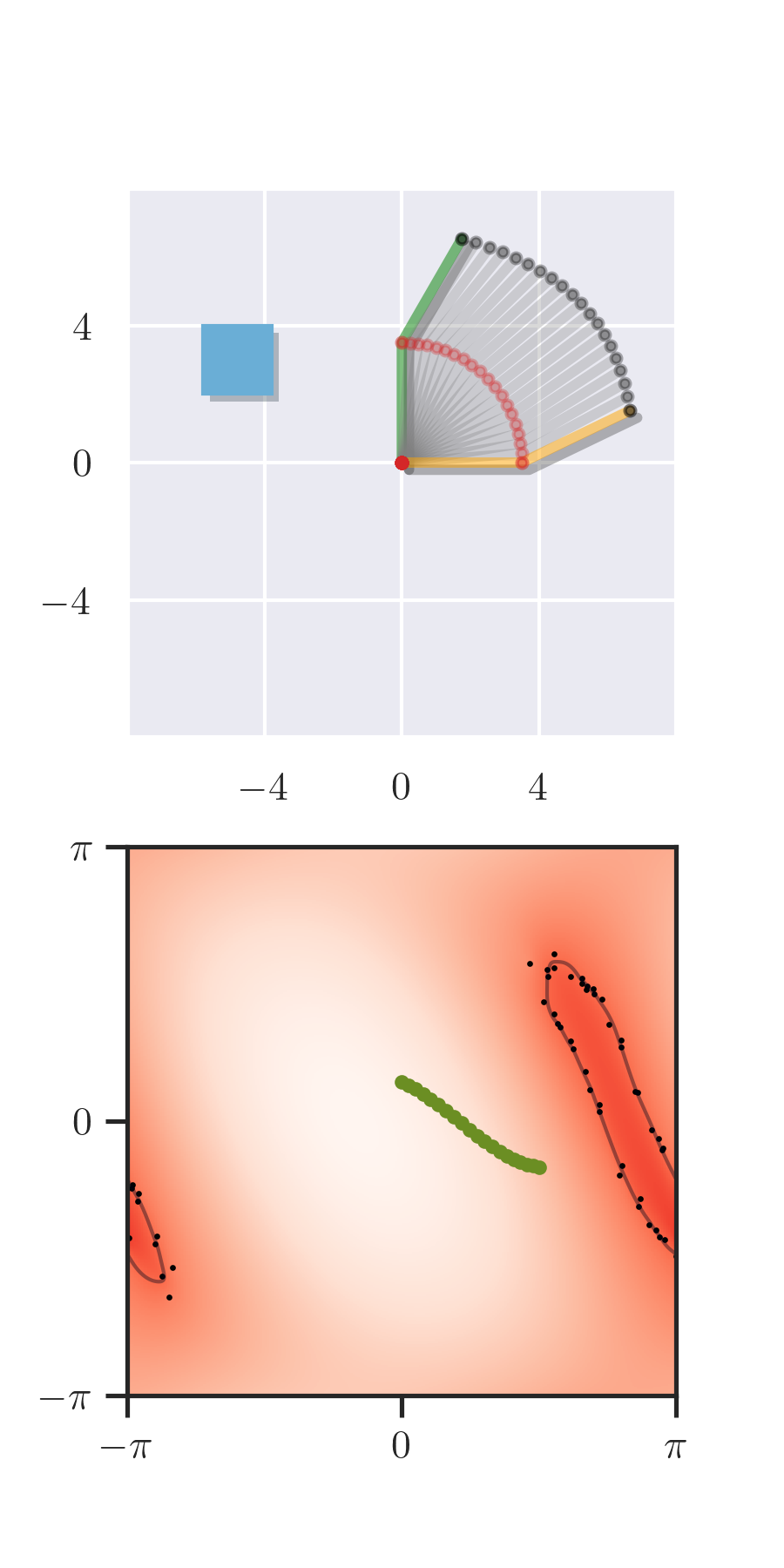}
    \includegraphics[width=.24\linewidth, trim=10 25 26 44, clip]{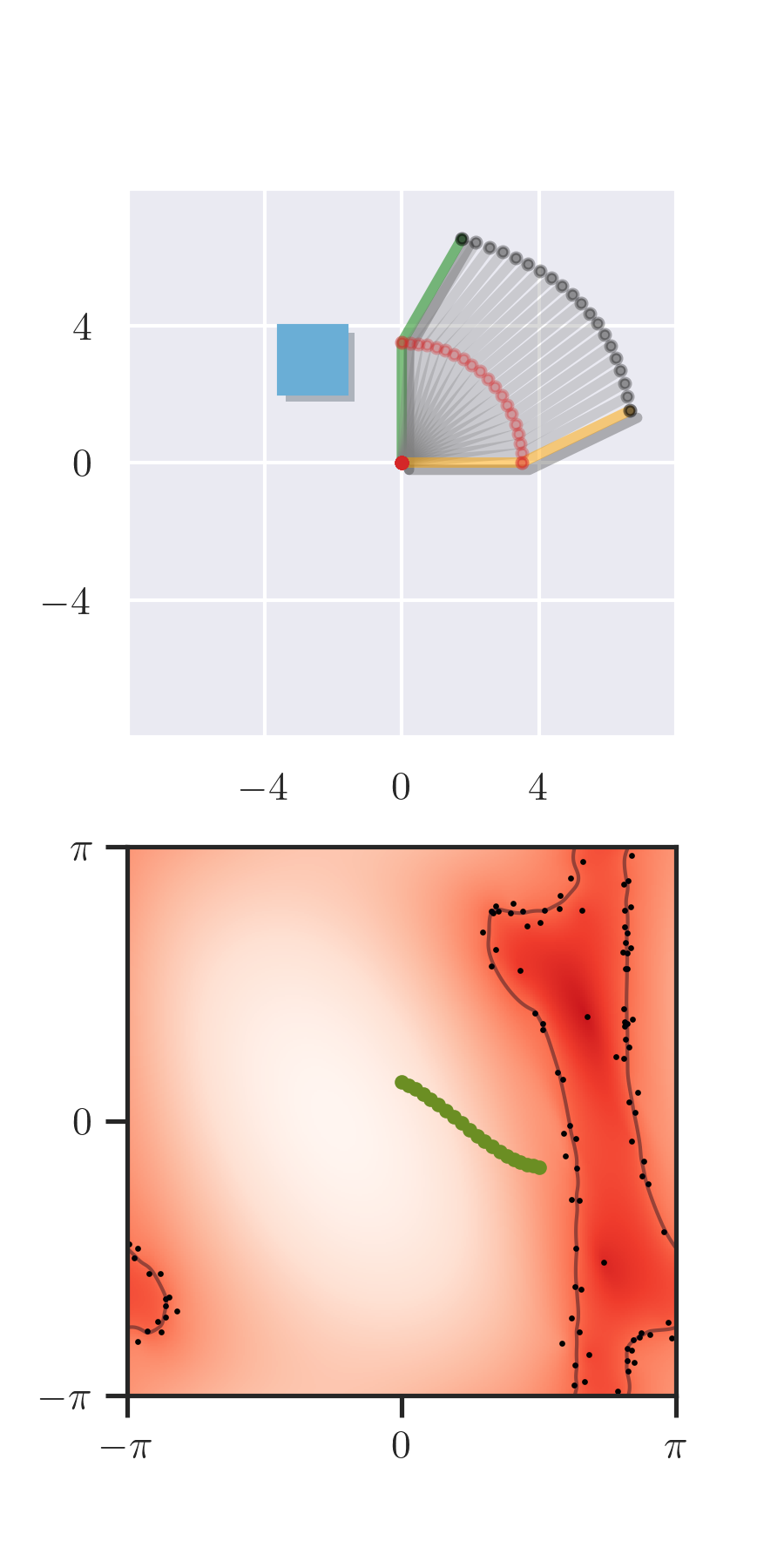}
    \includegraphics[width=.24\linewidth, trim=10 25 26 44, clip]{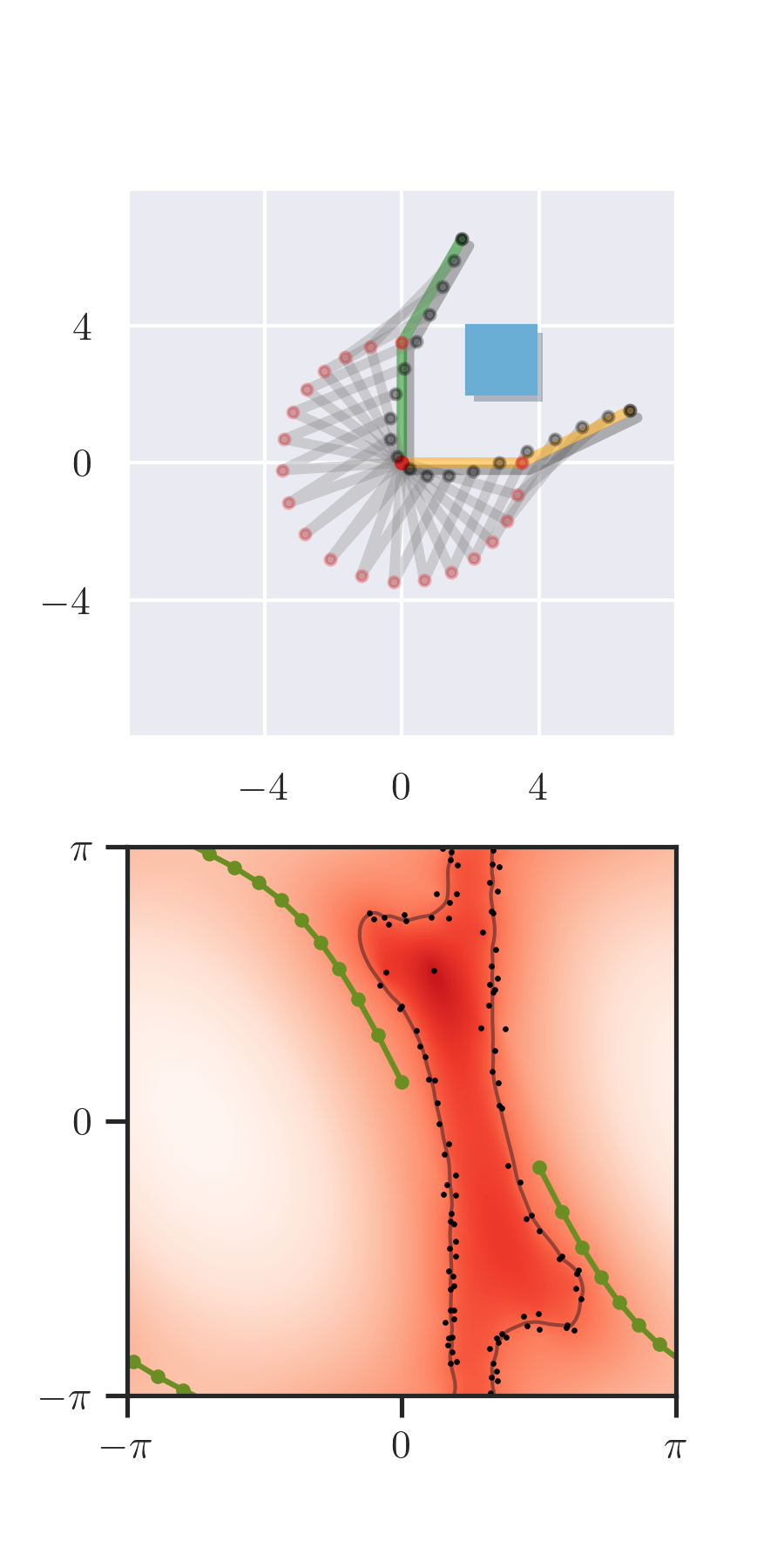}
    \includegraphics[width=.24\linewidth, trim=10 25 26 44, clip]{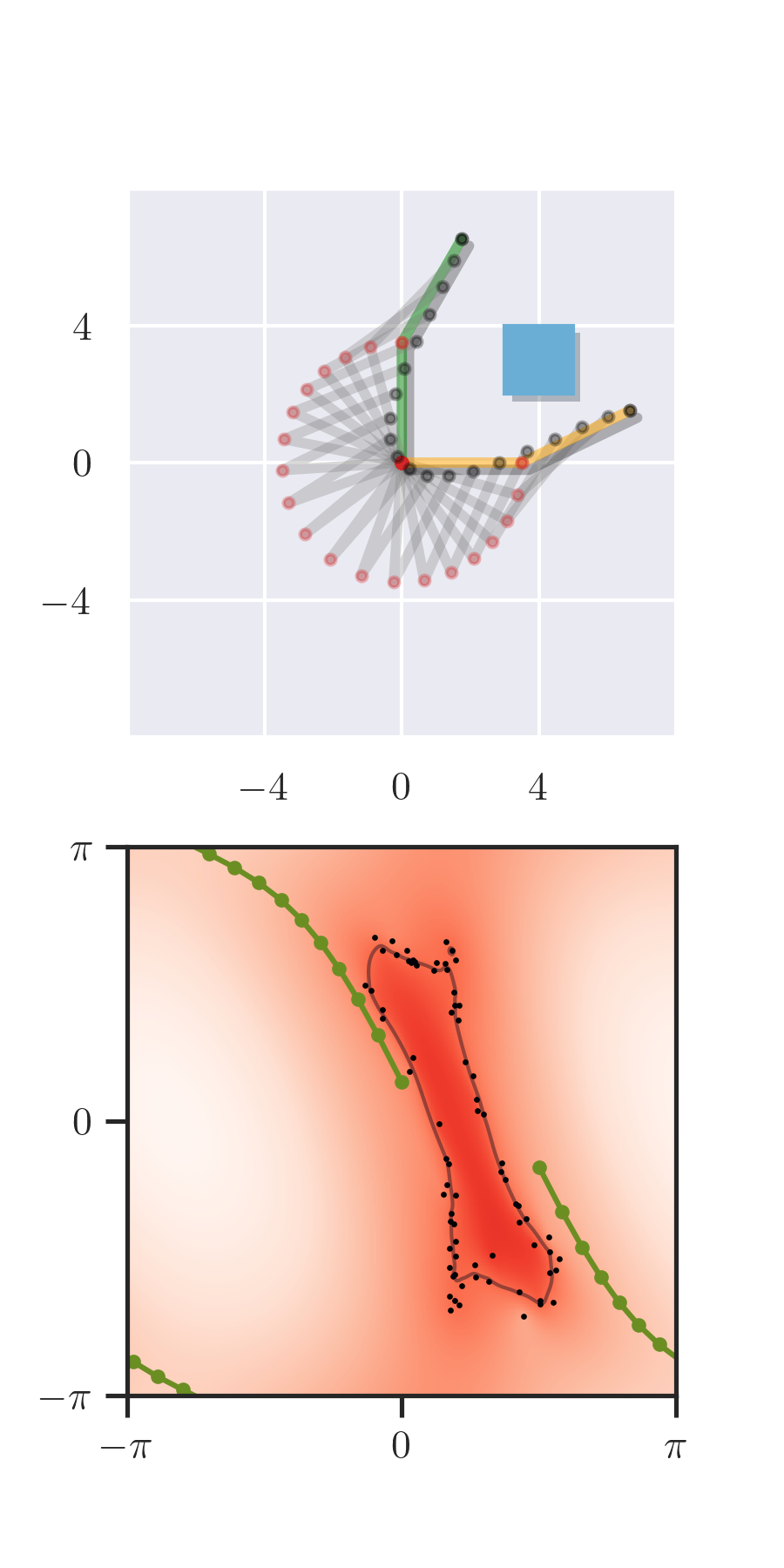}
    \caption{With active learning strategy enabled, the proxy collision detector is updated as the obstacle moves forward. The robot is able to choose the shortest path at the beginning, but has to switch to a further but safe trajectory when the obstacle gets in the way. (Note the wrapping-around property of the angular space; the path is still continuous.)}
    \label{fig:dynamic}
\end{figure*}
\begin{table*}[hb]
\centering
\caption{Comparing time consumed in trajectory optimization with dynamic objects and the costs of the generated paths when using FCL and \shortname as the collision detector. } 
\label{tab:active}
\begin{tabular}{@{}cccccc@{}}
\toprule
Collision Detector     & Initial Training (s) & Updating (s)     & Trajectory Optimization (s) & Total Time (s)      & Cost             \\ \midrule
FCL        & -                    & -                & $72.145\pm 109.605$   & $72.145\pm 109.605$ & $9.732\pm 8.958$ \\
\shortname & $2.423$              & $0.271\pm 0.054$ & $1.991\pm 2.499$      & $2.531\pm2.553$     & $9.732\pm 8.958$ \\ \bottomrule
\end{tabular}
\end{table*}

\updated{In the cluttered environment \second{with a 7-DOF planar robot} shown in Fig.~\ref{fig:narrow}, we compare the time and the number of collision checks used to sample 1000 collision-free configurations with and without \shortname. FCL is used as {\sc trueChecker}. When using \shortname, we use Adam optimizer with a learning rate of 0.2; the safety bias $\epsilon$ is set to 0, and the maximum number of optimizer steps $K$ is 3. Results in Table~\ref{tab:diffcosampling}
suggest that, with \shortname, and a proper trade-off between fixing sampled in-collision configurations and simply re-sampling, one may reduce the number of samples and the time required in sampling-based motion planning methods and end-to-end motion planning methods based on machine learning \cite{qureshi2019motion}.}

\subsection{Trajectory Optimization that Avoids Important Objects}
We show the proposed differentiable proxy collision detector can work with trajectory optimization to produce smooth C-space paths that keep larger safety margins with more important objects. In this experiment, we apply the method described in Sec.~\ref{sec:traj} to 3-DOF, 7-DOF planar robots, Baxter robot, \updated{and a Panda arm} in environments with different categories of obstacles. The Adam optimizer is used in this section.

In Fig.~\ref{fig:avoidimportantobj}, the robot trajectories are optimized while the safety bias $\epsilon_c$ added to the orange objects are $10\times$ larger than that of the blue ones, demonstrating its capability for semantically informed, safety-aware trajectory optimization. The robots try to stay away from the orange objects as much as possible in this case. As a comparison, when we set the safety biases of all objects to be equal, the robots would rather take trajectories that produce less total movement (Fig.~\ref{fig:equalmargin}). 

Fig.~\ref{fig:optimizerrt}(a,c,e,g\updated{,i,k}) visualize trajectories of the Baxter robot and Panda arm generated by RRT. These solutions are used as seeds in trajectory optimization that employs \shortname as a proxy collision detector. The optimized trajectories are presented in Fig.~\ref{fig:optimizerrt}(b,d,f,h\updated{,j,l}). 

The objective $f(\cfg)$ of the Baxter experiments are set to be a sum of the end-effector movement and joint angle changes,
\begin{equation}
    \begin{aligned}
    f_{\text{Baxter}}(\cfg) = \sum_{t=1}^{T-1} &w_p\|\text{ee}(\cfg)_{t+1}-\text{ee}(\cfg)_{t}\|^2 \\
    + &w_\theta\|\cfg_{t+1}-\cfg_{t}\|^2.
    \end{aligned}
\end{equation}
We use $w_p = w_\theta = 1$, and set the constraint weights $\mu_i=2, \forall i$. For the planar robot experiments, we calculate the movement of all joint positions in the objective, set $w_p=1$, $w_\theta=0$, and $\mu_i=10, \forall i$. The total number of waypoints $T=20$. 

\begin{figure*}[hbp]
    \centering
    \includegraphics[width=0.245\linewidth, trim={30, 23, 20, 23}, clip]{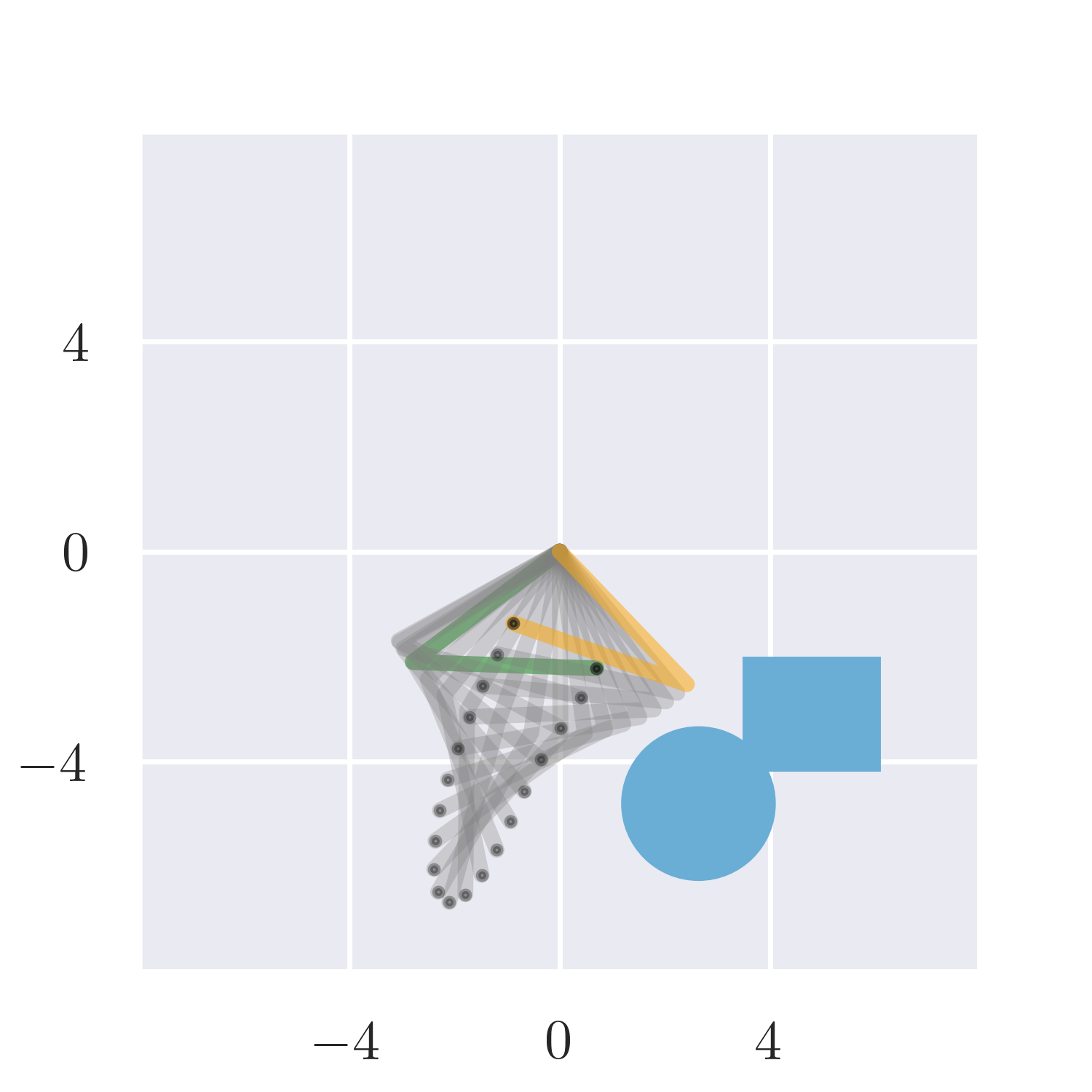}
    \includegraphics[width=0.245\linewidth, trim={30, 23, 20, 23}, clip]{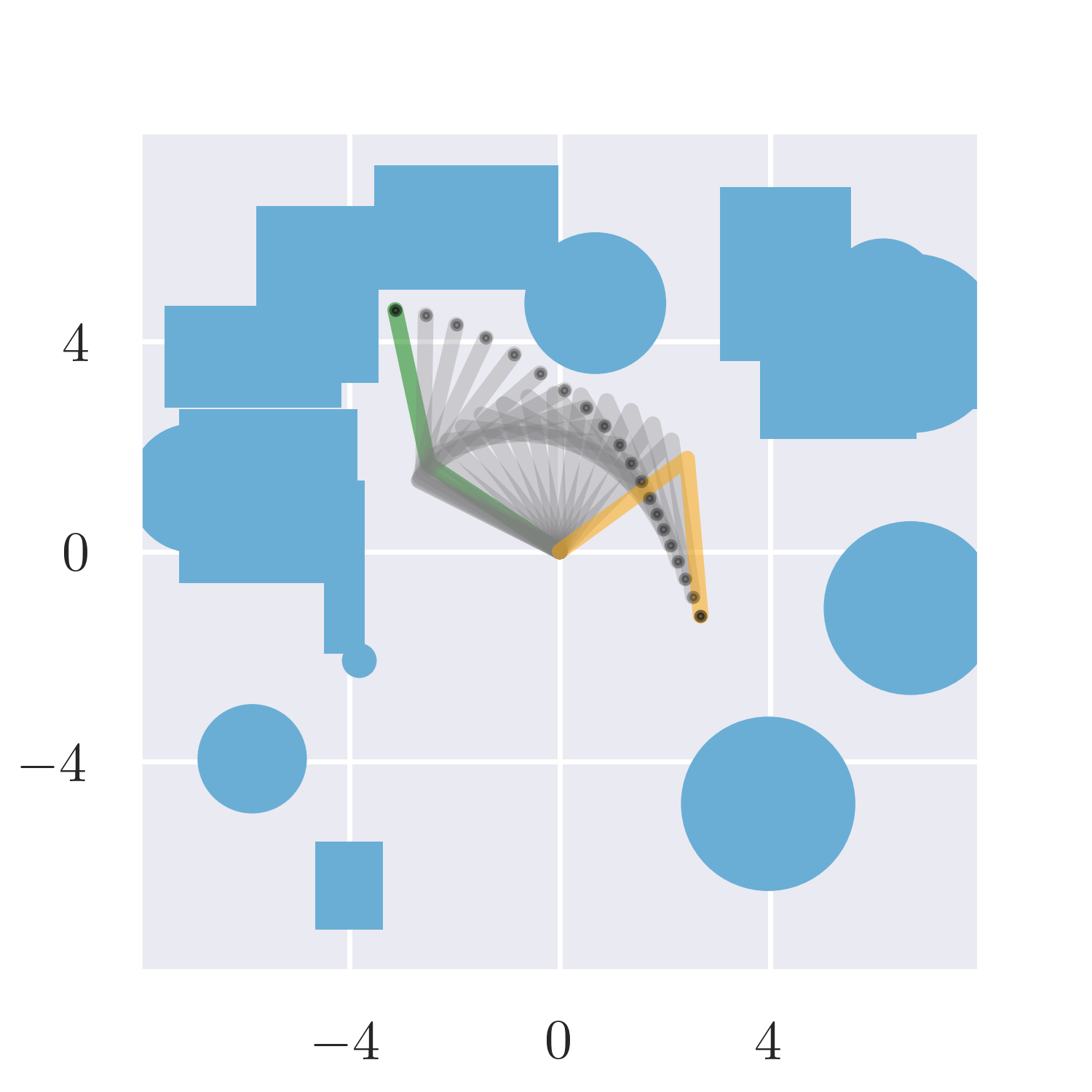}
    \includegraphics[width=0.245\linewidth, trim={30, 23, 20, 23}, clip]{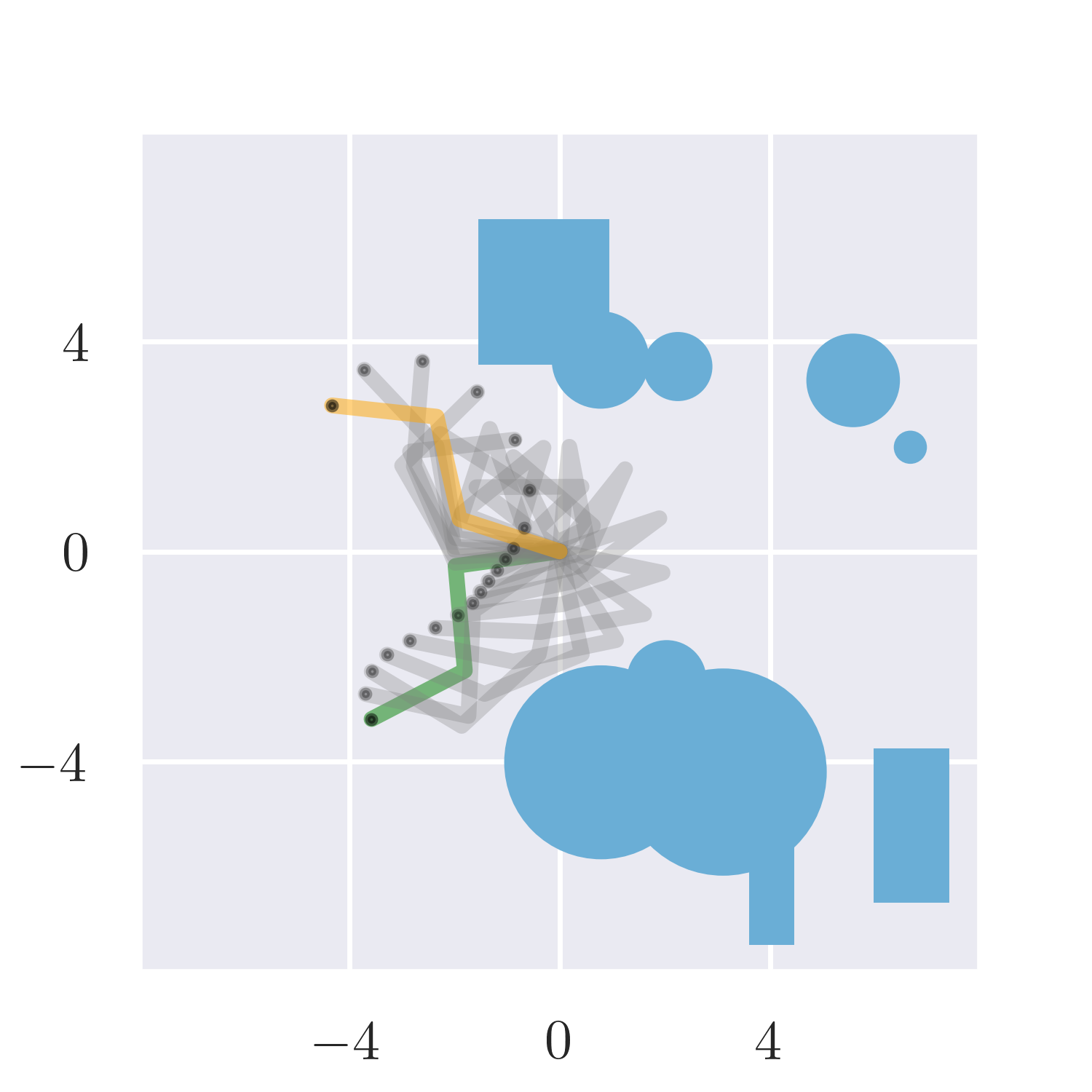}
    \includegraphics[width=0.245\linewidth, trim={30, 23, 20, 23}, clip]{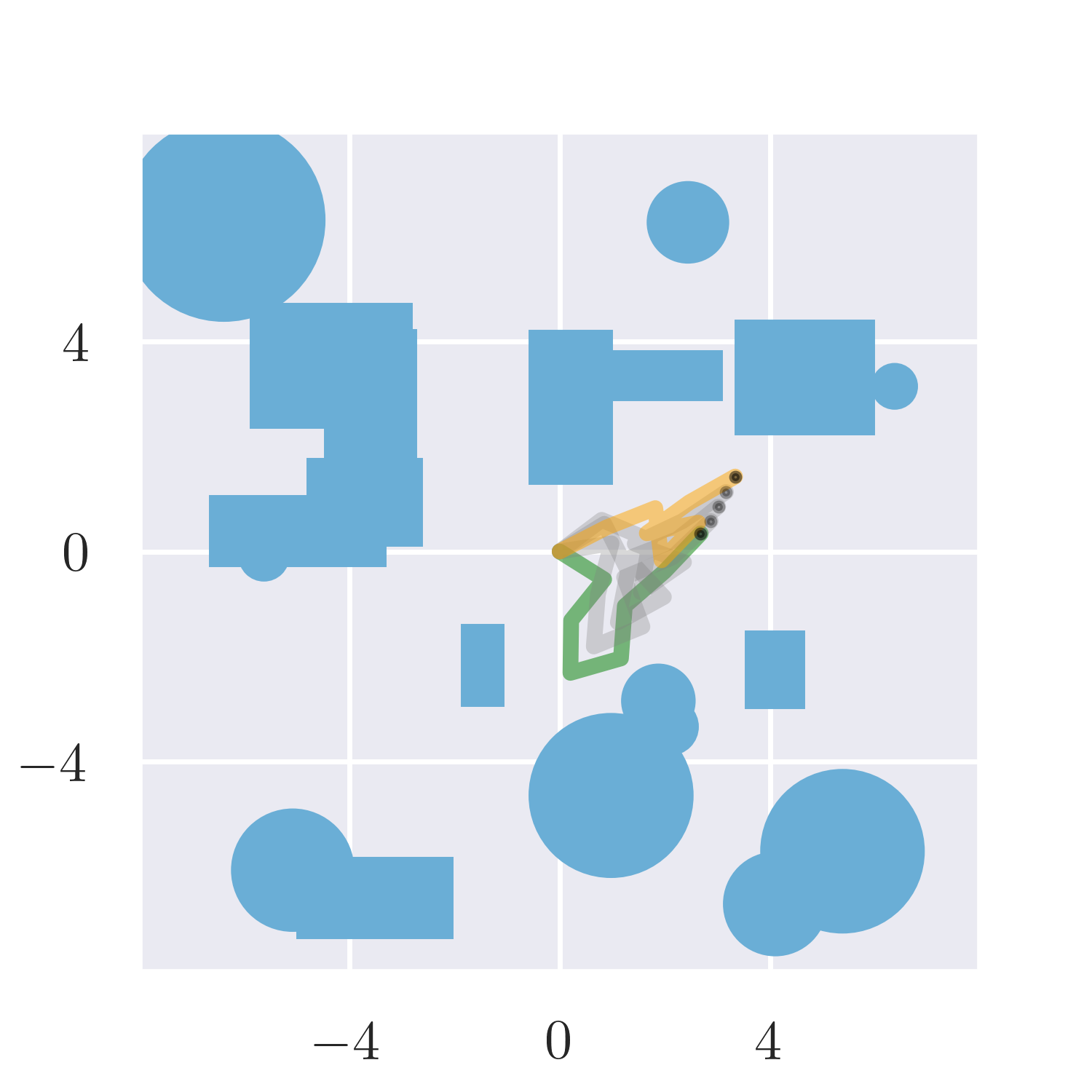}
    \caption{Examples of randomly generated environments used for quantitative comparisons, and trajectories generated by optimization using \shortname. Note that even though we only used convex geometrical shapes as basic elements of the obstacles, they can overlap with each other to form non-convex obstacles.} 
    \label{fig:planar_workspaces}
\end{figure*}

\subsection{Trajectory Optimization with Dynamic Objects}\label{sec:dynexp}
We test the dynamic adaptation functionality of the proposed active learning strategy in environments with moving obstacles. 
As Fig.~\ref{fig:dynamic} shows, the location of the obstacle is changing at every time step. The robot is asked to start from the same configuration towards the same target configuration at every time step.
Active learning strategy is able to update \shortname model accordingly, and consequently generate proper trajectories that do not collide with the moving obstacle. At first, when the obstacle does not get in the way between the start and goal configurations, the robot chose a trajectory that minimizes the movement cost. When the obstacle does get in the way, the robot chose another further but collision-free trajectory, with the help of an updated proxy collision detector. The \shortname models shown in this experiment are obtained by 
interpolating the geometrical collision distance measurements $\mathbf{D_S} \in \mathbb{R}^{M\times C}$ of the support configurations. 

\begin{figure*}[hbtp]
    \centering
    \includegraphics[width=0.97\textwidth]{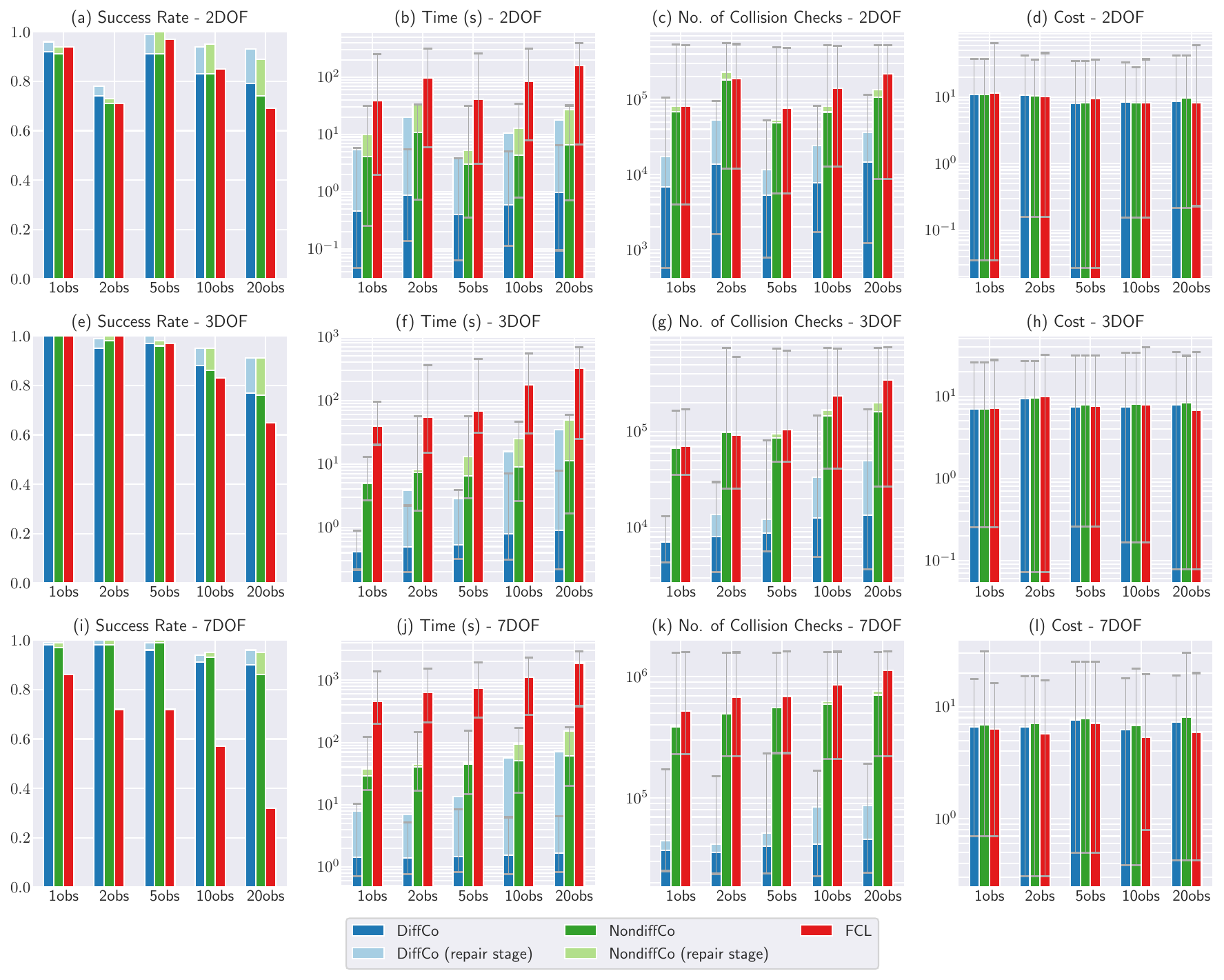} 
    \caption{Comparing \shortname and FCL \cite{pan2012fcl} when employed as collision detectors in trajectory optimization with various environments and robots. The x-axes separate results in environments containing different numbers of obstacles. Each row presents the results of the same DOF. The error bars mark the minimum and maximum value of the metric \textit{without} the repair stage. 
    When using \shortname, the optimization procedure requires less amount of time and collision checks up to 1-2 orders of magnitude, while maintaining comparable and sometimes better success rate and final solution cost, compared to the gold standard FCL. 
    The results given by \textit{NondiffCo}, the non-differentiable version of \shortname, further proves the speed advantage of using \shortname is not only due to the fact that numerical approximated gradients are less accurate than analytical solutions and take more function evaluations to compute, but also attributes to \shortname's simpler and faster model for collision detection. 
    \textbf{Note the y-axes of time, number of collision checks, and trajectory cost are in log scale}, demonstrating the orders-of-magnitude advantage \shortname provides to trajectory optimization over an classical approach. }
    \label{fig:quant}
\end{figure*}

In this example, it takes $2.423$ seconds to train a \shortname model initially, $0.271\pm 0.054$ seconds on average to update it at each time step, and $1.991\pm 2.499$ seconds on average to produce an admissible trajectory. The times of using a geometrical collision detector to produce ground-truth labels and distances are included. As a comparison, when using FCL \cite{pan2012fcl} as the collision checker and consequently using numerical differentiation during optimization in the same environment, it takes $72.145\pm 109.605$ seconds on average to produce an admissible trajectory. 
Even though the \shortname model needs updating at each time step, it still provides exceptional overall acceleration compared to FCL for trajectory optimization with dynamic objects. \updated{Trajectories obtained using FCL and \shortname have almost the same costs, both $9.732\pm 8.958$. The results are summarized in Table~\ref{tab:active}.}

\begin{figure}
    \centering
    \includegraphics[width=.55\linewidth, trim=42 38 15 15, clip ]{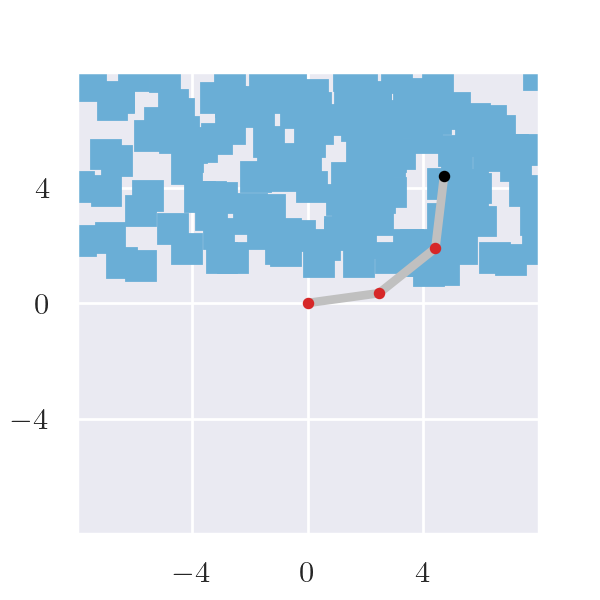}
    \caption{An environment used to compare the collision distance landscape produced by FCL and \shortname. It contains 150 randomly placed squares of size 1. The robot is a 3-link planar robot. } 
    \label{fig:halfnarrow}
\end{figure}

\begin{figure}
    \centering
    \includegraphics[width=\linewidth]{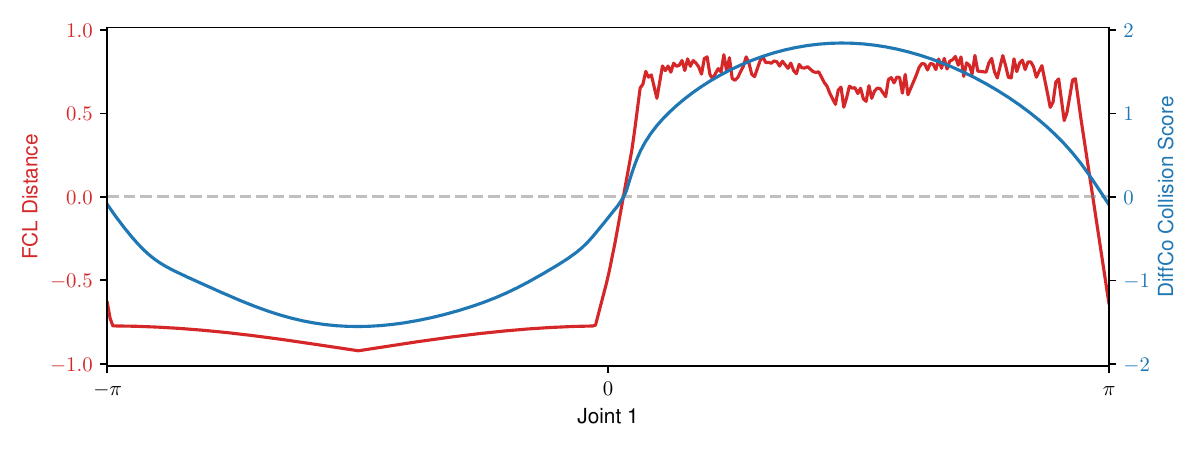}
    \caption{In the environment presented in Fig.~\ref{fig:halfnarrow}, this is how the collision distance (score) given by FCL and \shortname looks like when the first joint angle changes from $-\pi$ to $\pi$ while the other two joint angles are fixed at 0. The gradient of the distance given by FCL is very noisy in the in-collision area, which can potentially be a reason why the success rate of trajectory optimization is low when using FCL as the collision detector.} 
    \label{fig:landscape}
\end{figure}

\begin{figure*}[hbtp]
    \centering
    \includegraphics[width=\linewidth]{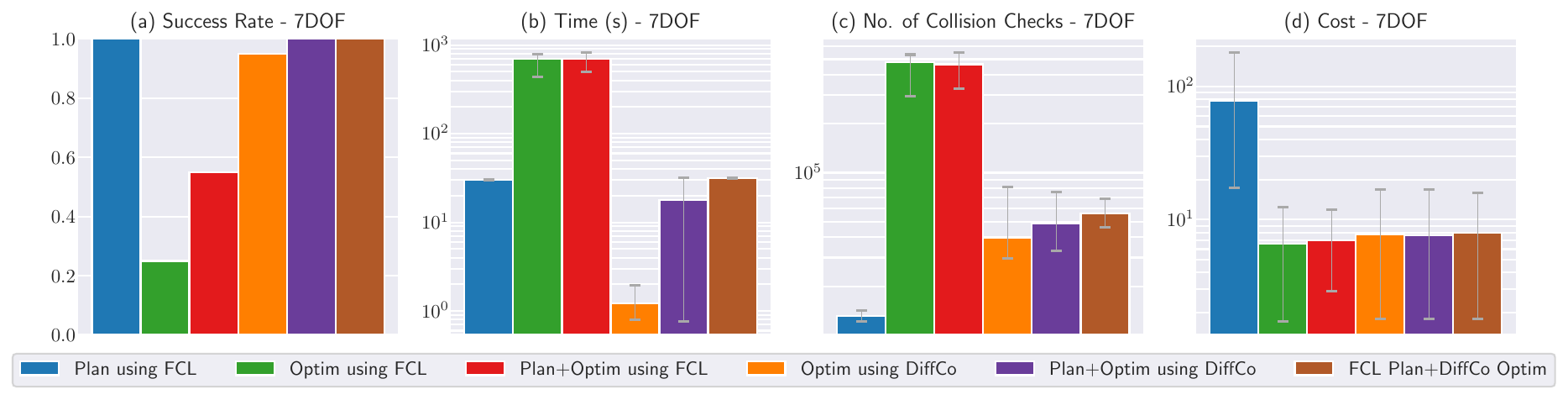}
    \caption{Quantitative comparison of different pipelines for optimal planning. Compared to a completeness-guaranteed optimal planner RRT*, optimization done with \shortname exhibits significant reduction in path cost while maintaining a comparable success rate and computational time. On the other hand, RRT* is able to warm start the optimization stage and improve the success rate. } 
    \label{fig:hybrid}
\end{figure*}
\begin{figure*}[hbtp]
    \centering
    \includegraphics[width=\linewidth]{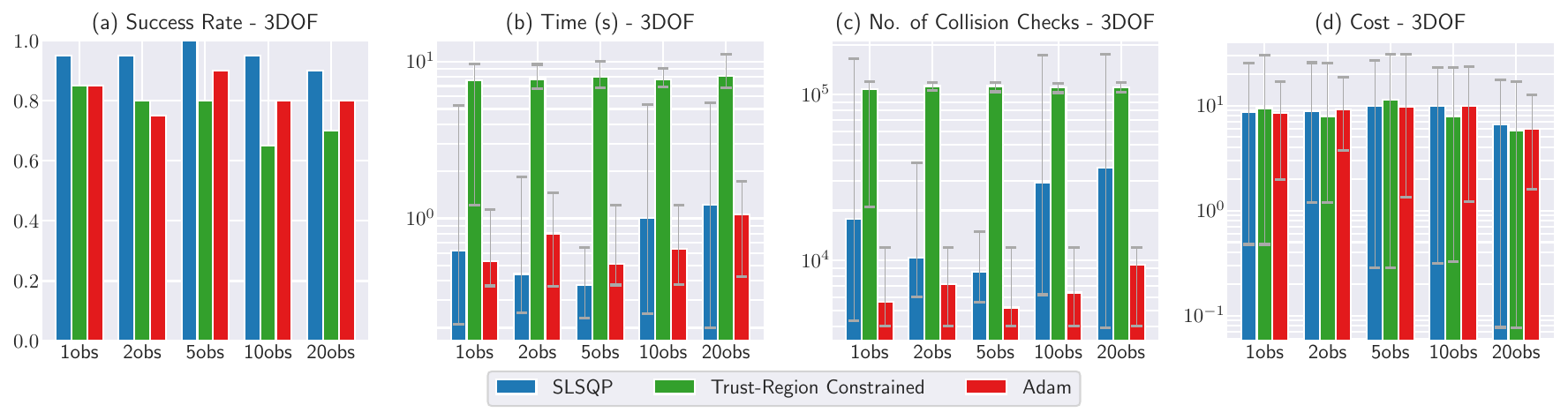}
    \caption{Comparison of different optimization algorithms using the gradients provided by \shortname. All algorithms are able to maintain comparable solution costs. Trust-Region Constrained \cite{trustregion} relies on Hessian matrix, which is numerically approximated using the analytical gradients provided by \shortname, resulting in a lower success rate, more computational time and an increased number of collision checks.} 
    \label{fig:differentoptimizer}
\end{figure*}

\subsection{Comparison in Randomly Generated Environments}\label{sec:quant} 
We test \shortname and FCL \cite{pan2012fcl} in trajectory optimization tasks in a set of randomly generated environments to demonstrate the benefit of using a differentiable collision detector in trajectory optimization. \updated{Fig.~\ref{fig:planar_workspaces} contains some examples of the environments and robot trajectories.}


Since FCL is not analytically differentiable, we selected SLSQP \cite{slsqpKraft1988software, slsqpKraft1994algorithm} as the optimizer, \updated{which has both an option for using approximated gradients by finite differencing}, 
and another option for using analytical gradients. 
Considering the fact that gradients obtained by finite differencing may provide a slower convergence rate, besides \shortname and FCL, we also included the results when employing the non-differentiable output of \shortname (denoted as \textit{NondiffCo}) as the collision detector. NondiffCo and \shortname outputs the exactly same collision scores, but instead of using backpropagation to compute analytical gradients, it relies on the numerical approximations provided by SLSQP like FCL does.

For each robot of 2, 3, and 7 DOF, 50 environments with different numbers of randomly placed and sized obstacles \updated{(circles and rectangles)} are tested. For each environment, every method is required to find collision-free paths between 10 pairs of randomly selected start and goal configurations. Every method is allowed up to 3 retries upon failure to find a feasible solution. The first try always starts from the initialization of a straight line between the start and goal in the $\mathcal{C}$-space. Later tries use way-points picked uniformly at random from the whole $\mathcal{C}$-space as initializations. Each try is allowed up to 200 iterations. 
For \shortname and NondiffCo, if the path is still inadmissible, we \textit{repair} the solution by using FCL as the collision detector \updated{and running SLSQP for at most another 200 iterations. The repair stages are exactly the same as the initial trajectory optimization, except the collision detector is switched to FCL and they use the best solutions seen so far as the seeds of optimization.} 
Metric changes brought in by the repair stages are marked in light color in Fig.~\ref{fig:quant}.

Note obstacles in environments with 1 or 2 obstacles are generated in a tighter area than the ones in environments with more obstacles, thus bringing worse performance metrics in some cases. All obstacles in all environments do not intersect with the origin of the robots, in which case the robot will always be in collision. The success rate, time consumed, number of collision checks, and the trajectory cost of admissible solutions are depicted in Fig.~\ref{fig:quant}.


Note that for methods using \shortname, the model needs to be trained once on ground-truth labels provided by the geometric collision detector, FCL \cite{pan2012fcl} before solving all 10 queries in each environment. 
Naturally, the more queries are made in the same environment, on average, the less time per query is spent training the model. 
Since training is not needed for every query and indeed should likely only be run when the robot is first turned on (whereas afterwards active learning should be used if the environment changes), the time consumed by initial training is not presented in Fig.~\ref{fig:quant} but in Table~\ref{tab:training_time} to avoid confusion. 
\begin{table}[hbtp]
\centering
\caption{Time consumed by one-time initial training of \shortname in each environment (in seconds). After the initial training, the model is ready to solve arbitrary number of trajectory queries in the same environment. }
\label{tab:training_time}
\resizebox{\linewidth}{!}{%
\begin{tabular}{@{}cccc@{}}
\toprule
\multirow[b]{2}{*}{\begin{tabular}[c]{@{}c@{}}Number of \\ obstacles\end{tabular}} & \multicolumn{3}{c}{Degrees of freedom} \\ \cmidrule(l){2-4} 
  & 2                  & 3                  & 7                  \\ \midrule
1  & $2.304\pm 0.093$   & $3.360 \pm 0.488$  & $5.913 \pm 1.695$  \\
2  & $2.489 \pm 0.146$  & $4.028 \pm 0.589$  & $8.379 \pm 2.007$  \\
5  & $ 2.578 \pm 0.193$ & $4.794 \pm 1.242$  & $8.116 \pm 3.074$  \\
10 & $3.060 \pm 0.153$  & $ 6.490 \pm 1.320$ & $10.587 \pm 2.404$ \\
20 & $3.476 \pm 0.148$  & $8.561 \pm 0.764$  & $13.341 \pm 1.917$ \\ \bottomrule
\end{tabular}%
}
\end{table}
The numbers include the time used to acquire ground-truth labels from the geometric collision detector and the time to train the model. 
Nonetheless, the orders-of-magnitude advantage in computation time of using \shortname persists even when taking the per-query average time spent on training into account. In any case, if one were to work in an environment with constantly changing objects, the active learning method presented in Section~\ref{sec:active} and \ref{sec:dynexp} should be used to \textit{update} the \shortname model, instead of re-training the model from scratch at every time step.

\updated{The lower success rate when doing trajectory optimization with FCL may raise some concerns because conceptually, FCL is the standard that \shortname tries to learn. We suspect this lower success rate may come from the fact that it is non-trivial to implement an algorithm to calculate (or even define) the penetration depth between a non-rigid robot and a group of obstacles. As a compromise, FCL computes the deepest penetration between all \textit{pairs} of robot links and obstacles.}

\updated{We attempt to show a negative effect of this compromise in the environment presented in Fig.~\ref{fig:halfnarrow}:
we set every except the first joint angle of the 3-link robot to 0, and plot the collision distance (score) given by FCL and a \shortname model when moving the first joint (Fig.~\ref{fig:landscape}). 
It can be observed that the gradient of FCL collision distance in the in-collision area is very noisy as a consequence of calculating only \textit{pairwise} rigid-body penetrations, which can make it hard for a gradient-based optimizer to get an in-collision configuration out of collision.
}

\updated{\shortname, on the other hand, steered around this problem by fitting the distance score of only the support points, which are critical configurations near the decision boundary empirically. This helps \shortname provide a smooth function of collision score, making it easy for optimizers to get a configuration out of collision. This also partially explained why the FCL distance and \shortname score in Fig.~\ref{fig:correlation} and Fig.~\ref{fig:selfcollisioncorrelation} are not perfectly correlated.}

\subsection{\updated{Comparison with a Sampling-based Optimal Planner}} 
We further compare trajectory optimization using \shortname and FCL against a sampling-based optimal planner, RRT* \cite{rrtstar} using a subset of environments, robots, and queries from Section~\ref{sec:quant}. Essentially 3 kinds of optimal planning pipelines are involved: 
\begin{enumerate}
    \item only using a sampling-based optimal planner,
    \item trajectory optimization starting from a straight-line seed, and
    \item trajectory optimization using a sampling-based solution as the seed.
\end{enumerate}
By alternating the collision detector, 6 optimal planning pipelines are compared: 1) RRT* using FCL; 2) optimization using FCL; 3) RRT* and optimization using FCL; 4) optimization using \shortname; 5) RRT* and optimization using \shortname; 6) RRT* using FCL and optimization using \shortname. RRT* is run for 30 seconds in all pipelines. We chose the most challenging cases with 20 obstacles in the workspace and a 7-DOF robot. For each environment, 2 queries are attempted by each pipeline. The success rate, computational times, numbers of collision checks, and the path cost are shown in Fig.~\ref{fig:hybrid}.

According to the costs of the solutions, methods that use optimization, no matter using FCL or \shortname, are more effective than the sampling-based optimal planner in achieving lower costs. With that said, optimization done with DiffCo is better at maintaining the computational time and the success rate.

Comparing the method using RRT* with FCL and optimization with \shortname (\textit{FCL Plan+\shortname Optim}) to only RRT* with FCL (\textit{Plan using FCL}) and only optimization using \shortname (\textit{Optim using \shortname)}, we can see a) warm starting the optimization procedure using admissible solution produced by completeness-guaranteed planner improves the success rate, and b) optimization as a post-processing can significantly reduce costs of the final solutions.

These results indicate that a proper combination of sampling-based planners and trajectory optimization with \shortname can achieve much more optimal solutions almost without sacrificing any robustness and computational time.

\subsection{\updated{Influence of Optimization Algorithms}} 
To show that the gradients provided by \shortname can generalize to more optimization algorithms, we let \shortname work with SLSQP, Trust-Region Constrained, and Adam \cite{slsqpKraft1994algorithm, trustregion, adam} on a subset of environments, robots, and queries from Section~\ref{sec:quant}.

It can be observed in Fig.~\ref{fig:differentoptimizer} that all algorithms are able to achieve comparable solution costs. However, Trust-Region Constrained achieved a lower success rate while taking more computational time and more collision checks than the other two algorithms. This is largely due to the fact that the Trust-Region method relies on Hessian matrix, which is numerically approximated using the analytical gradients provided by \shortname. Adam takes the least time in most cases because Adam does not contain nested loops for inexact line searches like the other two, so it triggers fewer collision checks. On the other hand, Adam has a lower success rate than SLSQP due to its limited ability to work with nonlinear constraints. 

In all cases, \shortname is able to help the optimizers solve most queries, producing admissible solutions of low costs.



\section{Conclusion}
In this paper, we presented a differentiable proxy collision detector, \shortname.
It models the configuration space using a non-parametric kernel perceptron, which allows faster computation than geometrical methods. It is able to provide instance-level proxy collision detection such that different levels of safety biases can be attached to individual objects. The computation of \shortname happens in a completely differentiable way, which enables the auto-differentiation packages widely available in deep learning libraries such as PyTorch to be directly applied to it. We further show its model can be updated online using an active learning approach. Finally, we show its application in constrained trajectory optimization where it is employed as the collision detector to help generate safety-optimized and semantically informed trajectories. Extensive quantitative results demonstrate its efficiency and efficacy compared to geometrical non-differentiable methods.

Since the collision score of \shortname is not a geometrical distance but just a kind of \textit{pseudo} distance, a future extension of interest is to approximate the geometrical distance to facilitate some robotic applications that require more accurate geometrical information. 
\second{Also, the active learning pipeline may be improved by a smarter sampling strategy, e.g., one that uses the movement of objects to bias the sampling distribution.}
Stochastic planning methods such as MPNet \cite{qureshi2019motion} often needs a re-planning procedure to fix inadmissible way-points, which takes a considerable portion of the overall time. \shortname, combined with optimization algorithms, may be used to accelerate this subroutine and improve the overall performance. \shortname may also be integrated into existing trajectory optimization frameworks such as Drake, CHOMP and TrajOpt \cite{drake, chomp, trajopt} to provide smooth analytical gradients for both in-collision and collision-free robot configurations. Last but not least, future work may try to tackle the challenge of completely eliminating ground-truth geometrical collision detection algorithms in the procedure of creating proxy collision detectors, which has been the main performance bottleneck of online model updates.



\appendix
\noindent\textit{Kernel Selection:} 
As all kernel-based learning methods are, the choice of the kernel function is vital to the performance of the kernel perceptron. A kernel function specially designed for proxy collision checking \cite{das2020forward}, forward-kinematics kernel $k_{\FK}$, is favored in our algorithm. On top of the rational quadratic (RQ) kernel,
\begin{equation}
    k_{\mathrm{RQ}}(\cfg, \cfg') = (1+\frac{\gamma}{p}\|\cfg-\cfg'\|^2)^{-p},
\end{equation}
$k_{\FK}$ adds forward-kinematics transforms to the input configurations:
\begin{align}
    & \FK_m(\cfg) = \mathrm{pos}(_w\mathbf{T}_m), \\
    & \FK(\cfg) = \left[\FK_1(\cfg), \FK_2(\cfg), ..., \FK_M(\cfg)\right], 
\end{align}
where each $m\in \{1, ..., M\}$ denotes an arbitrarily pre-defined \textit{control point} on a robot link, and $\mathrm{pos}(_w\mathbf{T}_m)$ extracts the position elements in the pose of the $m$-th control point in the world frame. So instead of directly inputting the configuration $\cfg$ and $\cfg'$ to $k_\mathrm{RQ}$,
\begin{align}
    k_{\FK}(\cfg, \cfg') = k_{\mathrm{RQ}}(\FK(\cfg), \FK(\cfg')).
\end{align}
In this paper, we select the control points to be unique joint positions on a robot, including the tip of the end effector. 

\begin{table}[hbtp]
\centering
\caption{Mathematical symbols used in this paper.}
\label{tab:symbols}
\begin{tabularx}{.9\linewidth}{@{}cX@{}}
\toprule
Symbol                                  & Definition                                                                                                                \\ \midrule
\begin{tabular}{c}$\mathcal{C}, \mathcal{C}_\mathrm{obs},$\\ $\mathcal{C}_\mathrm{free}$\end{tabular} & Configuration space, obstacle subspace, free subspace.                                                                     \\
$\mathcal{R}_i$                         & All points of the $i$-th link of a robot.                                                                                  \\
$o_j$                                   & All points of $j$-th object in the workspace.                                                                              \\
$\cfg$                                  & A configuration vector.                                                                                                    \\
$\Cfg$                                  & A set of configurations expressed in a matrix. Each row is a configuration.                                               \\
$N$                                     & The number of configurations.                                                                                             \\
$D$                                     & The dimension of a configuration, i.e., the DOF of a robot.                                                               \\
\updated{$d_\mathcal{R}$}                        & \updated{Distance to collision used in this paper.}                                                                                 \\
$\mathbf{y}$                            & A vector of collision labels with different categories of objects. +1 for in-collision, -1 for collision-free.                                                            \\
$\mathbf{Y}$                            & A set of collision labels expressed in a matrix. Each row is the collision labels of a configuration with different objects.                                    \\
$c$                                     & An obstacle category.                                                                                                     \\
$C$                                     & The number of obstacle categories. Sometimes used as the set of all obstacle categories.                                  \\
$\mathbf{S}$                            & Support configurations.                                                                                                   \\
$M$                                     & The number of support configurations.                                                                                      \\
$k$                                     & A kernel function.                                                                                                        \\
$\mathbf{K}$                            & A kernel matrix.                                                                                                          \\
$\mathbf{H}$                            & A hypothesis matrix.                                                                                                      \\
$\mathbf{M}$                            & A matrix of margins.                                                                                                      \\
$\mathbf{W}$                            & The weight matrix of the kernel perceptron.                                                                               \\
$\mathbf{A}$                            & The weight matrix of the differentiable model using the polyharmonic function.                                        \\
$\FK$                                   & The forward-kinematics transform.                                                                                         \\
\begin{tabular}{c}$\mathrm{EST}, $\\$\mathrm{EST}_c$\end{tabular}          & The proxy distance estimation to all categories of obstacles. The proxy distance estimation to obstacles of category $c$. \\
$\epsilon_c$                            & The safety bias attached to obstacles of the category $c$.                                                                           \\
$\mathrm{ee}$                           & The position of the end effector.                                                                                         \\
$\nu, \zeta, \sigma$                    & The number of configurations sampled in exploitation and exploration, and the scale of sampling in exploitation. \\
\bottomrule
\end{tabularx}%
\end{table}

\bibliographystyle{IEEEtran}
\bibliography{main}

\begin{thebibliography}{10}
\providecommand{\url}[1]{#1}
\csname url@samestyle\endcsname
\providecommand{\newblock}{\relax}
\providecommand{\bibinfo}[2]{#2}
\providecommand{\BIBentrySTDinterwordspacing}{\spaceskip=0pt\relax}
\providecommand{\BIBentryALTinterwordstretchfactor}{4}
\providecommand{\BIBentryALTinterwordspacing}{\spaceskip=\fontdimen2\font plus
\BIBentryALTinterwordstretchfactor\fontdimen3\font minus
  \fontdimen4\font\relax}
\providecommand{\BIBforeignlanguage}[2]{{%
\expandafter\ifx\csname l@#1\endcsname\relax
\typeout{** WARNING: IEEEtran.bst: No hyphenation pattern has been}%
\typeout{** loaded for the language `#1'. Using the pattern for}%
\typeout{** the default language instead.}%
\else
\language=\csname l@#1\endcsname
\fi
#2}}
\providecommand{\BIBdecl}{\relax}
\BIBdecl

\bibitem{pan2012fcl}
J.~Pan, S.~Chitta, and D.~Manocha, ``Fcl: A general purpose library for
  collision and proximity queries,'' in \emph{2012 IEEE International
  Conference on Robotics and Automation}.\hskip 1em plus 0.5em minus
  0.4em\relax IEEE, 2012, pp. 3859--3866.

\bibitem{libccd}
\BIBentryALTinterwordspacing
Libccd. [Online]. Available: \url{https://github.com/danfis/libccd}
\BIBentrySTDinterwordspacing

\bibitem{gjk}
E.~G. {Gilbert}, D.~W. {Johnson}, and S.~S. {Keerthi}, ``A fast procedure for
  computing the distance between complex objects in three-dimensional space,''
  \emph{IEEE Journal on Robotics and Automation}, vol.~4, no.~2, pp. 193--203,
  1988.

\bibitem{epa}
G.~Van Den~Bergen, ``Proximity queries and penetration depth computation on 3d
  game objects,'' in \emph{Game developers conference}, vol. 170, 2001.

\bibitem{kingston2018sampling}
Z.~Kingston, M.~Moll, and L.~E. Kavraki, ``Sampling-based methods for motion
  planning with constraints,'' \emph{Annual review of control, robotics, and
  autonomous systems}, vol.~1, pp. 159--185, 2018.

\bibitem{elbanhawi2014sampling}
M.~Elbanhawi and M.~Simic, ``Sampling-based robot motion planning: A review,''
  \emph{Ieee access}, vol.~2, pp. 56--77, 2014.

\bibitem{das2020learning}
N.~Das and M.~Yip, ``Learning-based proxy collision detection for robot motion
  planning applications,'' \emph{IEEE Transactions on Robotics}, 2020.

\bibitem{das2020forward}
N.~Das and M.~C. Yip, ``Forward kinematics kernel for improved proxy collision
  checking,'' \emph{IEEE Robotics and Automation Letters}, vol.~5, no.~2, pp.
  2349--2356, 2020.

\bibitem{pan2015efficient}
J.~Pan and D.~Manocha, ``Efficient configuration space construction and
  optimization for motion planning,'' \emph{Engineering}, vol.~1, no.~1, pp.
  046--057, 2015.

\bibitem{huh2016learning}
J.~Huh and D.~D. Lee, ``Learning high-dimensional mixture models for fast
  collision detection in rapidly-exploring random trees,'' in \emph{2016 IEEE
  International Conference on Robotics and Automation (ICRA)}.\hskip 1em plus
  0.5em minus 0.4em\relax IEEE, 2016, pp. 63--69.

\bibitem{trajopt}
\BIBentryALTinterwordspacing
J.~Schulman, J.~Ho, A.~Lee, I.~Awwal, H.~Bradlow, and P.~Abbeel, ``{Finding
  Locally Optimal, Collision-Free Trajectories with Sequential Convex
  Optimization},'' in \emph{Robotics: Science and Systems}.\hskip 1em plus
  0.5em minus 0.4em\relax Robotics: Science and Systems Foundation, jun 2013.
  [Online]. Available: \url{http://www.roboticsproceedings.org/rss09/p31.pdf}
\BIBentrySTDinterwordspacing

\bibitem{chomp}
M.~Zucker, N.~Ratliff, A.~D. Dragan, M.~Pivtoraiko, M.~Klingensmith, C.~M.
  Dellin, J.~A. Bagnell, and S.~S. Srinivasa, ``{CHOMP: Covariant Hamiltonian
  optimization for motion planning},'' \emph{International Journal of Robotics
  Research}, vol.~32, no. 9-10, pp. 1164--1193, 2013.

\bibitem{kalakrishnan2011stomp}
M.~Kalakrishnan, S.~Chitta, E.~Theodorou, P.~Pastor, and S.~Schaal, ``Stomp:
  Stochastic trajectory optimization for motion planning,'' in \emph{2011 IEEE
  international conference on robotics and automation}.\hskip 1em plus 0.5em
  minus 0.4em\relax IEEE, 2011, pp. 4569--4574.

\bibitem{heiden2018grips}
E.~Heiden, L.~Palmieri, S.~Koenig, K.~O. Arras, and G.~S. Sukhatme,
  ``Gradient-informed path smoothing for wheeled mobile robots,'' in
  \emph{International Conference on Robotics and Automation (ICRA)}.\hskip 1em
  plus 0.5em minus 0.4em\relax IEEE, 2018.

\bibitem{qureshi2019motion}
A.~H. Qureshi, Y.~Miao, A.~Simeonov, and M.~C. Yip, ``Motion planning networks:
  Bridging the gap between learning-based and classical motion planners,''
  \emph{IEEE Transactions on Robotics}, pp. 1--9, 2020.

\bibitem{ichter2018learning}
B.~Ichter, J.~Harrison, and M.~Pavone, ``Learning sampling distributions for
  robot motion planning,'' in \emph{2018 IEEE International Conference on
  Robotics and Automation (ICRA)}.\hskip 1em plus 0.5em minus 0.4em\relax IEEE,
  2018, pp. 7087--7094.

\bibitem{oraclenet}
M.~J. Bency, A.~H. Qureshi, and M.~C. Yip, ``Neural path planning: Fixed time,
  near-optimal path generation via oracle imitation,'' in \emph{2019 IEEE/RSJ
  International Conference on Intelligent Robots and Systems (IROS)}.\hskip 1em
  plus 0.5em minus 0.4em\relax IEEE, 2019, pp. 3965--3972.

\bibitem{qureshi2020constrained}
A.~H. Qureshi, J.~Dong, A.~Baig, and M.~C. Yip, ``Constrained motion planning
  networks x,'' \emph{arXiv preprint arXiv:2010.08707}, 2020.

\bibitem{qureshi2018deeply}
A.~H. Qureshi and M.~C. Yip, ``Deeply informed neural sampling for robot motion
  planning,'' in \emph{2018 IEEE/RSJ International Conference on Intelligent
  Robots and Systems (IROS)}.\hskip 1em plus 0.5em minus 0.4em\relax IEEE,
  2018, pp. 6582--6588.

\bibitem{gpmp2}
J.~Dong, M.~Mukadam, F.~Dellaert, and B.~Boots, ``Motion planning as
  probabilistic inference using gaussian processes and factor graphs.'' in
  \emph{Robotics: Science and Systems}, vol.~12, 2016, p.~4.

\bibitem{escande2014strictly}
A.~Escande, S.~Miossec, M.~Benallegue, and A.~Kheddar, ``A strictly convex hull
  for computing proximity distances with continuous gradients,'' \emph{IEEE
  Transactions on Robotics}, vol.~30, no.~3, pp. 666--678, 2014.

\bibitem{kew2019neural}
J.~C. Kew, B.~Ichter, M.~Bandari, T.-W.~E. Lee, and A.~Faust, ``Neural
  collision clearance estimator for fast robot motion planning,'' \emph{arXiv
  preprint arXiv:1910.05917}, 2019.

\bibitem{das2020stochastic}
N.~{Das} and M.~C. {Yip}, ``Stochastic modeling of distance to collision for
  robot manipulators,'' \emph{IEEE Robotics and Automation Letters}, vol.~6,
  no.~1, pp. 207--214, 2021.

\bibitem{wilcox2020solar}
B.~Wilcox and M.~C. Yip, ``Solar-gp: Sparse online locally adaptive regression
  using gaussian processes for bayesian robot model learning and control,''
  \emph{IEEE Robotics and Automation Letters}, vol.~5, no.~2, pp. 2832--2839,
  2020.

\bibitem{pan2013efficient}
J.~Pan, X.~Zhang, and D.~Manocha, ``Efficient penetration depth approximation
  using active learning,'' \emph{ACM Transactions on Graphics (TOG)}, vol.~32,
  no.~6, pp. 1--12, 2013.

\bibitem{flacco2012depth}
F.~Flacco, T.~Kr{\"o}ger, A.~De~Luca, and O.~Khatib, ``A depth space approach
  to human-robot collision avoidance,'' in \emph{2012 IEEE International
  Conference on Robotics and Automation}.\hskip 1em plus 0.5em minus
  0.4em\relax IEEE, 2012, pp. 338--345.

\bibitem{rakita2018relaxedik}
D.~Rakita, B.~Mutlu, and M.~Gleicher, ``Relaxedik: Real-time synthesis of
  accurate and feasible robot arm motion.'' in \emph{Robotics: Science and
  Systems}.\hskip 1em plus 0.5em minus 0.4em\relax Pittsburgh, PA, 2018, pp.
  26--30.

\bibitem{rajaram2016refinenet}
R.~N. Rajaram, E.~Ohn-Bar, and M.~M. Trivedi, ``Refinenet: Refining object
  detectors for autonomous driving,'' \emph{IEEE Transactions on Intelligent
  Vehicles}, vol.~1, no.~4, pp. 358--368, 2016.

\bibitem{objcontent2}
J.~Evans, P.~Patr{\'o}n, B.~Smith, and D.~M. Lane, ``Design and evaluation of a
  reactive and deliberative collision avoidance and escape architecture for
  autonomous robots,'' \emph{Autonomous Robots}, vol.~24, no.~3, pp. 247--266,
  2008.

\bibitem{objcontent3}
R.~Mojtahedzadeh, ``Safe robotic manipulation to extract objects from piles:
  from 3d perception to object selection,'' Ph.D. dissertation, {\"O}rebro
  university, 2016.

\bibitem{uber2020perceive}
A.~Sadat, S.~Casas, M.~Ren, X.~Wu, P.~Dhawan, and R.~Urtasun, ``Perceive,
  predict, and plan: Safe motion planning through interpretable semantic
  representations,'' in \emph{Proceedings of the European Conference on
  Computer Vision (ECCV)}, 2020.

\bibitem{adam}
\BIBentryALTinterwordspacing
D.~P. Kingma and J.~Ba, ``Adam: {A} method for stochastic optimization,'' in
  \emph{3rd International Conference on Learning Representations, {ICLR} 2015,
  San Diego, CA, USA, May 7-9, 2015, Conference Track Proceedings}, Y.~Bengio
  and Y.~LeCun, Eds., 2015. [Online]. Available:
  \url{http://arxiv.org/abs/1412.6980}
\BIBentrySTDinterwordspacing

\bibitem{slsqpKraft1994algorithm}
D.~Kraft, ``Algorithm 733: Tomp--fortran modules for optimal control
  calculations,'' \emph{ACM Transactions on Mathematical Software (TOMS)},
  vol.~20, no.~3, pp. 262--281, 1994.

\bibitem{slsqpKraft1988software}
D.~Kraft \emph{et~al.}, \emph{A software package for sequential quadratic
  programming}.\hskip 1em plus 0.5em minus 0.4em\relax DFVLR Obersfaffeuhofen,
  Germany, 1988.

\bibitem{svm}
J.~A. Suykens and J.~Vandewalle, ``Least squares support vector machine
  classifiers,'' \emph{Neural processing letters}, vol.~9, no.~3, pp. 293--300,
  1999.

\bibitem{kernelperceptron}
Y.~Freund and R.~E. Schapire, ``Large margin classification using the
  perceptron algorithm,'' \emph{Machine learning}, vol.~37, no.~3, pp.
  277--296, 1999.

\bibitem{hangelbroek2013density}
T.~Hangelbroek and J.~Levesley, ``On the density of polyharmonic splines,''
  \emph{Journal of Approximation Theory}, vol. 167, pp. 94--108, 2013.

\bibitem{paszke2019pytorch}
A.~Paszke, S.~Gross, F.~Massa, A.~Lerer, J.~Bradbury, G.~Chanan, T.~Killeen,
  Z.~Lin, N.~Gimelshein, L.~Antiga \emph{et~al.}, ``Pytorch: An imperative
  style, high-performance deep learning library,'' in \emph{Advances in neural
  information processing systems}, 2019, pp. 8026--8037.

\bibitem{pythonfcl}
\BIBentryALTinterwordspacing
Python-fcl. [Online]. Available:
  \url{https://github.com/BerkeleyAutomation/python-fcl}
\BIBentrySTDinterwordspacing

\bibitem{trustregion}
R.~H. Byrd, M.~E. Hribar, and J.~Nocedal, ``An interior point algorithm for
  large-scale nonlinear programming,'' \emph{SIAM Journal on Optimization},
  vol.~9, no.~4, pp. 877--900, 1999.

\bibitem{2020SciPy-NMeth}
P.~Virtanen, R.~Gommers, T.~E. Oliphant, M.~Haberland, T.~Reddy, D.~Cournapeau,
  E.~Burovski, P.~Peterson, W.~Weckesser, J.~Bright, S.~J. {van der Walt},
  M.~Brett, J.~Wilson, K.~J. Millman, N.~Mayorov, A.~R.~J. Nelson, E.~Jones,
  R.~Kern, E.~Larson, C.~J. Carey, {\.I}.~Polat, Y.~Feng, E.~W. Moore,
  J.~{VanderPlas}, D.~Laxalde, J.~Perktold, R.~Cimrman, I.~Henriksen, E.~A.
  Quintero, C.~R. Harris, A.~M. Archibald, A.~H. Ribeiro, F.~Pedregosa, P.~{van
  Mulbregt}, and {SciPy 1.0 Contributors}, ``{{SciPy} 1.0: Fundamental
  Algorithms for Scientific Computing in Python},'' \emph{Nature Methods},
  vol.~17, pp. 261--272, 2020.

\bibitem{Merkt2018}
W.~Merkt, V.~Ivan, and S.~Vijayakumar, ``{Leveraging Precomputation with
  Problem Encoding for Warm-Starting Trajectory Optimization in Complex
  Environments},'' in \emph{IEEE International Conference on Intelligent Robots
  and Systems}.\hskip 1em plus 0.5em minus 0.4em\relax Institute of Electrical
  and Electronics Engineers Inc., dec 2018, pp. 5877--5884.

\bibitem{collisionbook}
M.~C. Lin and D.~Manocha, \emph{Collision and proximity queries}.\hskip 1em
  plus 0.5em minus 0.4em\relax Citeseer, 2003.

\bibitem{rrtstar}
S.~Karaman and E.~Frazzoli, ``Sampling-based algorithms for optimal motion
  planning,'' \emph{The international journal of robotics research}, vol.~30,
  no.~7, pp. 846--894, 2011.

\bibitem{drake}
\BIBentryALTinterwordspacing
R.~Tedrake and the Drake Development~Team, ``Drake: Model-based design and
  verification for robotics,'' 2019. [Online]. Available:
  \url{https://drake.mit.edu}
\BIBentrySTDinterwordspacing

\end{thebibliography}
\balance

\end{document}